\documentclass[final,onefignum,onetabnum]{siamart190516}

\usepackage{lipsum}
\usepackage{amsfonts}
\usepackage{graphicx}
\usepackage{algorithmic}
\usepackage{todonotes}
\usepackage{amssymb}
\usepackage{amsmath}
\usepackage[section]{placeins}
\usepackage{fixfoot}
\usepackage[hyperfootnotes=false]{hyperref}

\usepackage{array,makecell}
\usepackage{multirow}
\usepackage{colortbl}
\usepackage{hhline} 
\newcommand{\mc}{\multicolumn{1}{c}}
\newcommand{\mcr}{\multicolumn{1}{c|}}

\definecolor{darkpastelgreen}{rgb}{0.01, 0.75, 0.24}

\usepackage[utf8]{inputenc}
\usepackage{pgfplots}
\DeclareUnicodeCharacter{2212}{−}
\usepgfplotslibrary{groupplots,dateplot}
\pgfplotsset{compat=newest}
\usepackage{varwidth}

\usepackage{tikz}
\usetikzlibrary{patterns,shapes.geometric, arrows}

\makeatletter
\newcommand*{\addFileDependency}[1]{
  \typeout{(#1)}
  \@addtofilelist{#1}
  \IfFileExists{#1}{}{\typeout{No file #1.}}
}
\makeatother

\ifpdf
  \DeclareGraphicsExtensions{.jpg,.eps,.pdf,.png} 
\else
  \DeclareGraphicsExtensions{.eps}
\fi

\newsiamremark{remark}{Remark}
\newsiamremark{hypothesis}{Hypothesis}
\crefname{hypothesis}{Hypothesis}{Hypotheses}
\newsiamthm{claim}{Claim}

\headers{Neural texture synthesis long range constraints}{N. Gonthier, Y. Gousseau, and S. Ladjal} 

\title{High resolution neural texture synthesis with long range constraints}

\author{Nicolas Gonthier$^{\dagger}$\thanks{Université Paris-Saclay, 91190, Saint-Aubin, France}
\and Yann Gousseau\thanks{LTCI, Télécom Paris, Institut polytechnique de Paris, 19 Place Marguerite Perey, 91120
Palaiseau, France (\email{nicolas.gonthier@telecom-paris.fr}).}
\and Saïd Ladjal\footnotemark[2]}
\usepackage{amsopn}

\ifpdf
\hypersetup{
  pdftitle={High resolution neural texture synthesis with long range constraints},
  pdfauthor={N. Gonthier, Y. Gousseau, and S. Ladjal}
}
\fi

\usepackage{fancyhdr}
\begin{document}

\maketitle

\begin{abstract}
The field of texture synthesis has witnessed important progresses over the last years, most notably through the use of Convolutional Neural Networks. However, neural synthesis methods still struggle to reproduce large scale structures, especially with high resolution textures. To address this issue, we first introduce a simple multi-resolution framework that efficiently accounts for long-range dependency. Then, we show that additional statistical constraints further improve the reproduction of textures with strong regularity. This can be achieved by constraining both the Gram matrices of a neural network and the power spectrum of the image. Alternatively one may constrain only the autocorrelation of the features of the network and drop the Gram matrices constraints. In an experimental part, the proposed methods are then extensively tested and compared to alternative approaches, both in an unsupervised way and through a user study. Experiments show the interest of the multi-scale scheme for high resolution textures and the interest of combining it with additional constraints for regular textures.   
\end{abstract}


\section{Introduction}

Examplar-based texture synthesis consists in automatically generating sample images from a given example texture image. These samples are required to be visually faithful to the example and as diverse as possible. For more than forty years, and despite its inherent ill-posedness, this problem has been a fruitful way to test visually the validity of various mathematical models, ranging from time series~\cite{mccormick1974time}, Markov random fields~\cite{cross1983} to wavelet decompositions~\cite{heeger_pyramidbased_1995,portilla_parametric_2000} or non-parametric Markovian modeling~\cite{efros_texture_1999}. More recently, Convolutional Neural Networks have permitted impressive progresses in the field, initiated by the work by Gatys et al.~\cite{gatys_texture_2015}, itself followed by numerous contributions, e.g.~\cite{ulyanov2016,liu_texture_2016,snelgrove_highresolution_2017}.       

One challenge that has been faced by all methods since the early days of texture synthesis is the multi-scale nature of texture samples, implying that models should be able to reproduce both small and large scales, possibly over several orders of magnitude. For instance, parametric models for Markov fields are known to be intrinsically badly suited to a multi-scale modeling. Zooming such a model by a given factor implies extremely heavy computations to derive the corresponding parameters~\cite{gidas1989}, inpairing the design of multi-scale such models. 
Wavelet models are more adapted by nature to multi-scale modeling, but the faithful reproduction of structured textures requires complex interactions between scales to be accounted for. The best such modeling up to date is the second order statistical model proposed in~\cite{portilla_parametric_2000}, but highly structured textures still represent a challenge to such approaches. Non parametric Markov modeling methods such as those presented in~\cite{efros_texture_1999} or~\cite{efros_image_2001} indeed have the ability to deal simultaneously with several scales, albeit at a high computational cost. However, they are also well known to produce textures with very little variety, often producing verbatim copies, see~\cite{aguerrebere_exemplarbased_2013} and the experiments in the present paper. The methods relying on convolutional neural networks, following the seminal work by Gatys et al.~\cite{gatys_texture_2015}, are currently the most efficient to capture multi-scale structures. Nevertheless, they still lacks efficiency when large scale regularity is needed, as we will see in detail in this paper. Moreover, they are prone to generate artefacts that prevent a satisfactory reproduction of small scale structures. 

In this work, we present several neural synthesis methods that significantly improves the ability to preserve the large scale organisation of textures. We first propose a simple multi-resolution framework that account for large-scale structures and permits the synthesis of high resolution images. We then show that, in this multi-resolution framework, additional constraints are useful in the case of regular textures. A first approach combines the classical statistical constraints of neural approaches~\cite{gatys_texture_2015} (Gram matrices) with Fourier frequency constraints, similar to those introduced by~\cite{galerne2011}. A preliminary, mono-scale version of this idea was presented in a conference paper~\cite{liu_texture_2016}. Alternatively, the multi-resolution framework can be combined with a statistical constraint relying on the full auto-correlation of the features of the network. 
 This approach is closely related to the one introduced in~\cite{sendik_deep_2017}, which combines correlations with Gram matrices and various additional constraints. We show that correlation terms alone yield excellent results and therefore that Gram matrices are not necessary in this case.

We then evaluate the proposed methods in an extensive experimental section. The evaluation of texture synthesis results is a challenging task. Some approaches draw on well chosen statistics to estimate the quality of the results (the closest to the examplar, the better), as for instance discussed in~\cite{clarke2011perceptual}. In this paper, we first evaluate results in this manner, relying on Kullback-Leibler divergence between wavelet marginals, following the texture indexing scheme from~\cite{do_waveletbased_2002}. 
Then, we also evaluate the proposed methodology through a perceptual user study. Indeed, it is shown in~\cite{clarke2011perceptual,dong2013importance,dong_perceptual_2020} through extensive experiments that feature-based evaluations do not approach well human-based visual evaluation of texture similarity, especially in the case of long range correlations, which is precisely one of the cases tackled in this paper. We therefore rely on a user study to compare the framework we propose to both the original method from~\cite{gatys_neural_2015} and some of its improvement that focus on the respect of large scale structures~\cite{snelgrove_highresolution_2017,sendik_deep_2017}. 

\section{Neuronal texture synthesis}
\label{sec:SOA}

A complete state-of-the-art on the subject of texture synthesis is out of the scope of this paper. In view of the method that we propose in this work, we focus in this section on the works involving CNNs that have followed the seminal contribution of Gatys et al.~\cite{gatys_texture_2015} and particularly on works proposing new statistical constraints and focusing on long-range structure. 

\paragraph{Accelerations and alternative sampling strategy} In a first direction, several works have proposed ways to speed-up the synthesis process, notably through feed forward networks  \cite{ulyanov2016,ulyanov2017,johnson_perceptual_2016,shi2020fast}. In \cite{jetchev_texture_2016}, Generative Adversarial Networks are used to synthesize textures. Such methods enables fast synthesis once the networks have been trained for specific textures, but the quality of results is still inferior to the original approach~\cite{gatys_texture_2015}, especially for structured textures. Zhu and other authors have proposed an evolution of the FRAME model \cite{zhu1998} in the context of neural networks \cite{lu2015a} under the name \textit{DeepFrame}. Textures are synthesized from an exponential model using features from a neural network. In \cite{debortoli_macrocanonical_2019}, this macrocanonical approach is pushed further and fully analyzed theoretically. It is worth noting that both approaches \cite{lu2015a,debortoli_macrocanonical_2019} rely on first order constraints on features and therefore drop the use of the Gram matrices. 

\paragraph{Statistical constraints and losses} In a different direction, a large body of works has been dedicated to add additional constraints to the synthesis, often relying on new or modified loss functions. In~\cite{gatys_image_2016}, the color of the synthesis is constrained to specified values. In~\cite{risser_stable_2017}, it is proposed to constrain the histograms of some feature maps, in order to reduce halo artefacts. In~\cite{johnson_perceptual_2016}, a total variation term is added in the loss function for perceptual reasons. 
Other works such as~\cite{berger_incorporating_2016,liu_texture_2016,sendik_deep_2017} also propose alternative losses to add further statistical constraints. Since they explicitly deal with long range dependency and structure, they will be reviewed in the next paragraph. 
It should be noted that these approaches propose to combine several statistical constraints by adding them to get the final loss function. Another possibility would be to alternate different projections as it is done in the seminal work of Portilla and Simoncelli~\cite{portilla_parametric_2000}. 
Alternative constraints have also been investigated for the closely related task of style transfer. In ~\cite{li_demystifying_2017}, it is shown that matching Gram matrices reduces to kernel-based comparison of features, and various kernels are investigated in this setting. Other works investigate alternatives to the original Gram matrices, such as cross-layers (rather than within-layers) Gram matrices as in \cite{yeh_improved_2018} or \cite{novak_improving_2016} (both inspired by  \cite{portilla_parametric_2000}). 

\paragraph{Multi-scale neural synthesis} Neural networks such as the VGG19 used in most texture synthesis methods intrinsically have a multi-scale structure by alternating convolutions, non-linearity and subsampling. However, as we will see in the experimental section, the size of the receptive fields in these networks is not sufficient to synthesize large scale structures, especially when the resolution increases. To the best of our knowledge, only one paper, \cite{snelgrove_highresolution_2017}, proposes to rely on a multi-scale strategy to synthesize high resolution textures. The idea is to feed the network with a multi-scale decomposition, in this case a Gaussian pyramid, instead of a single image.  In this paper we will propose an alternative approach to the multi-scale neural synthesis and compare our results with~\cite{snelgrove_highresolution_2017}.

\paragraph{Incorporating long distance dependency}

In \cite{berger_incorporating_2016}, long distance patterns are handled by adding in the loss function a cross-correlation term, made of the correlation between features maps and a shifted version of it. Different sets of shifts are used depending on the layer, up to about a sixth of the image size. 
In \cite{novak_improving_2016}, in the context of style transfer, a similar idea is investigated using only one-pixel shifts.
In \cite{sendik_deep_2017}, the same idea is pushed further, by considering all cross-correlations at once in order to impose long-range structure for regular textures. Several other terms (smoothness, diversity) are added to the loss function. This approach, to which we will compare our results, indeed yields long-range structure, nevertheless at the price of relatively strong artefacts. 
Apart from these work dealing with cross-correlation of features and closely related to the present paper, \cite{liu_texture_2016} proposed to incorporate the power spectrum in the loss function, thereby enabling the respect of highly structured textures. In a related work, \cite{schreiber_texture_2016}, it is proposed to impose the spectrum constraint by using a windowed Fourier Transform, enabling non-stationnary behavior to be accounted for, at the cost of the inherent stationary nature of textures.

\section{Multi-scale spectral control for texture synthesis}

In this section, we detail our method to synthesize high quality texture images. After recalling in \cref{sec:gatys}, the classical approach from Gatys et al., we introduce a simple  multi-scale framework  in \cref{sec:multiscale}, before presenting in \cref{sec:spectrum} the spectral constraint we propose in order to both control artefacts and preserve long-range structures. Finally, we present in \cref{sec:autocorr}, the use of the autocorrelation of the feature maps as a potential alternative to the Gram matrices.  

\subsection{Reminder on the work from~\cite{gatys_texture_2015}}
\label{sec:gatys}

The seminal work~\cite{gatys_texture_2015} is based on the idea that a network trained for classification purpose, in this case a VGG network as introduced in~\cite{simonyan_very_2014}, can be repurposed for a synthesis task. Roughly speaking, the synthesis is achieved by backpropagation of texture-adapted statistical constraints  from the inner layers of the network up to pixels of the synthesized image. 

More precisely, the method works as follows. We consider a given convolutional neural network~\footnote{In this work, as in \cite{gatys_texture_2015}, we consider the VGG19 network~\cite{simonyan_very_2014} but other choices are possible, including networks with random weights~\cite{he2016a}.} consisting of $l$ layers. For a given color texture exemplar $I\in \mathbb{R}^{h \times w \times 3}$, where $h,w$ are the dimensions of the image, we write $f_l$ for the output (that is, the activations) of layer $l$. This output will be called a {\it feature map} from now on. Each feature map $f^l$ belongs to $\mathbb{R}^{h_l \times w_l \times m_l}$, where $w_l, h_l$ are the spatial dimension of layer $l$ and $m_l$ the number of channels of the feature. We further write $N=h\times w$ and $N_l=h_l\times w_l$ for the spatial dimensions of respectively the image and the feature maps. 

To synthesis a new texture, some statistics are imposed on a subset\footnote{For simplicity's sake, we will consider layers from $1$ to $L$ in the rest of this document but the user can choose non-consecutive layers in the network.} $\mathcal{S}$ of the layers of the CNN. The statistics considered in \cite{gatys_neural_2015} are strongly inspired by the work from Portilla and Simoncelli~\cite{portilla_parametric_2000} and rely on the so-called Gram matrices $G^{l}\in \mathbb{R}^{m_l\times m_l}$, defined for each couple $p, q\in \{1,\cdots,m_l\}$ as 

\begin{align}
\label{eq:featgram}
   G^{l}_{p,q} = \frac{1}{N_l^2} \sum_{i=1}^{N_l} {f}^l_p(i) \cdot {f}^l_q(i) = \frac{1}{N_l^2} \langle {f}^l_q , {f}^l_p \rangle,
\end{align}

where $f^l_p \in \mathbb{R}^{N_l} $, for 
$p \in \{1,\cdots,m_l\}$, is the vectorized $p^\text{th}$ channel of feature $l$.

To generate a new texture image $\tilde{I}$ on the basis of a reference one $I$, a gradient descent is used, starting from a white noise image, to find an image that matches the reference statistics. Usually the L-BFGS-B  \cite{zhu1997} second-order optimization method is chosen.

The corresponding loss function 
on the features is defined as :
\begin{align}
\mathcal{L}_{Gram} =  \sum_{l=1}^{L} \omega_{l}  \lVert G^l - \tilde{G}^{l} \rVert_{2}^2,
 \label{eq:lossfct}
\end{align}
with $\omega_l \in \mathbb{R}$ the weight of the layer $l$. 

The loss function \cref{eq:lossfct} can be seen as multi-objective cost functions agglomerated into a single-objective cost function. 
Although comparing different objectives is generally difficult, choosing identical weights, i. e. $\omega_l = 1 ~ \forall l \in[1,L]$, yields perceptually acceptable results.

A central questions of texture synthesis is to identify the best sets of statistics to incorporate in this loss function and possibly the irreducible set of those statistics (\cite{julesz1962}). Although the method from~\cite{gatys_texture_2015} yields synthesis results of unprecedented quality, a strong limitation is its inability to respect long range dependency, particularly when large scale structures have some regularity. This can be seen in first row of \Cref{fig:BubbleMarbel}. Neural networks such as VGG-19 have a multiscale structure, through alternating convolution and subsampling, that allow some large scale structures to be accounted for. Nevertheless, the size of the filters used in CNNs such as VGG-19, and therefore the size of the corresponding receptive fields, are small with respect to the size of the image especially when synthesizing high resolution images (here $1024 \times 1024$). As we have mentioned in the introduction, several works have addressed this limitation~\cite{liu_texture_2016, berger_incorporating_2016,novak_improving_2016,schreiber_texture_2016,sendik_deep_2017}, but, as we will see in the experimental section, none is fully satisfactory. 
In the following sections, we propose several improvement of the original neural texture synthesis method in order to address this limitation. 

\subsection{Multi-scale synthesis}
\label{sec:multiscale}

The first modification we introduce to the method from~\cite{gatys_texture_2015} is a straightforward multi-scale framework that will help preserving the large scale organisation of images. This strategy is relatively classical for texture synthesis methods and has been used in the past in different settings~\cite{kwatra2005texture,tartavel2015}. This approach is much simpler than the related method introduced in~\cite{snelgrove_highresolution_2017} and, as we will see in the experimental section, yields better results. 

The idea is simply to first synthesize a coarse resolution image, which is then upsampled and given as initialization for a synthesis at the next scale. This process is repeated $K$ times until the desired resolution is reached. As illustrated in ~\cref{fig:multiscalesyn}, we first build an image pyramid from the examplar image $I$, iteratively down-sampling it by factors $2^{1},2^{2},\cdots, 2^{K}$, resulting in images $I^{(1)},I^{(2)},\cdots,I^{(K)}$. A first synthesis result is obtained by using the smallest image as the examplar and white noise as initialization. 
Then, for step $k \in {K,K-1,\cdots, 1}$, we generate a new result using $I^{(k)}$ as the exemplar and the obtained synthesis result $\tilde{I}^{(k-1)}$ as the initialization instead of white noise. The upsampling of $\tilde{I}^{(k-1)}$ is performed using bilinear interpolation. The only parameter of this generic multi-scale framework is the number of scales $K$. 

As can be seen in \Cref{fig:multiscalesyn_dif_K}, this strategy can yield strong improvements in some cases but is not enough to allow the reproduction of highly structured textures. In the next section, we show how the result can be improved by adding a careful control of the Fourier spectrum into the multi-scale scheme.

\tikzstyle{scaleRect} = [rectangle, rounded corners, minimum width=1.5cm, minimum height=1cm,text centered, draw=black]
\tikzstyle{CNN} = [ellipse, rounded corners, minimum width=2cm, minimum height=1cm,text centered, draw=black]
\tikzstyle{arrow} = [thick,->,>=stealth]
\tikzstyle{arrowdashed} = [dashed,thick,->,>=stealth]
\tikzstyle{textonly} = [minimum width=2cm, minimum height=1cm, text centered]

\begin{figure}[!ht]
\resizebox{\columnwidth}{!}{
\begin{tikzpicture}[node distance=2.2cm]

\node (scale0) [scaleRect] {Scale 0};
\node (scale1) [scaleRect, below of=scale0, yshift=-0cm] {Scale 1};
\node (scaleK) [scaleRect, below of=scale1, yshift=-0.5cm] {Scale K};

\node (CNN0) [CNN, right of=scale0, xshift=3cm] {CNN Synthesis};
\node (CNN0invisible) [textonly, below of=CNN0, yshift=1cm] {};
\node (CNN1) [CNN, right of=scale1, xshift=3cm] {CNN Synthesis};
\node (CNN1invisible) [textonly, below of=CNN1, yshift=1cm] {};
\node (CNNK) [CNN, right of=scaleK, xshift=3cm] {CNN Synthesis};

\node (syn_scale0) [scaleRect, right of=CNN0, xshift=3cm] {Synthesis 0};
\node (syn_scale1) [scaleRect, right of=CNN1, xshift=3cm] {Synthesis 1};
\node (syn_scaleK) [scaleRect, right of=CNNK, xshift=3cm] {Synthesis K};

\node (inputnoise) [textonly,below of=CNNK] {White Noise};
\node (finaloutput) [textonly,above of=syn_scale0,yshift=-1cm] {Final result};
\node (firstinput) [textonly,above of=scale0,yshift=-1cm] {Reference image};

\node (Reference0) [left of=scale0,xshift=-1cm] {\includegraphics[width=0.2\linewidth]{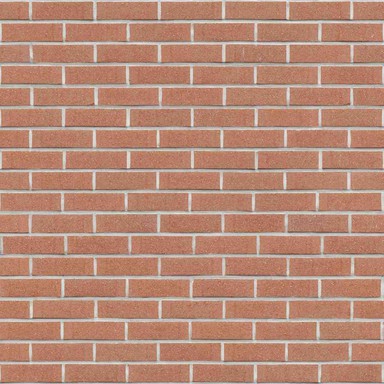}};
\node (Reference1) [left of=scale1,xshift=-1cm] {\includegraphics[width=0.1\linewidth]{im/MS_fig/TexturesCom_BrickSmallNew0099_1_seamless_S_1024}};
\node (ReferenceK) [left of=scaleK,xshift=-1cm] {\includegraphics[width=0.05\linewidth]{im/MS_fig/TexturesCom_BrickSmallNew0099_1_seamless_S_1024}};
\node (SynImK) [right of=syn_scaleK,xshift=1cm]	{\includegraphics[width=0.05\linewidth]{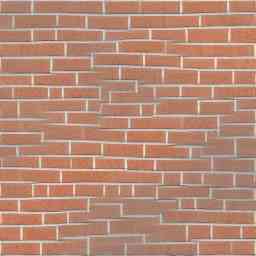}};
\node (SynIm1) [right of=syn_scale1,xshift=1cm] 	{\includegraphics[width=0.1\linewidth]{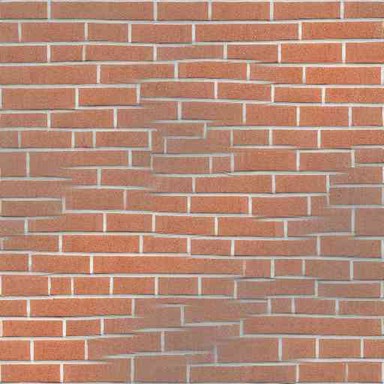}};
\node (SynIm0) [right of=syn_scale0,xshift=1cm] 
{\includegraphics[width=0.2\linewidth]{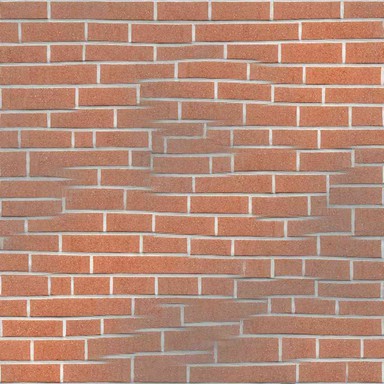}};

\draw [arrow] (scale0) -- node[anchor=west] {Down-sampling} (scale1);
\draw [arrowdashed] (scale1) -- node[anchor=west] {Down-sampling} (scaleK);
\draw [arrow] (syn_scale1)  |- (CNN0invisible) node[anchor=north] {Up-sampling} -|  (CNN0);
\draw [arrow] (scale0) --  (CNN0);
\draw [arrow] (CNN0) --   (syn_scale0);
\draw [arrow] (scale1) -- (CNN1);
\draw [arrow] (scaleK) -- node[anchor=south] {Reference} (CNNK);
\draw [arrow] (CNN1) --  (syn_scale1);
\draw [arrow] (CNNK) --  node[anchor=south] {Output} (syn_scaleK);
\draw [arrow] (inputnoise) -- node[anchor=west] {Input - Initialization} (CNNK);
\draw [arrowdashed] (syn_scaleK) |- (CNN1invisible) -| node[anchor=north] {Up-sampling} (CNN1);

\end{tikzpicture}

}

\caption{Multiscale strategy. The examplar is down-sampled by factors $2^{-1},2^{-2},\cdots, 2^{-k}$ to build a pyramid $I^{(1)},I^{(2)},\cdots,I^{(k)}$. At scale $K$, a new texture is synthesized by using $I^{(k)}$ as the exemplar and the upsampled result of the synthesis at scale $K-1$ as initialization (instead of white noise). We repeat this step until we reach the top of the image pyramid.}
	\label{fig:multiscalesyn}
\end{figure}

\subsection{Spectrum Constraint}
\label{sec:spectrum}

We propose to include in the synthesis a new constraint based on the Fourier spectrum of the image. It is known that such a constraint alone is an efficient way to reproduce the so-called {\it micro-textures}~\cite{galerne2011} made of uniformly distributed small details. This constraint has also been used in combination with more structured synthesis methods in~\cite{tartavel2015}. 

Let us write $\mathcal{F}(I)$ for the Discrete Fourier Transform (DFT) of an image $I$, $\mathcal{F}^{-1}$ for the inverse DFT and $|.|$ for the complex modulus. The idea is to constrain the synthesized image $\tilde{I}$ to have a Fourier spectrum $|\mathcal{F}(\tilde{I})|$ as similar as possible to  $|\mathcal{F}(I)|$, the spectrum of $I$. A simple way to do this is to first perform the multi-scale neural synthesis described above, and then to replace the phases of the Fourier transform of the synthesized image with random phases, before applying the inverse Fourier transform to the result~\cite{galerne2011}. Now, this sequential strategy is not satisfactory, since the randomization of phases would destroy the effect of both the statistical constraint on the VGG features and the effect of the multi-scale strategy. Therefore, we propose to incorporate the Fourier constraint into the multi-scale synthesis process. A preliminary, mono-scale version of this idea was presented in~\cite{liu_texture_2016}. 

In order to include the Fourier constraint into the loss function used for synthesizing images, we first introduce $\mathcal{E}_I$, the set of images having the same spectrum as $I$ the examplar image. In the case of color images, this is defined as 

$$\mathcal{E}_I=\left\{ J \in \mathbb{R}^{h \times w \times 3} | \exists \phi \in \mathbb{R}^{h \times w} : \mathcal{F}(J) = e^{i\phi}\mathcal{F}(I) \right\}.$$

Next, we define the Fourier loss associated to the image $\tilde{I}$ as the normalized Euclidean distance  between $\tilde{I}$ and $\mathcal{E}_I$,

\begin{align}
\label{eq:lossspe}
  \mathcal{L}_{spe} &= \frac{1}{2N} d(\tilde{I},\mathcal{E}_I)^2 = \frac{1}{2N}\| \tilde{I}-\mathcal{P}(\tilde{I})\|^2, 
\end{align}
 and the total loss as 
 
 \begin{align}
\label{eq:lossall}
\mathcal{L}        = \mathcal{L}_{Gram} + \beta \mathcal{L}_{spe},
\end{align}
where $\beta$ is a weighting parameter. Since the Fourier loss is the distance to $\mathcal{E}_I$, its gradient is given by 
$$\Delta_{spe} =  N^{-1} (\tilde{I} - \mathcal{P}(\tilde{I})),$$
where $\mathcal{P}$ is the projection operator on $\mathcal{E}_I$. This projection is given by (see~\cite{tartavel2015}, Appendix A)
\begin{align}
\label{eq:transfer_phase}
\mathcal{P}(\tilde{I_c}) = \mathcal{F}^{-1}\left( \frac{\mathcal{F}(\tilde{I})\cdot\mathcal{F}(I)}{| \mathcal{F}(\tilde{I})\cdot\mathcal{F}(I) |}\cdot \mathcal{F}(I_c)  \right), c\in\{r,g,b\}
\end{align}
where $\cdot$ is the scalar product in $\mathbb{C}^3$, that is 
$$\mathcal{F}(\tilde{I})\cdot\mathcal{F}(I)=\sum_{c=r,g,b}\mathcal{F}(\tilde{I_c})\mathcal{F}(I_c)^*,$$
$I_c$, for $c=r,g,b$, being the color channels of $I$ and $a^*$ the conjugate of complex number $a$.
This spectrum constraint can be seen as a regularization to the Ill-posed example based synthesis problem.

\subsection{Autocorrelation of the feature maps}
\label{sec:autocorr}

In this section, we consider an alternative way to impose  long-range consistency, based on the autocorrelation of the features maps. This is motivated by the fact that the autocorrelation is a proxy of repeating patterns, such as the presence of periodic elements in the signal. As explained in \cref{sec:SOA}, this idea has been explored with different modality in~\cite{berger_incorporating_2016,novak_improving_2016,sendik_deep_2017}. 

The autocorrelation function of an image is defined as the convolution of the image with itself. Let   $I \in \mathbb{R}^{h \times w}$, the autocorrelation $C(I)  =  \in \mathbb{R}^{h \times w}$ is defined,  for $ \forall k \in ~ \{1,\cdots,h\}$ and $\forall l \in ~  \{1,\cdots,w\}$, as 
\begin{align}
C(I)(k,l) & = \frac{1}{N^2} \sum_{i=1}^{h}  \sum_{j=1}^{w} I(i,j)I(\mid i+k \mid_h,\mid j+l\mid_w) \\
& =  \frac{1}{N^2} I * I 
 \label{eq:autocorr}
\end{align}
 $\mid \bullet \mid_h$ being the modulo operation with divisor h. 

And efficient way to compute the autocorrelation is to use the Discrete Fourier Transform (DFT).  According to the Wiener-Khintchin theorem we have :
$$C(I)  = \mathcal{F}^{-1} ( \mid \mathcal{F}(I) \mid^2 ). $$
Then, we define the autocorrelation constraint at the layer $l$ as  
$A^l \in \mathbb{R}^{h_l\times w_l \times m_l}$ the tensor of the squared modulus of the Fourier transform of the features maps, ie :
\begin{align}
\label{eq:f_autocorr}
    A^l_{p} = \frac{1}{N_l^2} \mid \mathcal{F}(f^l_p) \mid^2
\end{align}
with $p \in \{1,\cdots,m_l\}$ the corresponding indexes of the feature map $p$.
Using this toric representation allows one to consider all possible shifts between pixels. %

This constraint is similar to the one in \cite{sendik_deep_2017}, except that it is dealt with in the Fourier domain and there is no  weighting of the elements of the autocorrelation matrix.

\FloatBarrier %
\section{Experiments}
\label{sec:Experiments}

In this section, we perform experiments to illustrate both the multi-scale framework and the additional constraints we propose for neural texture synthesis. After briefly introducing the methods we compare ourselves to, we first show some visual results. Then, we propose a method to evaluate the innovation capacity of algorithms, and more precisely their tendency to produce verbatim copy of the input. Further, we evaluate the methods quantitatively using the Kullback-Leibler divergence between wavelet statistics. Despite the interest of such quantitative evaluations, it is known that they have severe limitations, in particular to evaluate results at large scales \cite{dong_perceptual_2020}. Therefore, we also have conducted a medium scale perceptual evaluation from human observers, the results of which we analyze in Section~\ref{sec:perceptualEval}. These different evaluations have been conducted on the 20 texture images visible in ~\cref{fig:PerceptualTest_RefImages}. These high resolution ($1024\times 1024$) textures have been chosen to include both structured and irregular textures. Some of them display strong long-range dependency. All results can be found in Supplementary Materials. Eventually, we study the effects of various parameters and briefly illustrate the ability of our method to produce higher resolution textures.  %

\subsection{Architecture and parameters} We use a VGG-19 network pre-trained on ImageNet with rescaled weights\footnote{The rescaled VGG-19 network can be found at \url{http://github.com/leongatys/DeepTextures}} as in \cite{gatys_texture_2015} and we also use the same layers i.e. : 'Conv1\_1', 'Pooling1', 'Pooling2', 'Pooling3', 'Pooling4'. The corresponding weights\footnote{Due to the numerical sensitivity of the LBFGS-B optimization algorithm.} are set to be $w_1=w_2=w_3=w_4=w_5 =10^9$. When the spectrum constraint is considered, we use a weighting parameter $\beta = 10^5$ unless otherwise specified. Synthesis are performed using  2000 iterations. We use Tensorflow as a deep learning framework and  Scipy as an optimization package.   
Synthesizing one texture of size $1024 \times 1024$ takes 60 minutes with a GeForce 1080 Ti for the method multi-scale "Gram + MSInit". The overhead compared to Gatys \cite{gatys_texture_2015} at the same scale is limited because the synthesis at lower resolutions are faster.

\subsection{Other texture synthesis methods} The first method we compare ourselves to is the original synthesis method from Gatys et al.~\cite{gatys_texture_2015}, that from now we refer to as "Gatys".
We also consider the method "Deep Corr", introduced in \cite{sendik_deep_2017}, using  the code from the authors\footnote{The code of \cite{sendik_deep_2017} can be found on Github : \url{https://github.com/omrysendik/DCor}}, using a maximum of 2000 iterations.
We also consider the multi-scale texture algorithm from \cite{snelgrove_highresolution_2017}, using the code from the author \footnote{The code of \cite{snelgrove_highresolution_2017} \url{https://github.com/wxs/subjective-functions}}, using layers 3 and 8 and 5 octaves in the Gaussian pyramid as in the original paper. We use a maximum of 2000 iterations. From now on, we refer to this method as "Snelgrove". Those last two methods have been chosen because they explicitly address the problem of reproducing large scale structures. 
We also consider the Feed Forward approach proposed in  \cite{ulyanov2016} using  a PyTorch  implementation by Jorge Gutierrez\footnote{\url{https://github.com/JorgeGtz/TextureNets_implementation}}. We refer to this method as "Ulyanov". Finally, we consider two patch-based methods, from the works  \cite{efros_texture_1999} and \cite{efros_image_2001}, using implementations from the online journal IPOL \cite{raad_efros_2017,aguerrebere_exemplarbased_2013}, with default parameters settings. We refer to these two methods respectively as "Efros Leung" and "Efros Freeman".

\subsection{Visual comparisons} 

\begin{figure}[!ht]
\centering
\begin{minipage}[b]{0.225\linewidth}
\centering Reference \\
\includegraphics[width=\linewidth]{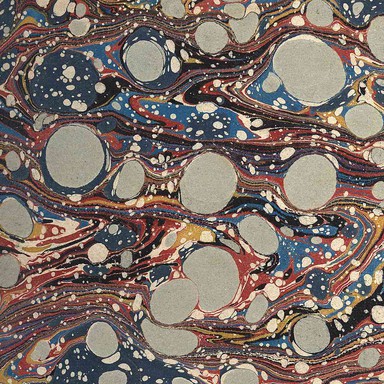}
\end{minipage}
\begin{minipage}[b]{0.225\linewidth}
\centering Gatys \cite{gatys_texture_2015} \\
\includegraphics[width=\linewidth]{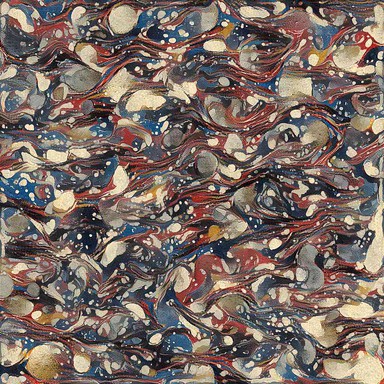}
\end{minipage}
\begin{minipage}[b]{0.225\linewidth}
\centering Efros Leung \cite{efros_texture_1999} \\
\includegraphics[width=\linewidth]{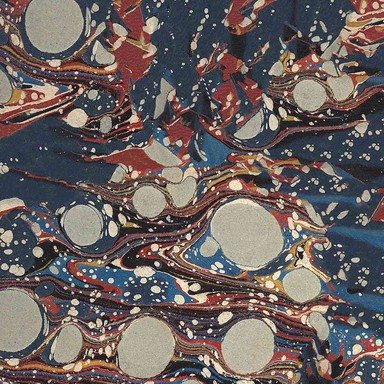}
\end{minipage}
\begin{minipage}[b]{0.225\linewidth}
\centering Efros Freeman \cite{efros_image_2001} \\
\includegraphics[width=\linewidth]{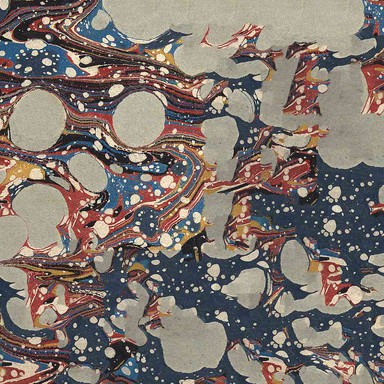}
\end{minipage}
\begin{minipage}[b]{0.225\linewidth}
\centering Ulyanov  \cite{ulyanov2016} \\
\includegraphics[width=\linewidth]{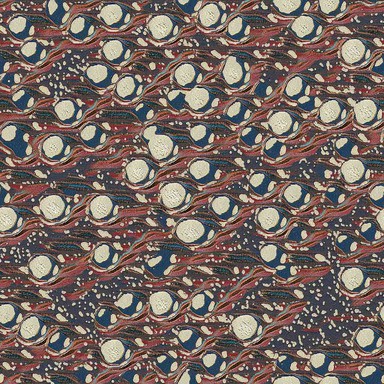}
\end{minipage}
\begin{minipage}[b]{0.225\linewidth}
\centering Gram + Spectrum \\
\includegraphics[width=\linewidth]{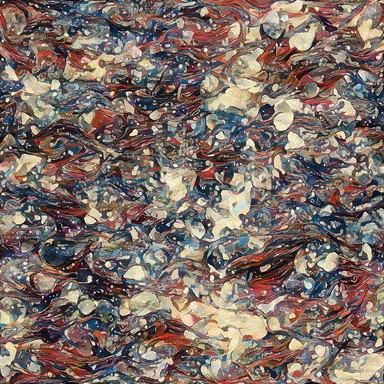}
\end{minipage}
\begin{minipage}[b]{0.225\linewidth}
\centering Deep Corr \cite{sendik_deep_2017} \\
\includegraphics[width=\linewidth]{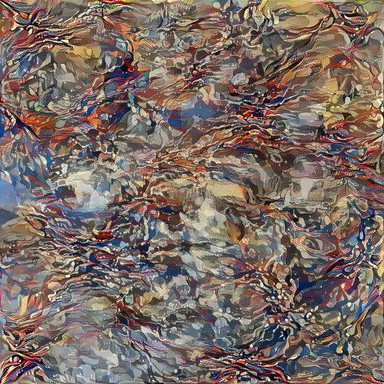}
\end{minipage}
\begin{minipage}[b]{0.225\linewidth}
\centering Autocorr \\
\includegraphics[width=\linewidth]{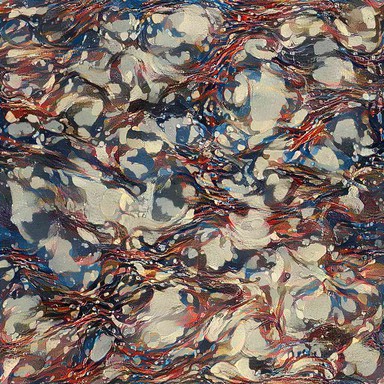}
\end{minipage}
\begin{minipage}[b]{0.225\linewidth}
\centering Snelgrove \cite{snelgrove_highresolution_2017} \\
\includegraphics[width=\linewidth]{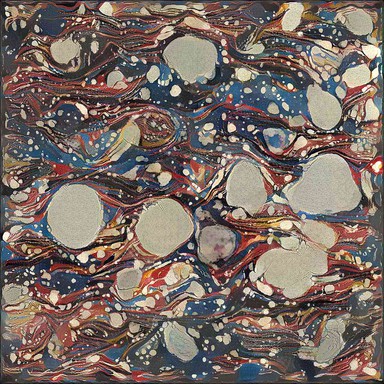}
\end{minipage}
\begin{minipage}[b]{0.225\linewidth}
\centering Gram + MSInit \\
\includegraphics[width=\linewidth]{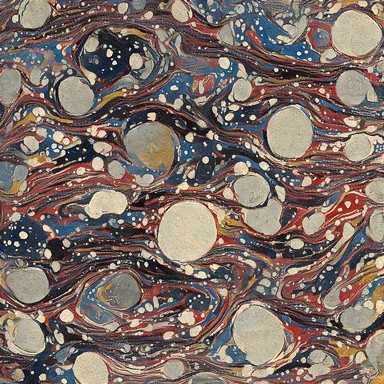}
\end{minipage}
\begin{minipage}[b]{0.225\linewidth}
\centering Gram + Spectrum + MSInit \\
\includegraphics[width=\linewidth]{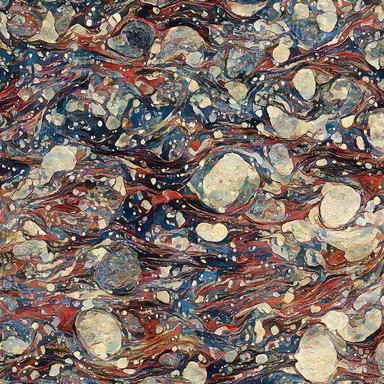}
\end{minipage}
\begin{minipage}[b]{0.225\linewidth}
\centering Autocorr + MSInit \\
\includegraphics[width=\linewidth]{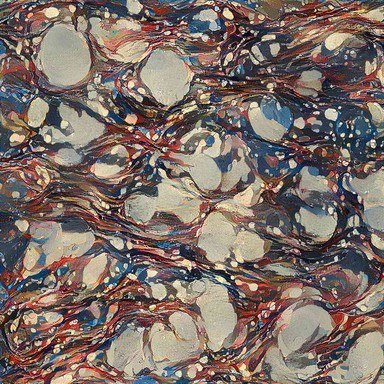}
\end{minipage}
\caption{Synthesis results using different methods for a given reference of size $1048 \times 1048$.}
\label{fig:BubbleMarbel}
\end{figure}

\begin{figure}[!ht]
\centering
\begin{minipage}[b]{0.225\linewidth}
\centering Reference \\
\includegraphics[width=\linewidth]{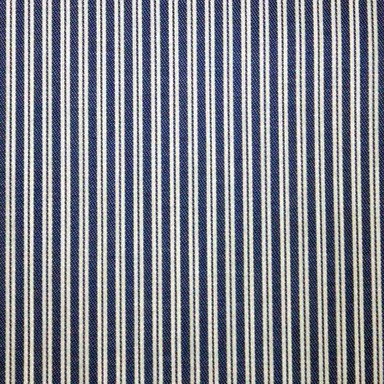}
\end{minipage}
\begin{minipage}[b]{0.225\linewidth}
\centering Gatys \cite{gatys_texture_2015} \\
\includegraphics[width=\linewidth]{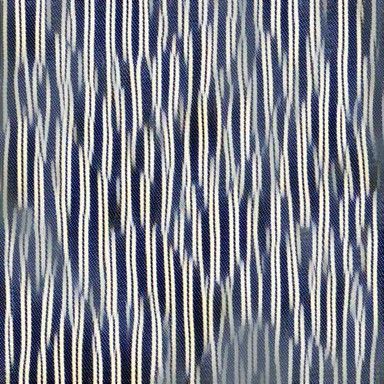}
\end{minipage}
\begin{minipage}[b]{0.225\linewidth}
\centering Efros Leung \cite{efros_texture_1999} \\
\includegraphics[width=\linewidth]{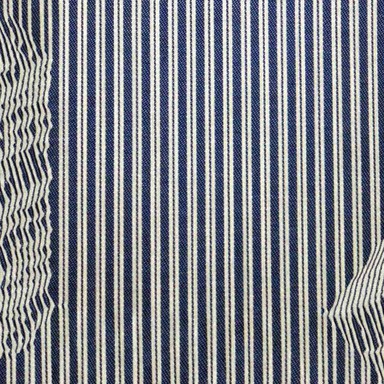}
\end{minipage}
\begin{minipage}[b]{0.225\linewidth}
\centering Efros Freeman \cite{efros_image_2001} \\
\includegraphics[width=\linewidth]{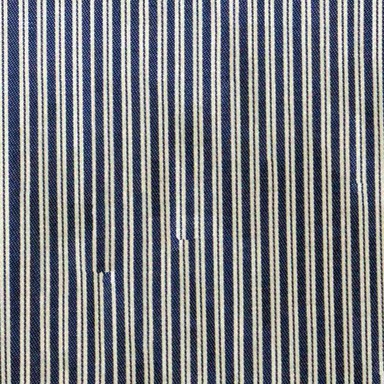}
\end{minipage}
\begin{minipage}[b]{0.225\linewidth}
\centering Ulyanov  \cite{ulyanov2016} \\
\includegraphics[width=\linewidth]{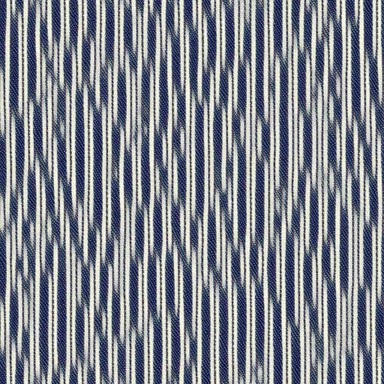}
\end{minipage}
\begin{minipage}[b]{0.225\linewidth}
\centering Gram + Spectrum \\
\includegraphics[width=\linewidth]{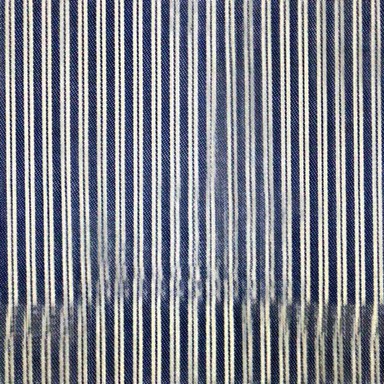}
\end{minipage}
\begin{minipage}[b]{0.225\linewidth}
\centering Deep Corr \cite{sendik_deep_2017} \\
\includegraphics[width=\linewidth]{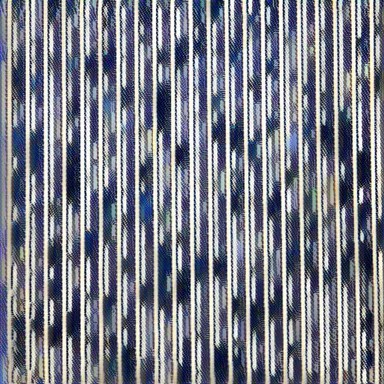}
\end{minipage}
\begin{minipage}[b]{0.225\linewidth}
\centering Autocorr \\
\includegraphics[width=\linewidth]{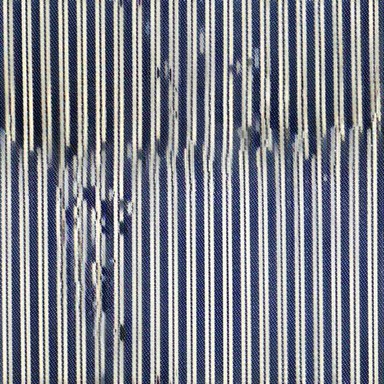}
\end{minipage}
\begin{minipage}[b]{0.225\linewidth}
\centering Snelgrove \cite{snelgrove_highresolution_2017} \\
\includegraphics[width=\linewidth]{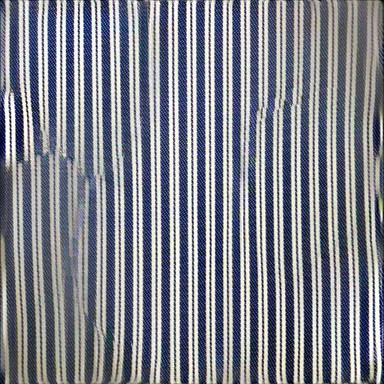}
\end{minipage}
\begin{minipage}[b]{0.225\linewidth}
\centering Gram + MSInit \\
\includegraphics[width=\linewidth]{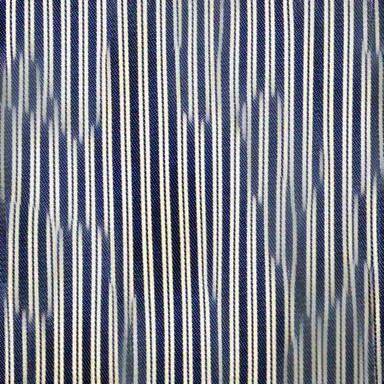}
\end{minipage}
\begin{minipage}[b]{0.225\linewidth}
\centering Gram + Spectrum + MSInit \\
\includegraphics[width=\linewidth]{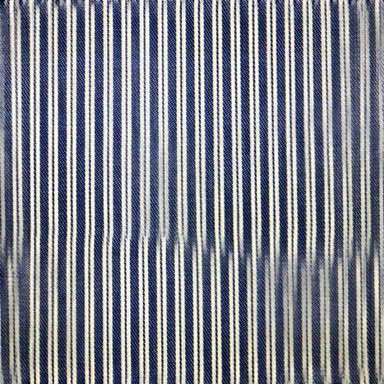}
\end{minipage}
\begin{minipage}[b]{0.225\linewidth}
\centering Autocorr + MSInit \\
\includegraphics[width=\linewidth]{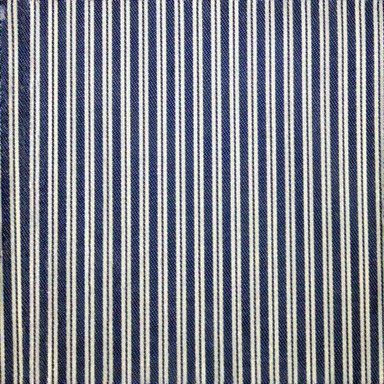}
\end{minipage}
\caption{Synthesis results using different methods for a given reference of size $1048 \times 1048$.}
\label{fig:fabric_white_blue_1024}
\end{figure}

\begin{figure}[!ht]
\centering
\begin{minipage}[b]{0.225\linewidth}
\centering Reference \\
\includegraphics[width=\linewidth]{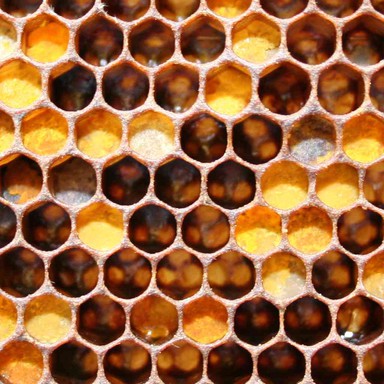}
\end{minipage}
\begin{minipage}[b]{0.225\linewidth}
\centering Gatys \cite{gatys_texture_2015} \\
\includegraphics[width=\linewidth]{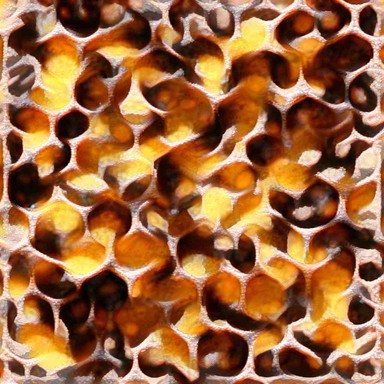}
\end{minipage}
\begin{minipage}[b]{0.225\linewidth}
\centering Efros Leung \cite{efros_texture_1999} \\
\includegraphics[width=\linewidth]{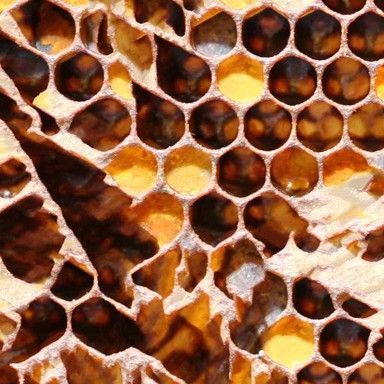}
\end{minipage}
\begin{minipage}[b]{0.225\linewidth}
\centering Efros Freeman \cite{efros_image_2001} \\
\includegraphics[width=\linewidth]{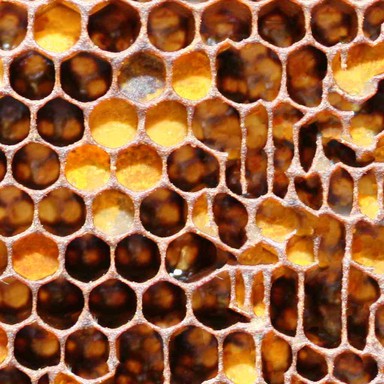}
\end{minipage}
\begin{minipage}[b]{0.225\linewidth}
\centering Ulyanov  \cite{ulyanov2016} \\
\includegraphics[width=\linewidth]{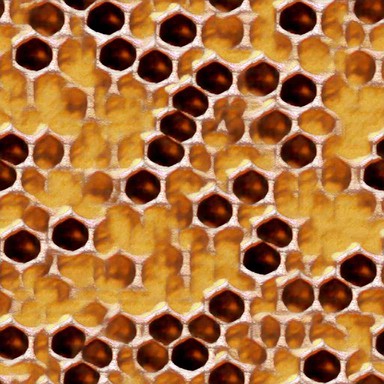}
\end{minipage}
\begin{minipage}[b]{0.225\linewidth}
\centering Gram + Spectrum \\
\includegraphics[width=\linewidth]{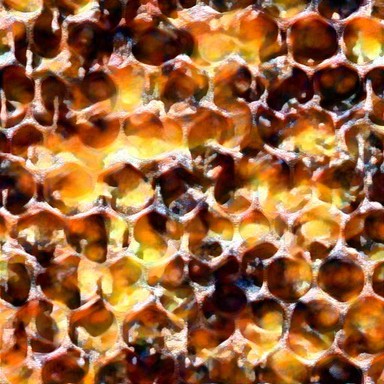}
\end{minipage}
\begin{minipage}[b]{0.225\linewidth}
\centering Deep Corr \cite{sendik_deep_2017} \\
\includegraphics[width=\linewidth]{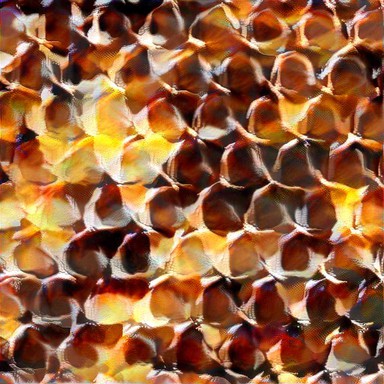}
\end{minipage}
\begin{minipage}[b]{0.225\linewidth}
\centering Autocorr \\
\includegraphics[width=\linewidth]{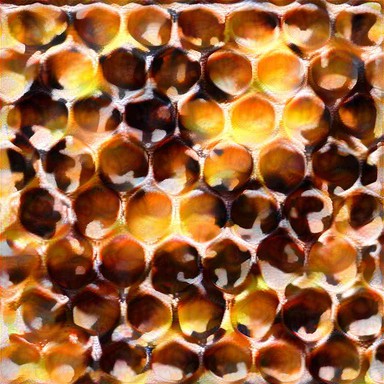}
\end{minipage}
\begin{minipage}[b]{0.225\linewidth}
\centering Snelgrove \cite{snelgrove_highresolution_2017} \\
\includegraphics[width=\linewidth]{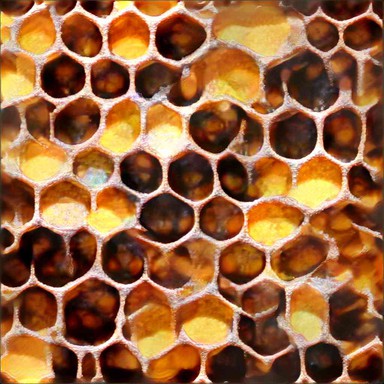}
\end{minipage}
\begin{minipage}[b]{0.225\linewidth}
\centering Gram + MSInit \\
\includegraphics[width=\linewidth]{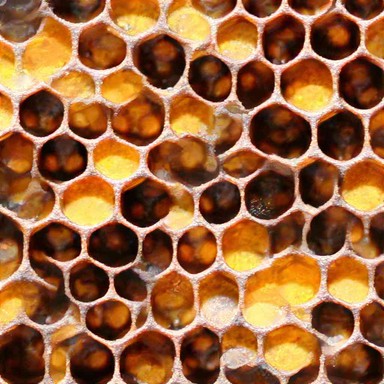}
\end{minipage}
\begin{minipage}[b]{0.225\linewidth}
\centering Gram + Spectrum + MSInit \\
\includegraphics[width=\linewidth]{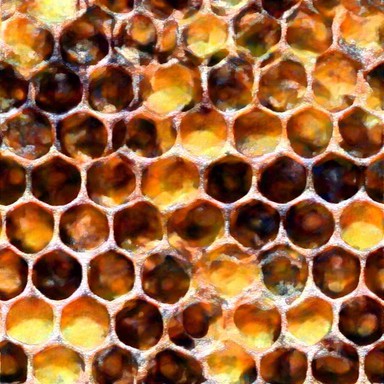}
\end{minipage}
\begin{minipage}[b]{0.225\linewidth}
\centering Autocorr + MSInit \\
\includegraphics[width=\linewidth]{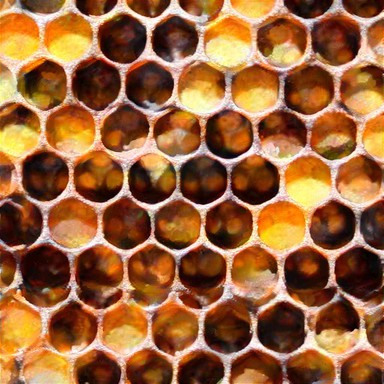}
\end{minipage}
\caption{Synthesis results using different methods for a given reference of size $1048 \times 1048$.}
\label{fig:Pierzga_2006_1024}
\end{figure}

\begin{figure}[!ht]
\centering
\begin{minipage}[b]{0.225\linewidth}
\centering Reference \\
\includegraphics[width=\linewidth]{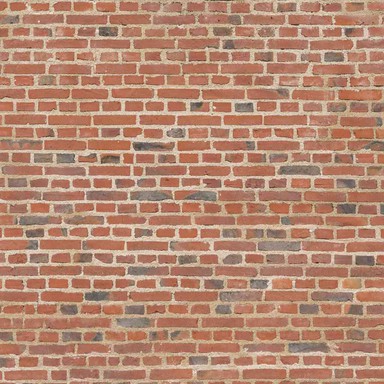}
\end{minipage}
\begin{minipage}[b]{0.225\linewidth}
\centering Gatys \cite{gatys_texture_2015} \\
\includegraphics[width=\linewidth]{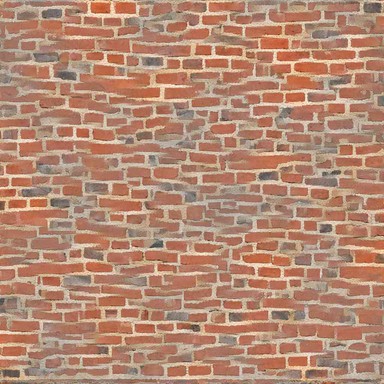}
\end{minipage}
\begin{minipage}[b]{0.225\linewidth}
\centering Efros Leung \cite{efros_texture_1999} \\
\includegraphics[width=\linewidth]{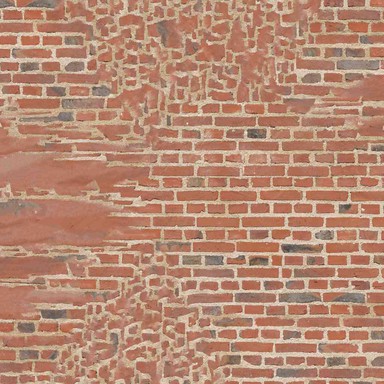}
\end{minipage}
\begin{minipage}[b]{0.225\linewidth}
\centering Efros Freeman \cite{efros_image_2001} \\
\includegraphics[width=\linewidth]{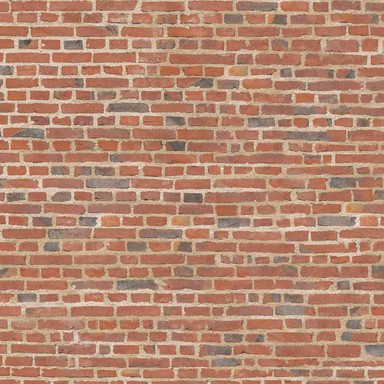}
\end{minipage}
\begin{minipage}[b]{0.225\linewidth}
\centering Ulyanov  \cite{ulyanov2016} \\
\includegraphics[width=\linewidth]{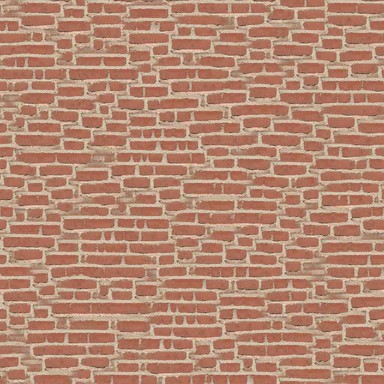}
\end{minipage}
\begin{minipage}[b]{0.225\linewidth}
\centering Gram + Spectrum \\
\includegraphics[width=\linewidth]{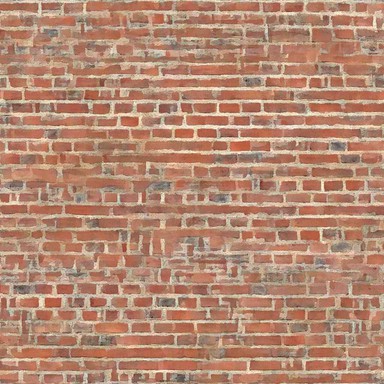}
\end{minipage}
\begin{minipage}[b]{0.225\linewidth}
\centering Deep Corr \cite{sendik_deep_2017} \\
\includegraphics[width=\linewidth]{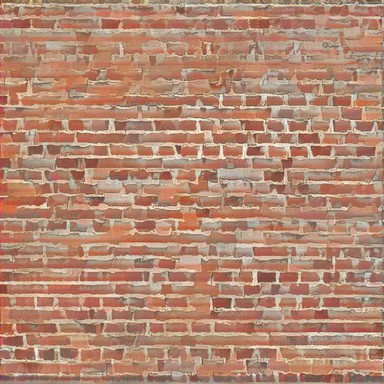}
\end{minipage}
\begin{minipage}[b]{0.225\linewidth}
\centering Autocorr \\
\includegraphics[width=\linewidth]{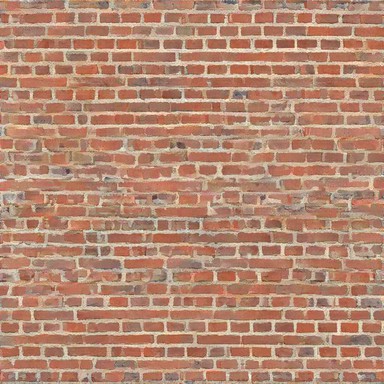}
\end{minipage}
\begin{minipage}[b]{0.225\linewidth}
\centering Snelgrove \cite{snelgrove_highresolution_2017} \\
\includegraphics[width=\linewidth]{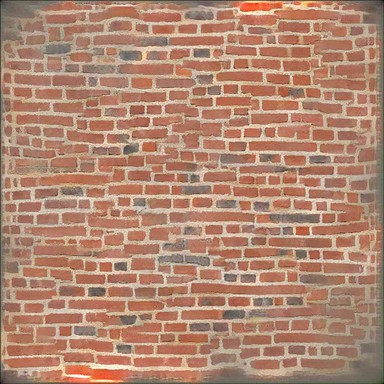}
\end{minipage}
\begin{minipage}[b]{0.225\linewidth}
\centering Gram + MSInit \\
\includegraphics[width=\linewidth]{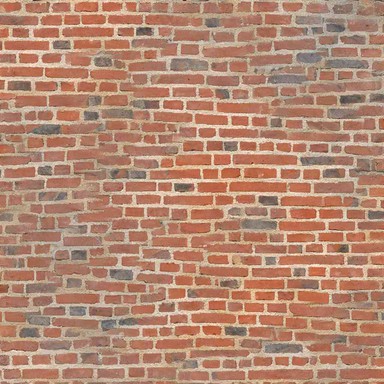}
\end{minipage}
\begin{minipage}[b]{0.225\linewidth}
\centering Gram + Spectrum + MSInit \\
\includegraphics[width=\linewidth]{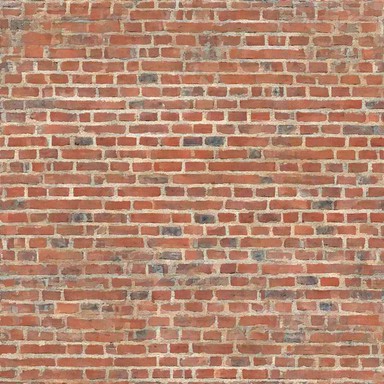}
\end{minipage}
\begin{minipage}[b]{0.225\linewidth}
\centering Autocorr + MSInit \\
\includegraphics[width=\linewidth]{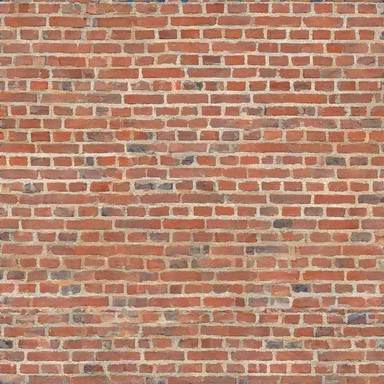}
\end{minipage}
\caption{Synthesis results using different methods for a given reference of size $1048 \times 1048$.}
\label{fig:TexturesCom_BrickSmallBrown0473_1_M_1024}
\end{figure}

In \Cref{fig:BubbleMarbel,fig:fabric_white_blue_1024,fig:Pierzga_2006_1024,fig:TexturesCom_BrickSmallBrown0473_1_M_1024} we display synthesis results using our methods and those presented in the previous paragraph. For space reason, we only consider 4 textures, all exhibiting some kind of long-range dependency. Their resolution is $1024 \times 1024$. Some details can be seen on \cref{fig:ZoomDetails}. All results can be seen in Supplementary Materials.%

We first notice that patch-based methods are very faithful to the reference image. However, they have the tendency to produce regions that are exact copy of the input, a verbatim effect already noticed in \cite{aguerrebere_exemplarbased_2013} and investigated in the next section. They also at times yield images with constant or repetitive patterns.%

Among neural methods, the original "Gatys" method is still competitive, but struggles to reproduce large scales on these high resolution textures. This is due to the size of the receptive fields, which is clearly not sufficient in this case. The method from "Ulyanov" is worse in this respect. The method "Deep Corr" improves the preservation of large scale structures, but results are not satisfying, some structures are lacking and artefacts are visible. In contrast, the plain use of the auto-correlation term as an additional constraint, as we propose in "Autocorr", yields better results, even though no use of the Gram matrices is made. The regularization and innovation terms present in the method from~\cite{sendik_deep_2017} may also be harmful in these cases. Next, we observe that adding the Fourier spectrum constraint alone (at a single scale) yields interesting results, but is not enough to get fully satisfying results. The multi-scale methods, be it "Snelgrove" or the one we propose, "Gram+MSInit", "Gram+Spectrum+MSInit", "Autocorr+MSInit", all improve the original synthesis method "Gatys". In the case of very regular textures, as in \Cref{fig:fabric_white_blue_1024,fig:Pierzga_2006_1024,fig:TexturesCom_BrickSmallBrown0473_1_M_1024}, our multi-scale methods "XXX+MSInit" yields better results, as will be confirmed by the user study in \cref{sec:perceptualEval}. The method "Autocorr+MSInit" sometimes yields results that are clearly better than others, especially for very structured textures, as can be seen in \cref{fig:Pierzga_2006_1024} or on the last line of \cref{fig:ZoomDetails}. Nevertheless, it also sometimes fails as in \cref{fig:BubbleMarbel} and may produce artefacts on some examples. For this reason and for human resources constraints, we choose, among our methods, to only include "Gram+MSInit" and "Gram+Spectrum+MSInit" in the user study presented in Section~\ref{sec:perceptualEval}.


\begin{figure}[!ht]
\centering
\begin{minipage}[c]{0.24\linewidth}
\centering Snelgrove \cite{snelgrove_highresolution_2017} \\
\end{minipage}
\begin{minipage}[c]{0.24\linewidth}
\centering Gram + MSInit \\
\end{minipage}
\begin{minipage}[c]{0.24\linewidth}
\centering Gram + Spectrum + MSInit \\
\end{minipage}
\begin{minipage}[c]{0.24\linewidth}
\centering Autocorr + MSInit \\
\end{minipage}
\begin{minipage}[c]{0.24\linewidth}
\includegraphics[width=\linewidth]{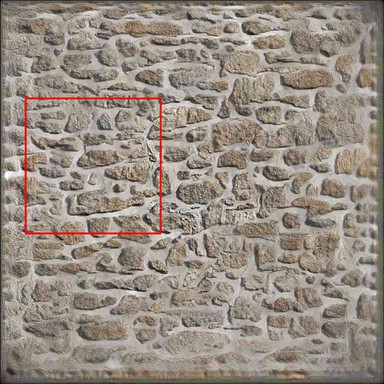}
\end{minipage}
\begin{minipage}[c]{0.24\linewidth}
\includegraphics[width=\linewidth]{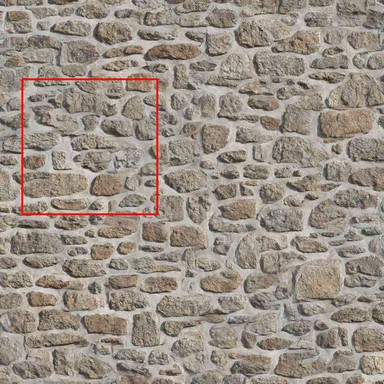}
\end{minipage}
\begin{minipage}[c]{0.24\linewidth}
\includegraphics[width=\linewidth]{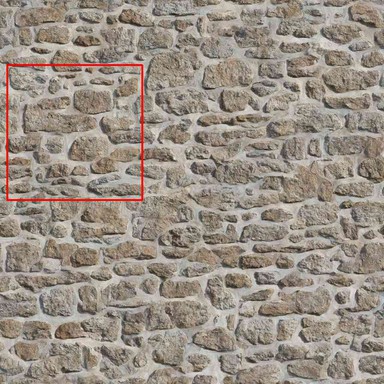}
\end{minipage}
\begin{minipage}[c]{0.24\linewidth}
\includegraphics[width=\linewidth]{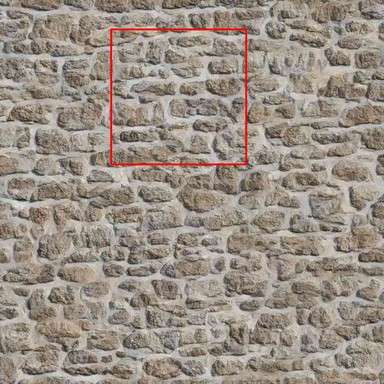}
\end{minipage}
\begin{minipage}[c]{0.24\linewidth}
\includegraphics[width=\linewidth]{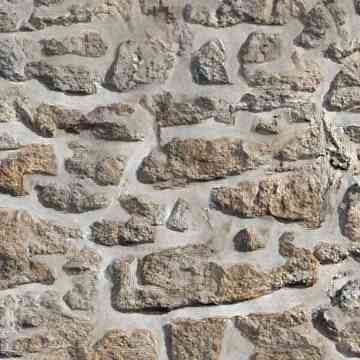}
\end{minipage}
\begin{minipage}[c]{0.24\linewidth}
\includegraphics[width=\linewidth]{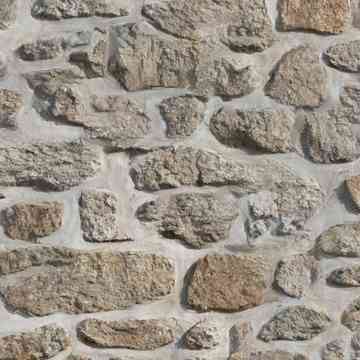}
\end{minipage}
\begin{minipage}[c]{0.24\linewidth}
\includegraphics[width=\linewidth]{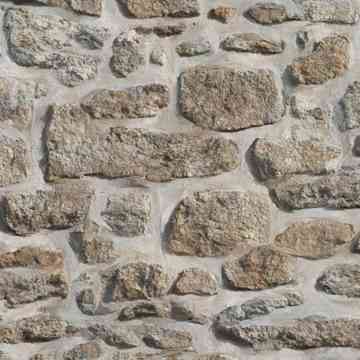}
\end{minipage}
\begin{minipage}[c]{0.24\linewidth}
\includegraphics[width=\linewidth]{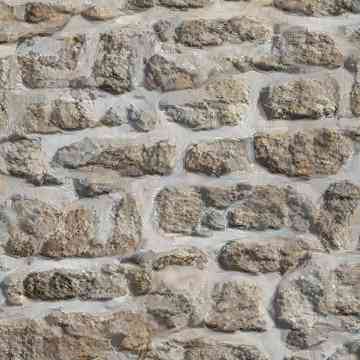}
\end{minipage}
\begin{minipage}[c]{0.24\linewidth}
\includegraphics[width=\linewidth]{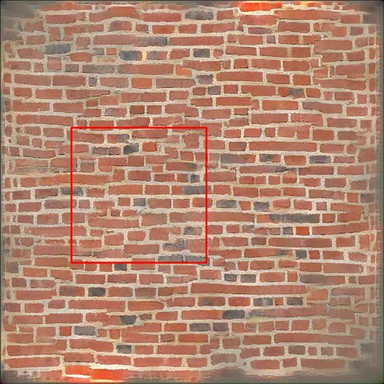}
\end{minipage}
\begin{minipage}[c]{0.24\linewidth}
\includegraphics[width=\linewidth]{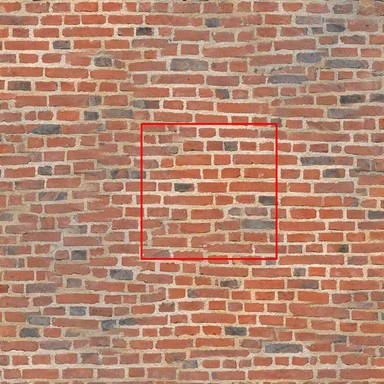}
\end{minipage}
\begin{minipage}[c]{0.24\linewidth}
\includegraphics[width=\linewidth]{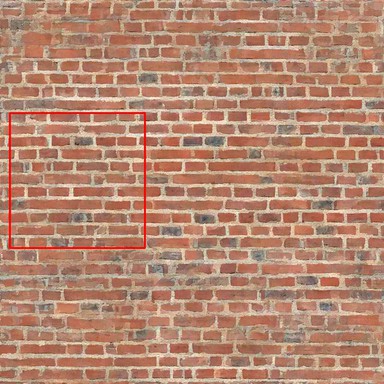}
\end{minipage}
\begin{minipage}[c]{0.24\linewidth}
\includegraphics[width=\linewidth]{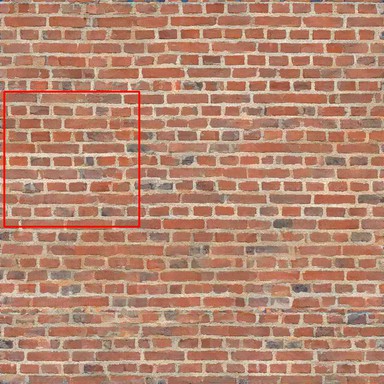}
\end{minipage}
\begin{minipage}[c]{0.24\linewidth}
\includegraphics[width=\linewidth]{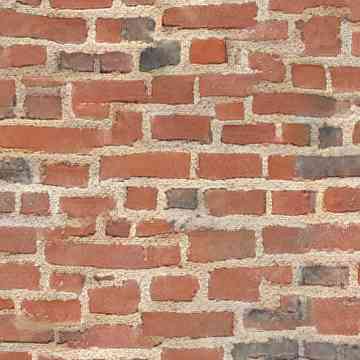}
\end{minipage}
\begin{minipage}[c]{0.24\linewidth}
\includegraphics[width=\linewidth]{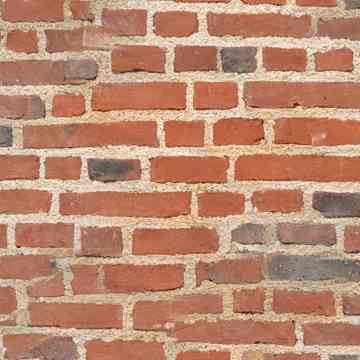}
\end{minipage}
\begin{minipage}[c]{0.24\linewidth}
\includegraphics[width=\linewidth]{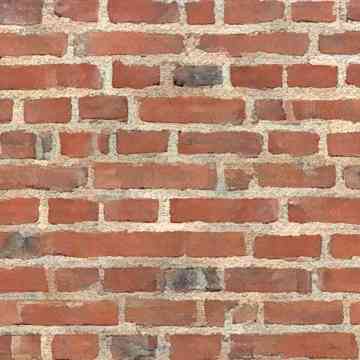}
\end{minipage}
\begin{minipage}[c]{0.24\linewidth}
\includegraphics[width=\linewidth]{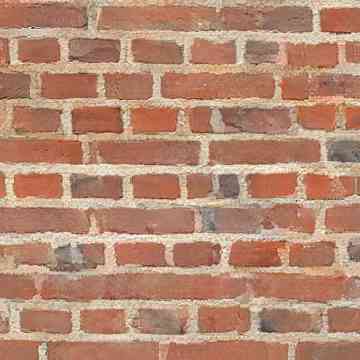}
\end{minipage}
\begin{minipage}[c]{0.24\linewidth}
\includegraphics[width=\linewidth]{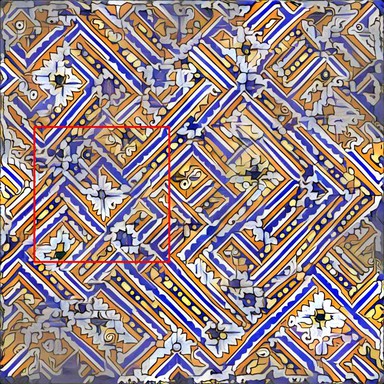}
\end{minipage}
\begin{minipage}[c]{0.24\linewidth}
\includegraphics[width=\linewidth]{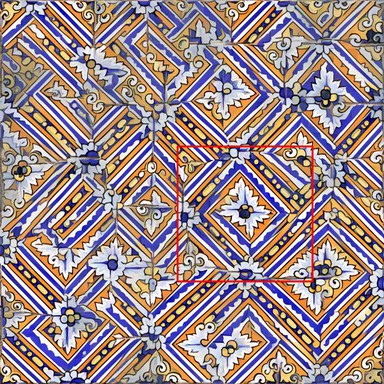}
\end{minipage}
\begin{minipage}[c]{0.24\linewidth}
\includegraphics[width=\linewidth]{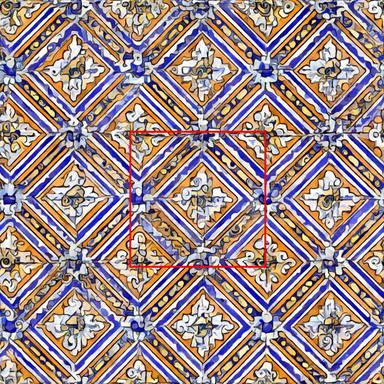}
\end{minipage}
\begin{minipage}[c]{0.24\linewidth}
\includegraphics[width=\linewidth]{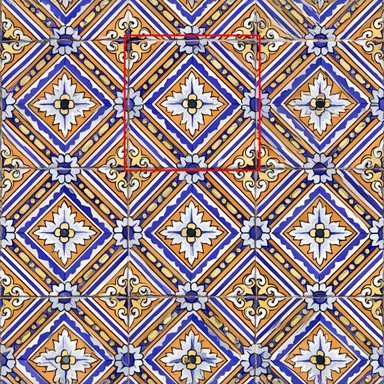}
\end{minipage}
\begin{minipage}[c]{0.24\linewidth}
\includegraphics[width=\linewidth]{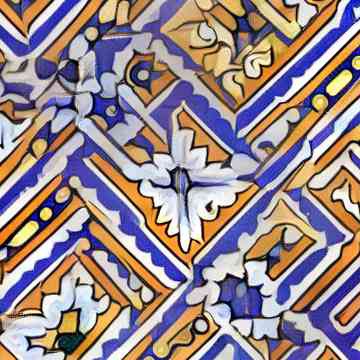}
\end{minipage}
\begin{minipage}[c]{0.24\linewidth}
\includegraphics[width=\linewidth]{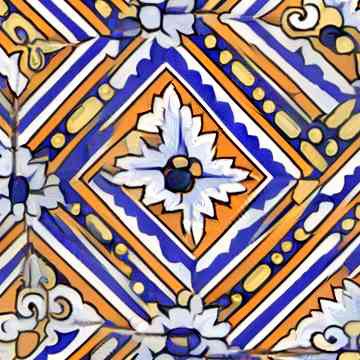}
\end{minipage}
\begin{minipage}[c]{0.24\linewidth}
\includegraphics[width=\linewidth]{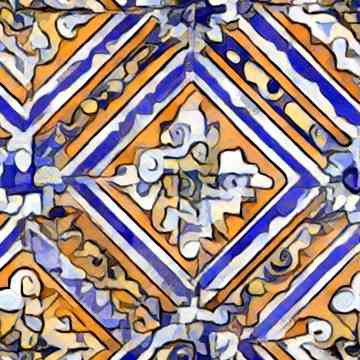}
\end{minipage}
\begin{minipage}[c]{0.24\linewidth}
\includegraphics[width=\linewidth]{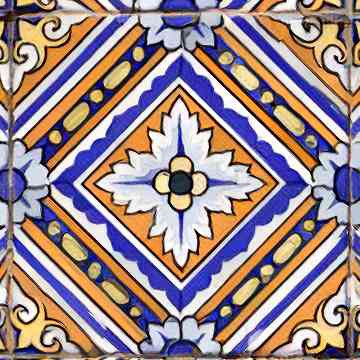}
\end{minipage}
\\
\caption{Zoom in some of the texture synthesis results. For the MSInit cases, we use $K = 2$. The region of each image framed by a red square is shown in the row below.}
\label{fig:ZoomDetails}
\end{figure}

\begin{figure}[!ht]
\centering
\begin{minipage}[c]{\linewidth}
\centering Regular images \\
\end{minipage}
\begin{minipage}[b]{0.09\linewidth}
\includegraphics[width=\linewidth]{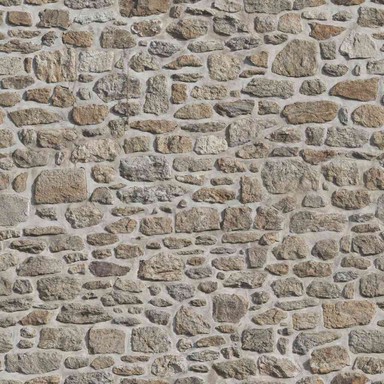}
\end{minipage}
\begin{minipage}[b]{0.09\linewidth}
\includegraphics[width=\linewidth]{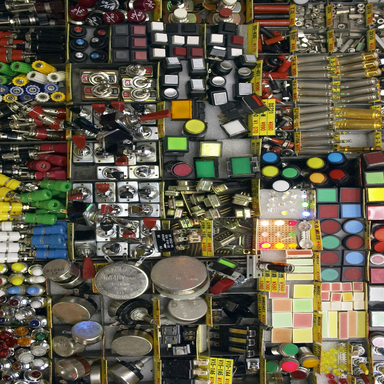}
\end{minipage}
\begin{minipage}[b]{0.09\linewidth}
\includegraphics[width=\linewidth]{./im/References/fabric_white_blue_1024}
\end{minipage}
\begin{minipage}[b]{0.09\linewidth}
\includegraphics[width=\linewidth]{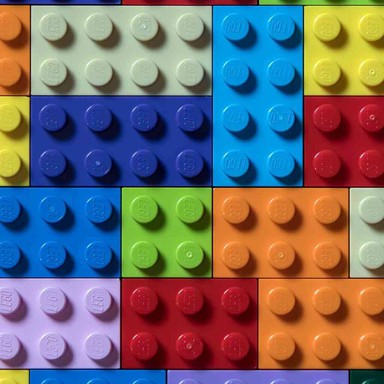}
\end{minipage}
\begin{minipage}[b]{0.09\linewidth}
\includegraphics[width=\linewidth]{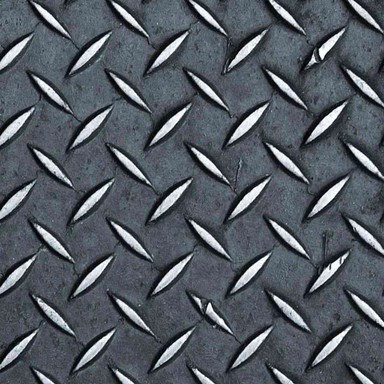}
\end{minipage}
\begin{minipage}[b]{0.09\linewidth}
\includegraphics[width=\linewidth]{./im/References/Pierzga_2006_1024}
\end{minipage}
\begin{minipage}[b]{0.09\linewidth}
\includegraphics[width=\linewidth]{./im/References/TexturesCom_BrickSmallBrown0473_1_M_1024}
\end{minipage}
\begin{minipage}[b]{0.09\linewidth}
\includegraphics[width=\linewidth]{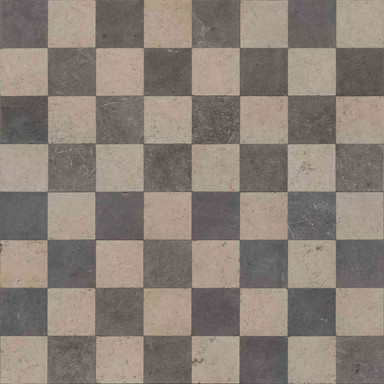}
\end{minipage}
\begin{minipage}[b]{0.09\linewidth}
\includegraphics[width=\linewidth]{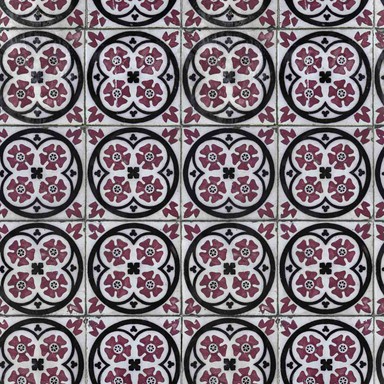}
\end{minipage}
\begin{minipage}[b]{0.09\linewidth}
\includegraphics[width=\linewidth]{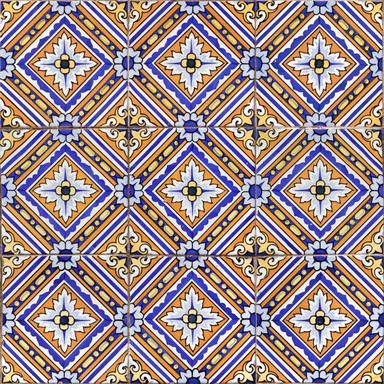}
\end{minipage}
\begin{minipage}[c]{\linewidth}
\centering Irregular images \\
\end{minipage}
\begin{minipage}[b]{0.09\linewidth}
\includegraphics[width=\linewidth]{./im/References/BubbleMarbel}
\end{minipage}
\begin{minipage}[b]{0.09\linewidth}
\includegraphics[width=\linewidth]{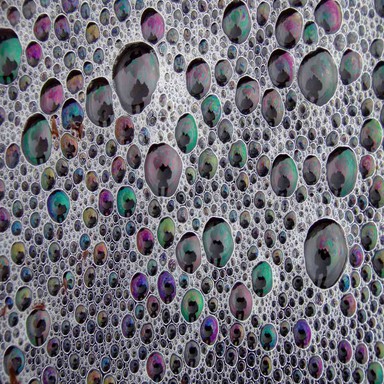}
\end{minipage}
\begin{minipage}[b]{0.09\linewidth}
\includegraphics[width=\linewidth]{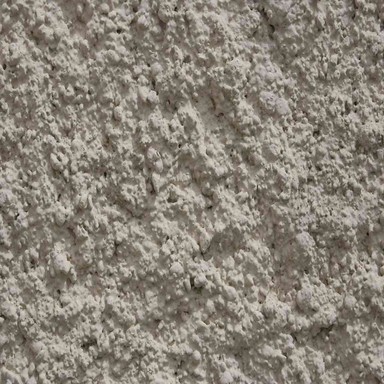}
\end{minipage}
\begin{minipage}[b]{0.09\linewidth}
\includegraphics[width=\linewidth]{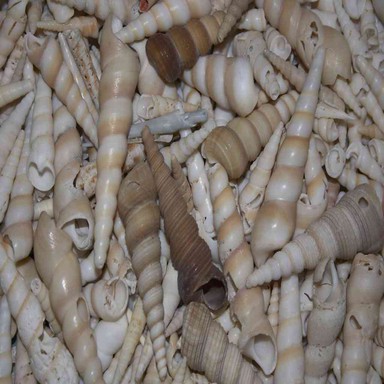}
\end{minipage}
\begin{minipage}[b]{0.09\linewidth}
\includegraphics[width=\linewidth]{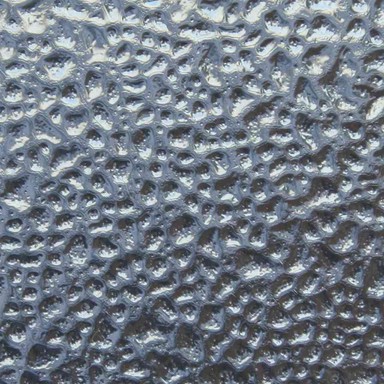}
\end{minipage}
\begin{minipage}[b]{0.09\linewidth}
\includegraphics[width=\linewidth]{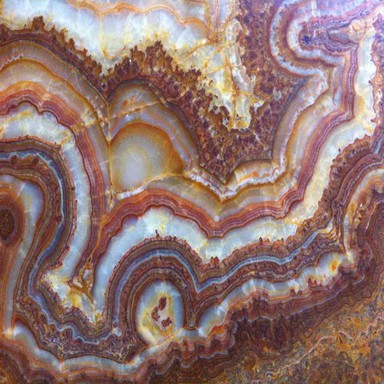}
\end{minipage}
\begin{minipage}[b]{0.09\linewidth}
\includegraphics[width=\linewidth]{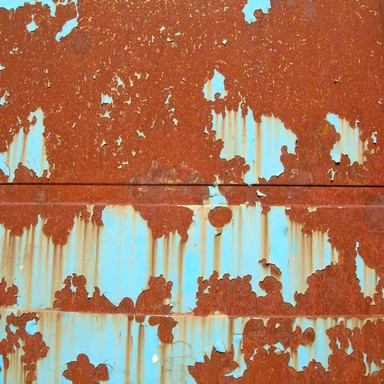}
\end{minipage}
\begin{minipage}[b]{0.09\linewidth}
\includegraphics[width=\linewidth]{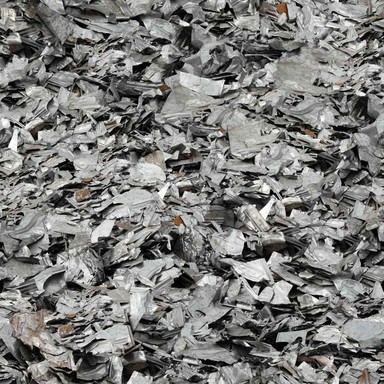}
\end{minipage}
\begin{minipage}[b]{0.09\linewidth}
\includegraphics[width=\linewidth]{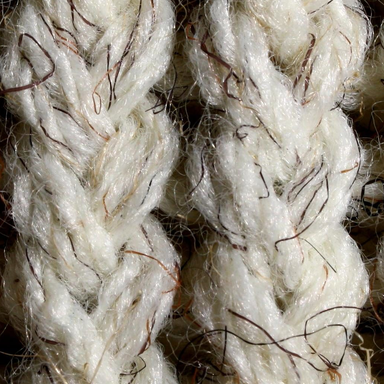}
\end{minipage}
\begin{minipage}[b]{0.09\linewidth}
\includegraphics[width=\linewidth]{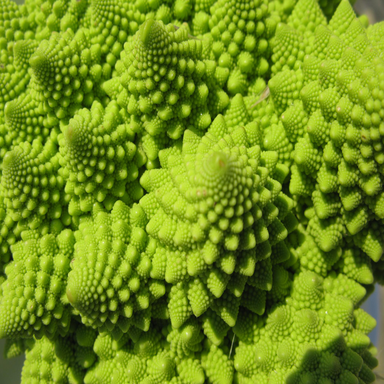}
\end{minipage}
\caption{Reference images used in the different evaluation methods.}
\label{fig:PerceptualTest_RefImages}
\end{figure}

\subsubsection{Verbatim copy} %
\label{sec:copy}

Texture synthesis methods should have the capacity to produce new images that are as diverse as possible. In the pioneering work FRAME~\cite{zhu1997}, this is achieved by maximizing the entropy. Similar ideas have recently been explored in~\cite{lu2015a,debortoli_macrocanonical_2019}. Following these ideas, texture synthesis methods could be evaluated based on their capacity to maximize the entropy under some given constraints. Such a quantitative evaluation, however, is far from being trivial and probably not tractable. In this section, we take a pragmatic and much more modest way. We propose a simple way to evaluate the tendency of methods to locally produce verbatim copy of the input. This is a known default of patch-based methods, see e.g. ~\cite{aguerrebere_exemplarbased_2013,raad_efros_2017}. 

\begin{figure}[!ht]
\centering
\begin{minipage}[b]{0.225\linewidth}
\includegraphics[width=\linewidth]{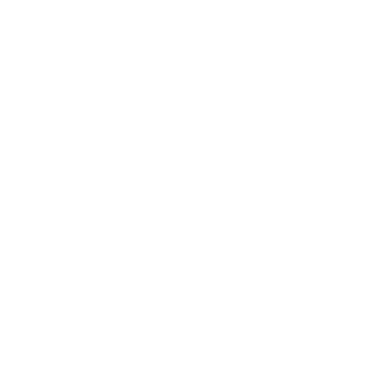}
\end{minipage}
\begin{minipage}[b]{0.225\linewidth}
\centering Gatys \cite{gatys_texture_2015} \\
\includegraphics[width=\linewidth]{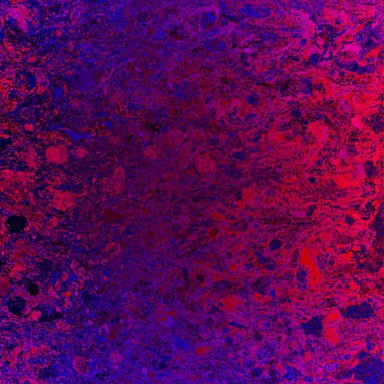}
\end{minipage}
\begin{minipage}[b]{0.225\linewidth}
\centering Efros Leung \cite{efros_texture_1999} \\
\includegraphics[width=\linewidth]{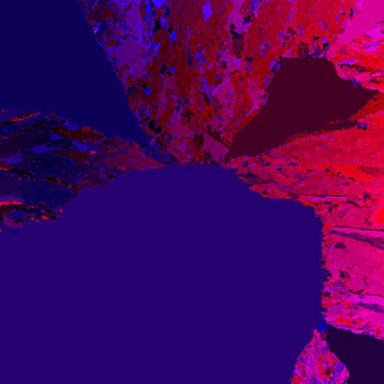}
\end{minipage}
\begin{minipage}[b]{0.225\linewidth}
\centering Efros Freeman \cite{efros_image_2001} \\
\includegraphics[width=\linewidth]{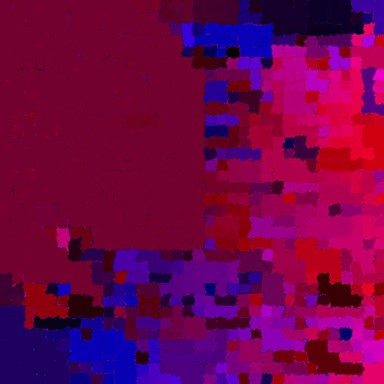}
\end{minipage}
\begin{minipage}[b]{0.225\linewidth}
\centering Ulyanov  \cite{ulyanov2016} \\
\includegraphics[width=\linewidth]{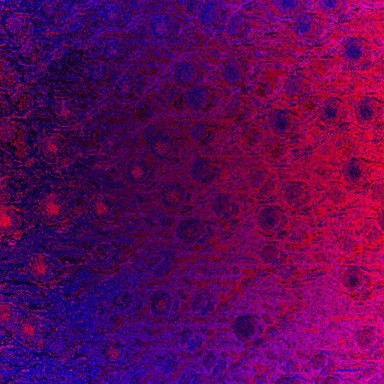}
\end{minipage}
\begin{minipage}[b]{0.225\linewidth}
\centering Gram + Spectrum \\
\includegraphics[width=\linewidth]{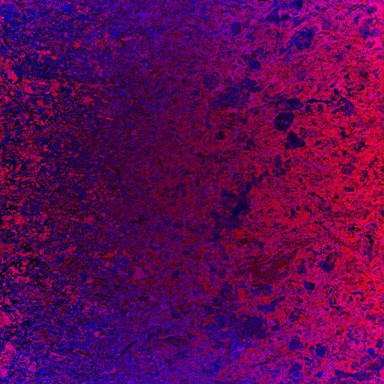}
\end{minipage}
\begin{minipage}[b]{0.225\linewidth}
\centering Deep Corr \cite{sendik_deep_2017} \\
\includegraphics[width=\linewidth]{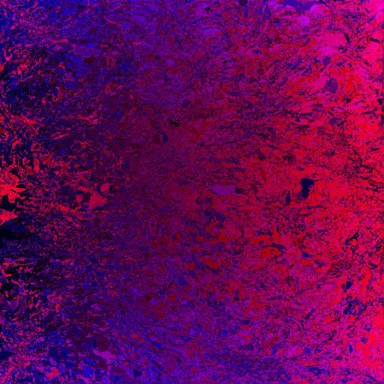}
\end{minipage}
\begin{minipage}[b]{0.225\linewidth}
\centering Autocorr \\
\includegraphics[width=\linewidth]{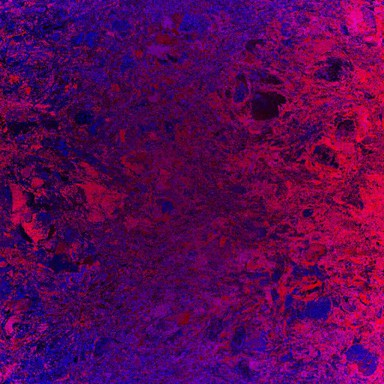}
\end{minipage}
\begin{minipage}[b]{0.225\linewidth}
\centering Snelgrove \cite{snelgrove_highresolution_2017} \\
\includegraphics[width=\linewidth]{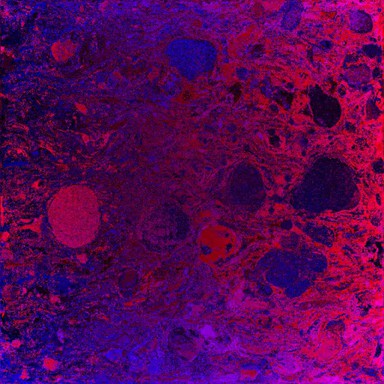}
\end{minipage}
\begin{minipage}[b]{0.225\linewidth}
\centering Gram + MSInit \\
\includegraphics[width=\linewidth]{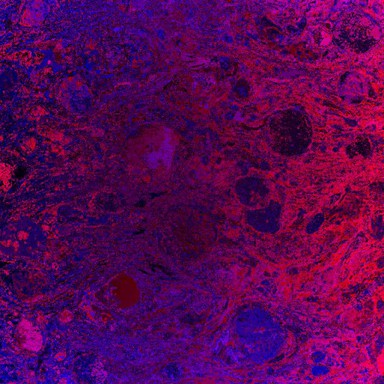}
\end{minipage}
\begin{minipage}[b]{0.225\linewidth}
\centering Gram + Spectrum + MSInit \\
\includegraphics[width=\linewidth]{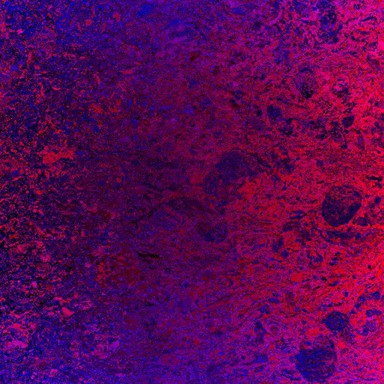}
\end{minipage}
\begin{minipage}[b]{0.225\linewidth}
\centering Autocorr + MSInit \\
\includegraphics[width=\linewidth]{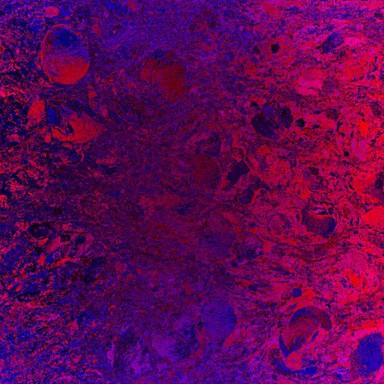}
\end{minipage}
\caption{Displacement map for results using different methods, for a given reference image.  An area with constant color indicates a verbatm copy of the input. The synthesis can be found in \ref{fig:BubbleMarbel}.}
\label{fig:DisMapBubbleMarbel}
\end{figure}

For each pixel of a given synthesis result, we look for its nearest neighbor in the input image. The notion of proximity is defined by comparing small square neighborhoods (patches) around each pixel. In Figure~\ref{fig:DisMapBubbleMarbel}, we display the corresponding displacement map. The used color scale is obtained by assigning the $x$ coordinate of the displacement map to red, and the $y$ coordinate to blue.
Verbatim copy of the input appear as constant regions in these displacement maps. As expected, the only two methods displaying large such regions are patch-based methods. All others seem to produce a reasonable amount of innovation, even though the multi-scale method from \cite{snelgrove_highresolution_2017} can very occasionally produce small verbatim copies, probably due to the strong constraints it puts on the Gaussian pyramid. %
 
In order to quantify the visual effect of the displacement maps, we propose to measure the flat regions corresponding to locally constant displacements. For each pixel of the displacement map, we count how many of its neighbors (in 4-connexity) share its color value. Denoting $n$ this number, a score is defined as $DS=1-n/N$, where $N$ is the total number of investigated neighbors. The more verbatim copy there are in the synthesis, the closest the score is to 0.

The box plots of this score for the different methods and the twenty reference images can be seen in \Cref{fig:DS}. They confirm the impression given by the displacement map that the patch-based methods yields significantly more verbatim copy than neural methods. It should be noted however that the proposed methodology is relatively rough and does not account neither for small perturbations on the pixel positions nor for noisy pixel values.

\begin{figure}    
\centering
 \resizebox{0.5\columnwidth}{!}{
 \input{imInTex/DS_Boxplots_per_method.tex}
 }
 \caption{Boxplots of the displacement score for the different methods on the twenty reference images of size $1024 \times 1024$.}
\label{fig:DS} 
\end{figure}

\subsubsection{Feature-based evaluation}
\label{sec:WaveletsKLStats}

Feature-based evaluation of textures is not straightforward, because no existing feature is considered as the reference one. Moreover, such evaluations are inherently biased. In the most extreme case, one could even try to optimize the chosen features to synthesize new textures. In this work, we choose to rely on wavelet filters, that both are classical texture features and are not used in any of the considered methods. More precisely, we rely on the texture features proposed in  \cite{do_waveletbased_2002}. In this paper, two textures are compared by computing the Kullback-Leibler divergence between parametric estimation (using generalized Gaussians) of the marginal distributions of wavelet coefficients. 

In order to quantify the proximity of a synthesized texture to the reference image, we propose to : 

\begin{enumerate}
 \item Compute the wavelets coefficients of the reference image and the synthesized one (in our case we choose a Daubechies 4 wavelets as in \cite{do_waveletbased_2002} with 8 scales instead of 3, in order to account for large scale structures). 
\item For each scale and orientations, estimate the parameters of a generalized Gaussian from the empirical distribution of wavelets coefficients
  \item  For each scale and orientations, compute the Kullback-Leibler divergence between the estimated generalized Gaussians (using a closed-form formula) 
\end{enumerate}

We display in \cref{fig:logKL} the boxplots of the log KL scores over the 20 considered images, for the different methods. For each box, the horizontal orange line corresponds to the average result and the star to the median. On the average, the best method for this evaluation scheme appears to be "Gram+MSInit". Then follow the two patch-based methods. This is in agreement with results from the previous paragraph, since indeed a verbatim copy will have a perfect score. The next method is "Gram+Spectrum+MSInit", followed by "Snelgrove" and "Autocorr + MSInit". This evaluation confirms the good quality of results produced by the proposed "XXX+MSInit" methods, as well as "Snelgrove", at least on this image dataset containing a relatively high proportion of structured textures.

\begin{figure}    
\centering
 \resizebox{0.5\columnwidth}{!}{
 \input{imInTex/KL_NScaleNone_logBoxplots_per_method.tex}
 }
 \caption{Boxplots of the displacement score for the different methods on the twenty reference images of size $1024 \times 1024$.}
\label{fig:logKL} 
\end{figure}

\subsubsection{Perceptual evaluation of texture synthesis methods}
\label{sec:perceptualEval}

Next, we further evaluate the proposed methods by performing a medium scale user study. Indeed, as shown in~\cite{dong_perceptual_2020}, feature-based methods such as the one of the previous section may not correlate very well with human observations, especially for long range structures.  

For ethical reasons, we decided not to rely on micro-work platforms. Most users involved are volunteer PhD students or researchers, which certainly induces some bias. The total number of persons involved was 93, each having the possibility to answer up to 40 questions. %

\paragraph{Methodology}

\begin{figure}[!tbp]
\centering
    \includegraphics[width=0.5\textwidth]{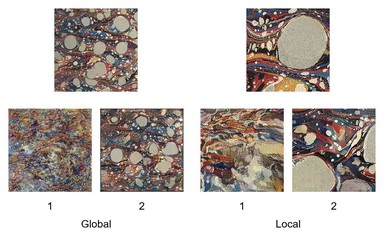}
\caption{Example of the layout for one question.}   
\label{fig:Exemple}
\end{figure}

Each question aims at comparing two methods on a given texture. In order to evaluate results at different scales, both the complete synthesis and a detail are presented to the user, see \cref{fig:Exemple}. The evaluation is performed on the twenty $1024 \times 1024$ images considered in this paper. In order to get further insight on the methods, we have split the textures in two groups : regular and irregular, see \Cref{fig:PerceptualTest_RefImages}.

Following the results of the previous sections, we chose to include in the study the five following methods : "Gatys" \cite{gatys_texture_2015}, "Gram+MSInit", "Gram+Spectrum+MSInit", "Snelgrove" \cite{snelgrove_highresolution_2017} and "Deep Coor" \cite{sendik_deep_2017}. The first four correspond to the best feature-based score and visual impression. The last one appears to us as the most directly related to the present work in the literature, since it explicitly aims at preserving large scale structures through additional statistical constraints. 

For each couple of methods (out of five) and each image, we build up two setups corresponding to the two possible respective position of methods (right and left) to avoid a possible lateral bias. This results in 400 different questions, for which we got 3170 answers. 

For each question, four images are presented corresponding to the two methods at two different scales (global and local). There are 4 possible answers (method 1 is the best for the global and the local scale, method 1 is the best for the local scale and method 2 the best for the global scale, etc.). Even though this is presented as a single question to the user, we treat its answers as two answers, one for the local scale and one for the global scale. This survey has been made with PsyToolkit servers \cite{stoet_psytoolkit_2017,stoet_psytoolkit_2010}.

It should be noted that asking a question such as "which result is most similar to the reference" is not trivial. Users were indicated that by "the most similar", it should be understood "which gives the most similar visual impression". Images are not expected to correspond pixel by pixel. Ideally, a synthesized image should give the impression to correspond to a different region of the same material as the reference.

\paragraph{Bradley-Terry model}

In order to quantify the results of this study, we rely on the Bradley-Terry model, as used in other perceptual study, see~\cite{um_spot_2019}. 

Let $\beta_{i} \in \mathbb{R}$ represent the strength of method $i$ (also called performance score), and let the outcome of a duel between methods $i$ and $j$ be determined by $\beta_{i} - \beta_{j}$.  The Bradley-Terry model treats these outcomes as  independent Bernoulli random variables with parameter $p_{ij}$,  where the log-odds corresponding to the probability $p_{ij}$ that method $i$ beats method $j$ is modeled as : 
\begin{equation}
\log \frac{p_{i j}}{1-p_{i j}}=\beta_{i}-\beta_{j}
\label{eq:beta_def}
\end{equation}
Equivalently, solving for $p_{i j}$ yields
\begin{equation}
p_{i j}=\frac{e^{\beta_{i}-\beta_{j}}}{1+e^{\beta_{i}-\beta_{j}}}=\frac{e^{\beta_{i}}}{e^{\beta_{i}}+e^{\beta_{j}}}
\end{equation}

This model is over-parameterized in the sense that it is exactly the same if we add a fixed constant to all values .
The Bradley-Terry model assigns scores to a fixed set of items based on pairwise comparisons of these items, where the log-odds of item "beating" item is given by the difference of their scores. 
The strength is estimated by second order optimization of the maximum likelihood and the standard deviation of the difference is approximated with the Hessian of this likelihood.

\paragraph{Duel results}

First, we can consider all the duels between all pairs of methods and all reference images, either from the complete set (20 images) or from the subsets of regular and irregular images separately. Results can be averaged for the global and local scale or treated separately. The results can be found on \Cref{fig:beta_global,fig:beta_local,fig:beta_both}. 

Overall, the two best methods for this evaluation appear to be "Gram+MSInit" and "Gram+Spectrum+MSInit".

For the global scale, there is a draw for the complete dataset and for the irregular images, while "Gram+Spectrum+MSInit" wins on the regular images. For the local scale, "Gram+MInit" always win.   

From this, we may deduce that the spectrum constraint may be useful for preserving large scale structure of regular texture, possibly at the price of a slight degradation at a more local scale. For more irregular textures, method "Gram + MSInit" should be preferred. 
When we consider all images and both scales (\cref{fig:beta_global}) we can extract a full ranking : "Gram+MSInit" $>$ "Gram+Spectrum+MSInit"  $>$ Snelgrov $>$ Gatys $>$ Deep Cor.

\begin{table}[!tbp]
\begin{center}
    Global case
\end{center}
   \begin{minipage}[b]{0.32\textwidth}
   \centering \small{All images} \\
   \resizebox{\columnwidth}{!}{
    \begin{tabular}{*{6}{p{1.75cm}|}}\mc{} & \mc{\makecell[c]{Gatys \cite{gatys_texture_2015}}} & \mc{\makecell[c]{Gram + \\ MSInit}} & \mc{\makecell[c]{Gram + \\ Spectrum + \\ MSInit}} & \mc{\makecell[c]{Snelgrove \cite{snelgrove_highresolution_2017}}} & \mc{\makecell[c]{Deep Corr \cite{sendik_deep_2017}}}\\ \cline{3-6}
\mc{\makecell*[c]{Gatys \cite{gatys_texture_2015}}} & \mcr{}  & \makecell*[c]{{\color{red} -2.78e+00 } \\ {\color{red} (2.24e-01)}} & \makecell*[c]{{\color{red} -2.68e+00 } \\ {\color{red} (2.21e-01)}} & \makecell*[c]{{\color{red} -2.17e+00 } \\ {\color{red} (2.11e-01)}} & \makecell*[c]{{ 4.00e-02 } \\ { (1.86e-01)}}\\ \cline{2-6}
\mcr{\makecell*[c]{Gram + \\ MSInit}} & \makecell*[c]{{\color{darkpastelgreen} 2.78e+00 } \\ {\color{darkpastelgreen} (2.24e-01)}} & \mcr{}  & \makecell*[c]{{ 1.07e-01 } \\ { (1.66e-01)}} & \makecell*[c]{{\color{darkpastelgreen} 6.12e-01 } \\ {\color{darkpastelgreen} (1.70e-01)}} & \makecell*[c]{{\color{darkpastelgreen} 2.82e+00 } \\ {\color{darkpastelgreen} (2.26e-01)}}\\ \cline{2-6}
\mcr{\makecell*[c]{Gram + \\ Spectrum + \\ MSInit}} & \makecell*[c]{{\color{darkpastelgreen} 2.68e+00 } \\ {\color{darkpastelgreen} (2.21e-01)}} & \makecell*[c]{{ -1.07e-01 } \\ { (1.66e-01)}} & \mcr{}  & \makecell*[c]{{\color{darkpastelgreen} 5.05e-01 } \\ {\color{darkpastelgreen} (1.67e-01)}} & \makecell*[c]{{\color{darkpastelgreen} 2.72e+00 } \\ {\color{darkpastelgreen} (2.23e-01)}}\\ \cline{2-6}
\mcr{\makecell*[c]{Snelgrove \cite{snelgrove_highresolution_2017}}} & \makecell*[c]{{\color{darkpastelgreen} 2.17e+00 } \\ {\color{darkpastelgreen} (2.11e-01)}} & \makecell*[c]{{\color{red} -6.12e-01 } \\ {\color{red} (1.70e-01)}} & \makecell*[c]{{\color{red} -5.05e-01 } \\ {\color{red} (1.67e-01)}} & \mcr{}  & \makecell*[c]{{\color{darkpastelgreen} 2.21e+00 } \\ {\color{darkpastelgreen} (2.14e-01)}}\\ \cline{2-6}
\mcr{\makecell*[c]{Deep Corr \cite{sendik_deep_2017}}} & \makecell*[c]{{ -4.00e-02 } \\ { (1.86e-01)}} & \makecell*[c]{{\color{red} -2.82e+00 } \\ {\color{red} (2.26e-01)}} & \makecell*[c]{{\color{red} -2.72e+00 } \\ {\color{red} (2.23e-01)}} & \makecell*[c]{{\color{red} -2.21e+00 } \\ {\color{red} (2.14e-01)}} & \mc{} \\ \cline{2-5}
\end{tabular}
    }
     \end{minipage} \hfill
       \begin{minipage}[b]{0.32\textwidth}
         \centering \small{Regular images} \\
   \resizebox{\columnwidth}{!}{
  \begin{tabular}{*{6}{p{1.75cm}|}}\mc{} & \mc{\makecell[c]{Gatys \cite{gatys_texture_2015}}} & \mc{\makecell[c]{Gram + \\ MSInit}} & \mc{\makecell[c]{Gram + \\ Spectrum + \\ MSInit}} & \mc{\makecell[c]{Snelgrove \cite{snelgrove_highresolution_2017}}} & \mc{\makecell[c]{Deep Corr \cite{sendik_deep_2017}}}\\ \cline{3-6}
\mc{\makecell*[c]{Gatys \cite{gatys_texture_2015}}} & \mcr{}  & \makecell*[c]{{\color{red} -2.67e+00 } \\ {\color{red} (3.19e-01)}} & \makecell*[c]{{\color{red} -3.54e+00 } \\ {\color{red} (3.50e-01)}} & \makecell*[c]{{\color{red} -1.93e+00 } \\ {\color{red} (2.95e-01)}} & \makecell*[c]{{ -2.17e-01 } \\ { (2.62e-01)}}\\ \cline{2-6}
\mcr{\makecell*[c]{Gram + \\ MSInit}} & \makecell*[c]{{\color{darkpastelgreen} 2.67e+00 } \\ {\color{darkpastelgreen} (3.19e-01)}} & \mcr{}  & \makecell*[c]{{\color{red} -8.73e-01 } \\ {\color{red} (2.61e-01)}} & \makecell*[c]{{\color{darkpastelgreen} 7.35e-01 } \\ {\color{darkpastelgreen} (2.49e-01)}} & \makecell*[c]{{\color{darkpastelgreen} 2.45e+00 } \\ {\color{darkpastelgreen} (3.09e-01)}}\\ \cline{2-6}
\mcr{\makecell*[c]{Gram + \\ Spectrum + \\ MSInit}} & \makecell*[c]{{\color{darkpastelgreen} 3.54e+00 } \\ {\color{darkpastelgreen} (3.50e-01)}} & \makecell*[c]{{\color{darkpastelgreen} 8.73e-01 } \\ {\color{darkpastelgreen} (2.61e-01)}} & \mcr{}  & \makecell*[c]{{\color{darkpastelgreen} 1.61e+00 } \\ {\color{darkpastelgreen} (2.81e-01)}} & \makecell*[c]{{\color{darkpastelgreen} 3.33e+00 } \\ {\color{darkpastelgreen} (3.40e-01)}}\\ \cline{2-6}
\mcr{\makecell*[c]{Snelgrove \cite{snelgrove_highresolution_2017}}} & \makecell*[c]{{\color{darkpastelgreen} 1.93e+00 } \\ {\color{darkpastelgreen} (2.95e-01)}} & \makecell*[c]{{\color{red} -7.35e-01 } \\ {\color{red} (2.49e-01)}} & \makecell*[c]{{\color{red} -1.61e+00 } \\ {\color{red} (2.81e-01)}} & \mcr{}  & \makecell*[c]{{\color{darkpastelgreen} 1.72e+00 } \\ {\color{darkpastelgreen} (2.85e-01)}}\\ \cline{2-6}
\mcr{\makecell*[c]{Deep Corr \cite{sendik_deep_2017}}} & \makecell*[c]{{ 2.17e-01 } \\ { (2.62e-01)}} & \makecell*[c]{{\color{red} -2.45e+00 } \\ {\color{red} (3.09e-01)}} & \makecell*[c]{{\color{red} -3.33e+00 } \\ {\color{red} (3.40e-01)}} & \makecell*[c]{{\color{red} -1.72e+00 } \\ {\color{red} (2.85e-01)}} & \mc{} \\ \cline{2-5}
\end{tabular}
  }
    \end{minipage} \hfill
       \begin{minipage}[b]{0.32\textwidth}
          \centering \small{Irregular images} \\
   \resizebox{\columnwidth}{!}{
    \begin{tabular}{*{6}{p{1.75cm}|}}\mc{} & \mc{\makecell[c]{Gatys \cite{gatys_texture_2015}}} & \mc{\makecell[c]{Gram + \\ MSInit}} & \mc{\makecell[c]{Gram + \\ Spectrum + \\ MSInit}} & \mc{\makecell[c]{Snelgrove \cite{snelgrove_highresolution_2017}}} & \mc{\makecell[c]{Deep Corr \cite{sendik_deep_2017}}}\\ \cline{3-6}
\mc{\makecell*[c]{Gatys \cite{gatys_texture_2015}}} & \mcr{}  & \makecell*[c]{{\color{red} -2.78e+00 } \\ {\color{red} (2.24e-01)}} & \makecell*[c]{{\color{red} -2.68e+00 } \\ {\color{red} (2.21e-01)}} & \makecell*[c]{{\color{red} -2.17e+00 } \\ {\color{red} (2.11e-01)}} & \makecell*[c]{{ 4.00e-02 } \\ { (1.86e-01)}}\\ \cline{2-6}
\mcr{\makecell*[c]{Gram + \\ MSInit}} & \makecell*[c]{{\color{darkpastelgreen} 2.78e+00 } \\ {\color{darkpastelgreen} (2.24e-01)}} & \mcr{}  & \makecell*[c]{{ 1.07e-01 } \\ { (1.66e-01)}} & \makecell*[c]{{\color{darkpastelgreen} 6.12e-01 } \\ {\color{darkpastelgreen} (1.70e-01)}} & \makecell*[c]{{\color{darkpastelgreen} 2.82e+00 } \\ {\color{darkpastelgreen} (2.26e-01)}}\\ \cline{2-6}
\mcr{\makecell*[c]{Gram + \\ Spectrum + \\ MSInit}} & \makecell*[c]{{\color{darkpastelgreen} 2.68e+00 } \\ {\color{darkpastelgreen} (2.21e-01)}} & \makecell*[c]{{ -1.07e-01 } \\ { (1.66e-01)}} & \mcr{}  & \makecell*[c]{{\color{darkpastelgreen} 5.05e-01 } \\ {\color{darkpastelgreen} (1.67e-01)}} & \makecell*[c]{{\color{darkpastelgreen} 2.72e+00 } \\ {\color{darkpastelgreen} (2.23e-01)}}\\ \cline{2-6}
\mcr{\makecell*[c]{Snelgrove \cite{snelgrove_highresolution_2017}}} & \makecell*[c]{{\color{darkpastelgreen} 2.17e+00 } \\ {\color{darkpastelgreen} (2.11e-01)}} & \makecell*[c]{{\color{red} -6.12e-01 } \\ {\color{red} (1.70e-01)}} & \makecell*[c]{{\color{red} -5.05e-01 } \\ {\color{red} (1.67e-01)}} & \mcr{}  & \makecell*[c]{{\color{darkpastelgreen} 2.21e+00 } \\ {\color{darkpastelgreen} (2.14e-01)}}\\ \cline{2-6}
\mcr{\makecell*[c]{Deep Corr \cite{sendik_deep_2017}}} & \makecell*[c]{{ -4.00e-02 } \\ { (1.86e-01)}} & \makecell*[c]{{\color{red} -2.82e+00 } \\ {\color{red} (2.26e-01)}} & \makecell*[c]{{\color{red} -2.72e+00 } \\ {\color{red} (2.23e-01)}} & \makecell*[c]{{\color{red} -2.21e+00 } \\ {\color{red} (2.14e-01)}} & \mc{} \\ \cline{2-5}
\end{tabular}
    }
     \end{minipage}
\caption{Difference between the methods strengths ($\beta_i - \beta_j$) (eq. \cref{eq:beta_def})
Index $i$ corresponds to rows and index $j$ to  columns. When $|\beta_i - \beta_j| > 1.96\hat{\mathrm{se}}_{i j}$ the method $i$ is considered as beatting  the method $j$ and the cell is displayed in green. In the opposite case, the cell is red. When the cell is white, the difference is not significant.}   
\label{fig:beta_global}
\end{table}

\begin{table}[!tbp]
\begin{center}
    Local case
\end{center}
   \begin{minipage}[b]{0.3\textwidth}
   \centering \small{All images} \\
   \resizebox{\columnwidth}{!}{
    \begin{tabular}{*{6}{p{1.75cm}|}}\mc{} & \mc{\makecell[c]{Gatys \cite{gatys_texture_2015}}} & \mc{\makecell[c]{Gram + \\ MSInit}} & \mc{\makecell[c]{Gram + \\ Spectrum + \\ MSInit}} & \mc{\makecell[c]{Snelgrove \cite{snelgrove_highresolution_2017}}} & \mc{\makecell[c]{Deep Corr \cite{sendik_deep_2017}}}\\ \cline{3-6}
\mc{\makecell*[c]{Gatys \cite{gatys_texture_2015}}} & \mcr{}  & \makecell*[c]{{\color{red} -1.62e+00 } \\ {\color{red} (1.77e-01)}} & \makecell*[c]{{\color{red} -9.65e-01 } \\ {\color{red} (1.65e-01)}} & \makecell*[c]{{\color{red} -1.02e+00 } \\ {\color{red} (1.64e-01)}} & \makecell*[c]{{\color{darkpastelgreen} 1.45e+00 } \\ {\color{darkpastelgreen} (2.04e-01)}}\\ \cline{2-6}
\mcr{\makecell*[c]{Gram + \\ MSInit}} & \makecell*[c]{{\color{darkpastelgreen} 1.62e+00 } \\ {\color{darkpastelgreen} (1.77e-01)}} & \mcr{}  & \makecell*[c]{{\color{darkpastelgreen} 6.59e-01 } \\ {\color{darkpastelgreen} (1.61e-01)}} & \makecell*[c]{{\color{darkpastelgreen} 6.04e-01 } \\ {\color{darkpastelgreen} (1.63e-01)}} & \makecell*[c]{{\color{darkpastelgreen} 3.07e+00 } \\ {\color{darkpastelgreen} (2.27e-01)}}\\ \cline{2-6}
\mcr{\makecell*[c]{Gram + \\ Spectrum + \\ MSInit}} & \makecell*[c]{{\color{darkpastelgreen} 9.65e-01 } \\ {\color{darkpastelgreen} (1.65e-01)}} & \makecell*[c]{{\color{red} -6.59e-01 } \\ {\color{red} (1.61e-01)}} & \mcr{}  & \makecell*[c]{{ -5.59e-02 } \\ { (1.54e-01)}} & \makecell*[c]{{\color{darkpastelgreen} 2.41e+00 } \\ {\color{darkpastelgreen} (2.15e-01)}}\\ \cline{2-6}
\mcr{\makecell*[c]{Snelgrove \cite{snelgrove_highresolution_2017}}} & \makecell*[c]{{\color{darkpastelgreen} 1.02e+00 } \\ {\color{darkpastelgreen} (1.64e-01)}} & \makecell*[c]{{\color{red} -6.04e-01 } \\ {\color{red} (1.63e-01)}} & \makecell*[c]{{ 5.59e-02 } \\ { (1.54e-01)}} & \mcr{}  & \makecell*[c]{{\color{darkpastelgreen} 2.47e+00 } \\ {\color{darkpastelgreen} (2.16e-01)}}\\ \cline{2-6}
\mcr{\makecell*[c]{Deep Corr \cite{sendik_deep_2017}}} & \makecell*[c]{{\color{red} -1.45e+00 } \\ {\color{red} (2.04e-01)}} & \makecell*[c]{{\color{red} -3.07e+00 } \\ {\color{red} (2.27e-01)}} & \makecell*[c]{{\color{red} -2.41e+00 } \\ {\color{red} (2.15e-01)}} & \makecell*[c]{{\color{red} -2.47e+00 } \\ {\color{red} (2.16e-01)}} & \mc{} \\ \cline{2-5}
\end{tabular}
    }
     \end{minipage} \hfill
       \begin{minipage}[b]{0.3\textwidth}
         \centering \small{Regular images} \\
   \resizebox{\columnwidth}{!}{
  \begin{tabular}{*{6}{p{1.75cm}|}}\mc{} & \mc{\makecell[c]{Gatys \cite{gatys_texture_2015}}} & \mc{\makecell[c]{Gram + \\ MSInit}} & \mc{\makecell[c]{Gram + \\ Spectrum + \\ MSInit}} & \mc{\makecell[c]{Snelgrove \cite{snelgrove_highresolution_2017}}} & \mc{\makecell[c]{Deep Corr \cite{sendik_deep_2017}}}\\ \cline{3-6}
\mc{\makecell*[c]{Gatys \cite{gatys_texture_2015}}} & \mcr{}  & \makecell*[c]{{\color{red} -2.31e+00 } \\ {\color{red} (2.89e-01)}} & \makecell*[c]{{\color{red} -1.74e+00 } \\ {\color{red} (2.74e-01)}} & \makecell*[c]{{\color{red} -1.80e+00 } \\ {\color{red} (2.74e-01)}} & \makecell*[c]{{\color{darkpastelgreen} 8.95e-01 } \\ {\color{darkpastelgreen} (2.78e-01)}}\\ \cline{2-6}
\mcr{\makecell*[c]{Gram + \\ MSInit}} & \makecell*[c]{{\color{darkpastelgreen} 2.31e+00 } \\ {\color{darkpastelgreen} (2.89e-01)}} & \mcr{}  & \makecell*[c]{{\color{darkpastelgreen} 5.74e-01 } \\ {\color{darkpastelgreen} (2.33e-01)}} & \makecell*[c]{{\color{darkpastelgreen} 5.15e-01 } \\ {\color{darkpastelgreen} (2.38e-01)}} & \makecell*[c]{{\color{darkpastelgreen} 3.21e+00 } \\ {\color{darkpastelgreen} (3.32e-01)}}\\ \cline{2-6}
\mcr{\makecell*[c]{Gram + \\ Spectrum + \\ MSInit}} & \makecell*[c]{{\color{darkpastelgreen} 1.74e+00 } \\ {\color{darkpastelgreen} (2.74e-01)}} & \makecell*[c]{{\color{red} -5.74e-01 } \\ {\color{red} (2.33e-01)}} & \mcr{}  & \makecell*[c]{{ -5.90e-02 } \\ { (2.29e-01)}} & \makecell*[c]{{\color{darkpastelgreen} 2.64e+00 } \\ {\color{darkpastelgreen} (3.16e-01)}}\\ \cline{2-6}
\mcr{\makecell*[c]{Snelgrove \cite{snelgrove_highresolution_2017}}} & \makecell*[c]{{\color{darkpastelgreen} 1.80e+00 } \\ {\color{darkpastelgreen} (2.74e-01)}} & \makecell*[c]{{\color{red} -5.15e-01 } \\ {\color{red} (2.38e-01)}} & \makecell*[c]{{ 5.90e-02 } \\ { (2.29e-01)}} & \mcr{}  & \makecell*[c]{{\color{darkpastelgreen} 2.69e+00 } \\ {\color{darkpastelgreen} (3.18e-01)}}\\ \cline{2-6}
\mcr{\makecell*[c]{Deep Corr \cite{sendik_deep_2017}}} & \makecell*[c]{{\color{red} -8.95e-01 } \\ {\color{red} (2.78e-01)}} & \makecell*[c]{{\color{red} -3.21e+00 } \\ {\color{red} (3.32e-01)}} & \makecell*[c]{{\color{red} -2.64e+00 } \\ {\color{red} (3.16e-01)}} & \makecell*[c]{{\color{red} -2.69e+00 } \\ {\color{red} (3.18e-01)}} & \mc{} \\ \cline{2-5}
\end{tabular}
  }
    \end{minipage} \hfill
       \begin{minipage}[b]{0.3\textwidth}
          \centering \small{Irregular images} \\
   \resizebox{\columnwidth}{!}{
    \begin{tabular}{*{6}{p{1.75cm}|}}\mc{} & \mc{\makecell[c]{Gatys \cite{gatys_texture_2015}}} & \mc{\makecell[c]{Gram + \\ MSInit}} & \mc{\makecell[c]{Gram + \\ Spectrum + \\ MSInit}} & \mc{\makecell[c]{Snelgrove \cite{snelgrove_highresolution_2017}}} & \mc{\makecell[c]{Deep Corr \cite{sendik_deep_2017}}}\\ \cline{3-6}
\mc{\makecell*[c]{Gatys \cite{gatys_texture_2015}}} & \mcr{}  & \makecell*[c]{{\color{red} -1.62e+00 } \\ {\color{red} (1.77e-01)}} & \makecell*[c]{{\color{red} -9.65e-01 } \\ {\color{red} (1.65e-01)}} & \makecell*[c]{{\color{red} -1.02e+00 } \\ {\color{red} (1.64e-01)}} & \makecell*[c]{{\color{darkpastelgreen} 1.45e+00 } \\ {\color{darkpastelgreen} (2.04e-01)}}\\ \cline{2-6}
\mcr{\makecell*[c]{Gram + \\ MSInit}} & \makecell*[c]{{\color{darkpastelgreen} 1.62e+00 } \\ {\color{darkpastelgreen} (1.77e-01)}} & \mcr{}  & \makecell*[c]{{\color{darkpastelgreen} 6.59e-01 } \\ {\color{darkpastelgreen} (1.61e-01)}} & \makecell*[c]{{\color{darkpastelgreen} 6.04e-01 } \\ {\color{darkpastelgreen} (1.63e-01)}} & \makecell*[c]{{\color{darkpastelgreen} 3.07e+00 } \\ {\color{darkpastelgreen} (2.27e-01)}}\\ \cline{2-6}
\mcr{\makecell*[c]{Gram + \\ Spectrum + \\ MSInit}} & \makecell*[c]{{\color{darkpastelgreen} 9.65e-01 } \\ {\color{darkpastelgreen} (1.65e-01)}} & \makecell*[c]{{\color{red} -6.59e-01 } \\ {\color{red} (1.61e-01)}} & \mcr{}  & \makecell*[c]{{ -5.59e-02 } \\ { (1.54e-01)}} & \makecell*[c]{{\color{darkpastelgreen} 2.41e+00 } \\ {\color{darkpastelgreen} (2.15e-01)}}\\ \cline{2-6}
\mcr{\makecell*[c]{Snelgrove \cite{snelgrove_highresolution_2017}}} & \makecell*[c]{{\color{darkpastelgreen} 1.02e+00 } \\ {\color{darkpastelgreen} (1.64e-01)}} & \makecell*[c]{{\color{red} -6.04e-01 } \\ {\color{red} (1.63e-01)}} & \makecell*[c]{{ 5.59e-02 } \\ { (1.54e-01)}} & \mcr{}  & \makecell*[c]{{\color{darkpastelgreen} 2.47e+00 } \\ {\color{darkpastelgreen} (2.16e-01)}}\\ \cline{2-6}
\mcr{\makecell*[c]{Deep Corr \cite{sendik_deep_2017}}} & \makecell*[c]{{\color{red} -1.45e+00 } \\ {\color{red} (2.04e-01)}} & \makecell*[c]{{\color{red} -3.07e+00 } \\ {\color{red} (2.27e-01)}} & \makecell*[c]{{\color{red} -2.41e+00 } \\ {\color{red} (2.15e-01)}} & \makecell*[c]{{\color{red} -2.47e+00 } \\ {\color{red} (2.16e-01)}} & \mc{} \\ \cline{2-5}
\end{tabular}
    }
     \end{minipage}
\caption{Difference between the methods strengths ($\beta_i - \beta_j$) (eq. \cref{eq:beta_def})
Index $i$ corresponds to rows and index $j$ to  columns. When $|\beta_i - \beta_j| > 1.96\hat{\mathrm{se}}_{i j}$ the method $i$ is considered as beatting  the method $j$ and the cell is displayed in green. In the opposite case, the cell is red. When the cell is white, the difference is not significant.}   
\label{fig:beta_local}
\end{table}

\begin{table}[!tbp]
\begin{center}
   Gloabl and local case
\end{center}
   \begin{minipage}[b]{0.3\textwidth}
   \centering \small{All images} \\
   \resizebox{\columnwidth}{!}{
    \begin{tabular}{*{6}{p{1.75cm}|}}\mc{} & \mc{\makecell[c]{Gatys \cite{gatys_texture_2015}}} & \mc{\makecell[c]{Gram + \\ MSInit}} & \mc{\makecell[c]{Gram + \\ Spectrum + \\ MSInit}} & \mc{\makecell[c]{Snelgrove \cite{snelgrove_highresolution_2017}}} & \mc{\makecell[c]{Deep Corr \cite{sendik_deep_2017}}}\\ \cline{3-6}
\mc{\makecell*[c]{Gatys \cite{gatys_texture_2015}}} & \mcr{}  & \makecell*[c]{{\color{red} -2.10e+00 } \\ {\color{red} (1.36e-01)}} & \makecell*[c]{{\color{red} -1.70e+00 } \\ {\color{red} (1.30e-01)}} & \makecell*[c]{{\color{red} -1.50e+00 } \\ {\color{red} (1.26e-01)}} & \makecell*[c]{{\color{darkpastelgreen} 7.48e-01 } \\ {\color{darkpastelgreen} (1.31e-01)}}\\ \cline{2-6}
\mcr{\makecell*[c]{Gram + \\ MSInit}} & \makecell*[c]{{\color{darkpastelgreen} 2.10e+00 } \\ {\color{darkpastelgreen} (1.36e-01)}} & \mcr{}  & \makecell*[c]{{\color{darkpastelgreen} 3.93e-01 } \\ {\color{darkpastelgreen} (1.14e-01)}} & \makecell*[c]{{\color{darkpastelgreen} 5.98e-01 } \\ {\color{darkpastelgreen} (1.17e-01)}} & \makecell*[c]{{\color{darkpastelgreen} 2.85e+00 } \\ {\color{darkpastelgreen} (1.53e-01)}}\\ \cline{2-6}
\mcr{\makecell*[c]{Gram + \\ Spectrum + \\ MSInit}} & \makecell*[c]{{\color{darkpastelgreen} 1.70e+00 } \\ {\color{darkpastelgreen} (1.30e-01)}} & \makecell*[c]{{\color{red} -3.93e-01 } \\ {\color{red} (1.14e-01)}} & \mcr{}  & \makecell*[c]{{\color{darkpastelgreen} 2.05e-01 } \\ {\color{darkpastelgreen} (1.12e-01)}} & \makecell*[c]{{\color{darkpastelgreen} 2.45e+00 } \\ {\color{darkpastelgreen} (1.47e-01)}}\\ \cline{2-6}
\mcr{\makecell*[c]{Snelgrove \cite{snelgrove_highresolution_2017}}} & \makecell*[c]{{\color{darkpastelgreen} 1.50e+00 } \\ {\color{darkpastelgreen} (1.26e-01)}} & \makecell*[c]{{\color{red} -5.98e-01 } \\ {\color{red} (1.17e-01)}} & \makecell*[c]{{\color{red} -2.05e-01 } \\ {\color{red} (1.12e-01)}} & \mcr{}  & \makecell*[c]{{\color{darkpastelgreen} 2.25e+00 } \\ {\color{darkpastelgreen} (1.45e-01)}}\\ \cline{2-6}
\mcr{\makecell*[c]{Deep Corr \cite{sendik_deep_2017}}} & \makecell*[c]{{\color{red} -7.48e-01 } \\ {\color{red} (1.31e-01)}} & \makecell*[c]{{\color{red} -2.85e+00 } \\ {\color{red} (1.53e-01)}} & \makecell*[c]{{\color{red} -2.45e+00 } \\ {\color{red} (1.47e-01)}} & \makecell*[c]{{\color{red} -2.25e+00 } \\ {\color{red} (1.45e-01)}} & \mc{} \\ \cline{2-5}
\end{tabular}
    }
     \end{minipage} \hfill
       \begin{minipage}[b]{0.3\textwidth}
         \centering \small{Regular images} \\
   \resizebox{\columnwidth}{!}{
  \begin{tabular}{*{6}{p{1.75cm}|}}\mc{} & \mc{\makecell[c]{Gatys \cite{gatys_texture_2015}}} & \mc{\makecell[c]{Gram + \\ MSInit}} & \mc{\makecell[c]{Gram + \\ Spectrum + \\ MSInit}} & \mc{\makecell[c]{Snelgrove \cite{snelgrove_highresolution_2017}}} & \mc{\makecell[c]{Deep Corr \cite{sendik_deep_2017}}}\\ \cline{3-6}
\mc{\makecell*[c]{Gatys \cite{gatys_texture_2015}}} & \mcr{}  & \makecell*[c]{{\color{red} -2.42e+00 } \\ {\color{red} (2.10e-01)}} & \makecell*[c]{{\color{red} -2.50e+00 } \\ {\color{red} (2.12e-01)}} & \makecell*[c]{{\color{red} -1.82e+00 } \\ {\color{red} (1.98e-01)}} & \makecell*[c]{{\color{darkpastelgreen} 3.27e-01 } \\ {\color{darkpastelgreen} (1.84e-01)}}\\ \cline{2-6}
\mcr{\makecell*[c]{Gram + \\ MSInit}} & \makecell*[c]{{\color{darkpastelgreen} 2.42e+00 } \\ {\color{darkpastelgreen} (2.10e-01)}} & \mcr{}  & \makecell*[c]{{ -8.77e-02 } \\ { (1.64e-01)}} & \makecell*[c]{{\color{darkpastelgreen} 5.94e-01 } \\ {\color{darkpastelgreen} (1.68e-01)}} & \makecell*[c]{{\color{darkpastelgreen} 2.74e+00 } \\ {\color{darkpastelgreen} (2.19e-01)}}\\ \cline{2-6}
\mcr{\makecell*[c]{Gram + \\ Spectrum + \\ MSInit}} & \makecell*[c]{{\color{darkpastelgreen} 2.50e+00 } \\ {\color{darkpastelgreen} (2.12e-01)}} & \makecell*[c]{{ 8.77e-02 } \\ { (1.64e-01)}} & \mcr{}  & \makecell*[c]{{\color{darkpastelgreen} 6.82e-01 } \\ {\color{darkpastelgreen} (1.68e-01)}} & \makecell*[c]{{\color{darkpastelgreen} 2.83e+00 } \\ {\color{darkpastelgreen} (2.20e-01)}}\\ \cline{2-6}
\mcr{\makecell*[c]{Snelgrove \cite{snelgrove_highresolution_2017}}} & \makecell*[c]{{\color{darkpastelgreen} 1.82e+00 } \\ {\color{darkpastelgreen} (1.98e-01)}} & \makecell*[c]{{\color{red} -5.94e-01 } \\ {\color{red} (1.68e-01)}} & \makecell*[c]{{\color{red} -6.82e-01 } \\ {\color{red} (1.68e-01)}} & \mcr{}  & \makecell*[c]{{\color{darkpastelgreen} 2.15e+00 } \\ {\color{darkpastelgreen} (2.08e-01)}}\\ \cline{2-6}
\mcr{\makecell*[c]{Deep Corr \cite{sendik_deep_2017}}} & \makecell*[c]{{\color{red} -3.27e-01 } \\ {\color{red} (1.84e-01)}} & \makecell*[c]{{\color{red} -2.74e+00 } \\ {\color{red} (2.19e-01)}} & \makecell*[c]{{\color{red} -2.83e+00 } \\ {\color{red} (2.20e-01)}} & \makecell*[c]{{\color{red} -2.15e+00 } \\ {\color{red} (2.08e-01)}} & \mc{} \\ \cline{2-5}
\end{tabular}
  }
    \end{minipage} \hfill
       \begin{minipage}[b]{0.3\textwidth}
          \centering \small{Irregular images} \\
   \resizebox{\columnwidth}{!}{
    \begin{tabular}{*{6}{p{1.75cm}|}}\mc{} & \mc{\makecell[c]{Gatys \cite{gatys_texture_2015}}} & \mc{\makecell[c]{Gram + \\ MSInit}} & \mc{\makecell[c]{Gram + \\ Spectrum + \\ MSInit}} & \mc{\makecell[c]{Snelgrove \cite{snelgrove_highresolution_2017}}} & \mc{\makecell[c]{Deep Corr \cite{sendik_deep_2017}}}\\ \cline{3-6}
\mc{\makecell*[c]{Gatys \cite{gatys_texture_2015}}} & \mcr{}  & \makecell*[c]{{\color{red} -2.10e+00 } \\ {\color{red} (1.36e-01)}} & \makecell*[c]{{\color{red} -1.70e+00 } \\ {\color{red} (1.30e-01)}} & \makecell*[c]{{\color{red} -1.50e+00 } \\ {\color{red} (1.26e-01)}} & \makecell*[c]{{\color{darkpastelgreen} 7.48e-01 } \\ {\color{darkpastelgreen} (1.31e-01)}}\\ \cline{2-6}
\mcr{\makecell*[c]{Gram + \\ MSInit}} & \makecell*[c]{{\color{darkpastelgreen} 2.10e+00 } \\ {\color{darkpastelgreen} (1.36e-01)}} & \mcr{}  & \makecell*[c]{{\color{darkpastelgreen} 3.93e-01 } \\ {\color{darkpastelgreen} (1.14e-01)}} & \makecell*[c]{{\color{darkpastelgreen} 5.98e-01 } \\ {\color{darkpastelgreen} (1.17e-01)}} & \makecell*[c]{{\color{darkpastelgreen} 2.85e+00 } \\ {\color{darkpastelgreen} (1.53e-01)}}\\ \cline{2-6}
\mcr{\makecell*[c]{Gram + \\ Spectrum + \\ MSInit}} & \makecell*[c]{{\color{darkpastelgreen} 1.70e+00 } \\ {\color{darkpastelgreen} (1.30e-01)}} & \makecell*[c]{{\color{red} -3.93e-01 } \\ {\color{red} (1.14e-01)}} & \mcr{}  & \makecell*[c]{{\color{darkpastelgreen} 2.05e-01 } \\ {\color{darkpastelgreen} (1.12e-01)}} & \makecell*[c]{{\color{darkpastelgreen} 2.45e+00 } \\ {\color{darkpastelgreen} (1.47e-01)}}\\ \cline{2-6}
\mcr{\makecell*[c]{Snelgrove \cite{snelgrove_highresolution_2017}}} & \makecell*[c]{{\color{darkpastelgreen} 1.50e+00 } \\ {\color{darkpastelgreen} (1.26e-01)}} & \makecell*[c]{{\color{red} -5.98e-01 } \\ {\color{red} (1.17e-01)}} & \makecell*[c]{{\color{red} -2.05e-01 } \\ {\color{red} (1.12e-01)}} & \mcr{}  & \makecell*[c]{{\color{darkpastelgreen} 2.25e+00 } \\ {\color{darkpastelgreen} (1.45e-01)}}\\ \cline{2-6}
\mcr{\makecell*[c]{Deep Corr \cite{sendik_deep_2017}}} & \makecell*[c]{{\color{red} -7.48e-01 } \\ {\color{red} (1.31e-01)}} & \makecell*[c]{{\color{red} -2.85e+00 } \\ {\color{red} (1.53e-01)}} & \makecell*[c]{{\color{red} -2.45e+00 } \\ {\color{red} (1.47e-01)}} & \makecell*[c]{{\color{red} -2.25e+00 } \\ {\color{red} (1.45e-01)}} & \mc{} \\ \cline{2-5}
\end{tabular}
    }
     \end{minipage}
\caption{Difference between the methods strengths ($\beta_i - \beta_j$) (eq. \cref{eq:beta_def})
Index $i$ corresponds to rows and index $j$ to  columns. When $|\beta_i - \beta_j| > 1.96\hat{\mathrm{se}}_{i j}$ the method $i$ is considered as beatting  the method $j$ and the cell is displayed in green. In the opposite case, the cell is red. When the cell is white, the difference is not significant.}   
\label{fig:beta_both}
\end{table}

\paragraph{Winning probability} An alternative evaluation consists in calculating the probability that a method $i$ is chosen among all candidates. This "winning probability" is given by the average over $j$ of the probability $p_{i j}$ that a participant chooses the candidate $i$ over $j$:
\begin{equation}
W_{i}=\frac{1}{N-1} \sum_{j \neq i}^{N} p_{i j}=\frac{1}{N-1} \sum_{j \neq i}^{N} \frac{e^{\beta_{i}-\beta_{j}}}{1+e^{\beta_{i}-\beta_{j}}}
\label{eq:winning_proba}
\end{equation}
In contrast to the pairwise probability $p_{i j}, W_{i}$ represents the probability that a candidate $i$ was preferred over all other candidates.

We can estimate the standard error of $W_i$ as :
\begin{equation}
\Sigma_{i}= \frac{1}{N-1}  \sqrt{ \sum_{j \neq i}^{N}  \hat{\sigma}_{ij}^2 }
\label{eq:winning_proba_std}
\end{equation}
under the hypothesis that the $p_{ij}$ are independent. 

\begin{figure}[!tbp]
\centering
    Global and local case \\
  \begin{minipage}[b]{0.3\textwidth}
   \centering \small{All images} \\
   \resizebox{\columnwidth}{!}{
 \begin{tikzpicture}

\definecolor{color0}{rgb}{0.215686274509804,0.494117647058824,0.72156862745098}
\definecolor{color1}{rgb}{1,0.498039215686275,0}
\definecolor{color2}{rgb}{0.301960784313725,0.686274509803922,0.290196078431373}
\definecolor{color3}{rgb}{0.968627450980392,0.505882352941176,0.749019607843137}
\definecolor{color4}{rgb}{0.650980392156863,0.337254901960784,0.156862745098039}

\begin{axis}[
axis line style={white!80!black},
tick align=outside,
tick pos=both,
x grid style={white!80!black},
xmajorgrids,
xmin=-0.64, xmax=4.64,
xtick style={color=white!15!black},
xtick={0,1,2,3,4},
xticklabel style = {rotate=45.0,align=center},
xticklabels={Gatys \cite{gatys_texture_2015},{Gram +\\ MSInit},{Gram +\\ Spectrum +\\ MSInit},Snelgrove \cite{snelgrove_highresolution_2017},Deep Corr \cite{sendik_deep_2017}},
y grid style={white!80!black},
ylabel={Winning Prob},
ymajorgrids,
ymin=0, ymax=1,
ytick style={color=white!15!black},
ytick={0,0.2,0.4,0.6,0.8,1},
yticklabels={\(\displaystyle 0.0\),\(\displaystyle 0.2\),\(\displaystyle 0.4\),\(\displaystyle 0.6\),\(\displaystyle 0.8\),\(\displaystyle 1.0\)}
]
\draw[draw=white,fill=color0,opacity=0.5] (axis cs:-0.4,0) rectangle (axis cs:0.4,0.281097568181724);
\draw[draw=white,fill=color1,opacity=0.5] (axis cs:0.6,0) rectangle (axis cs:1.4,0.769561124502862);
\draw[draw=white,fill=color2,opacity=0.5] (axis cs:1.6,0) rectangle (axis cs:2.4,0.680236405479519);
\draw[draw=white,fill=color3,opacity=0.5] (axis cs:2.6,0) rectangle (axis cs:3.4,0.631393329572408);
\draw[draw=white,fill=color4,opacity=0.5] (axis cs:3.6,0) rectangle (axis cs:4.4,0.137711572263487);
\path [draw=black, semithick]
(axis cs:0,0.275938843663703)
--(axis cs:0,0.286256292699745);

\path [draw=black, semithick]
(axis cs:1,0.764286126737983)
--(axis cs:1,0.77483612226774);

\path [draw=black, semithick]
(axis cs:2,0.674644014241977)
--(axis cs:2,0.68582879671706);

\path [draw=black, semithick]
(axis cs:3,0.62570090145208)
--(axis cs:3,0.637085757692737);

\path [draw=black, semithick]
(axis cs:4,0.133374787577258)
--(axis cs:4,0.142048356949716);

\addplot [semithick, black, mark=-, mark size=10, mark options={solid}, only marks]
table {%
0 0.275938843663703
1 0.764286126737983
2 0.674644014241977
3 0.62570090145208
4 0.133374787577258
};
\addplot [semithick, black, mark=-, mark size=10, mark options={solid}, only marks]
table {%
0 0.286256292699745
1 0.77483612226774
2 0.68582879671706
3 0.637085757692737
4 0.142048356949716
};
\end{axis}

\end{tikzpicture}}
     \end{minipage}
      \begin{minipage}[b]{0.3\textwidth}
   \centering \small{Regular images} \\
   \resizebox{\columnwidth}{!}{
 \begin{tikzpicture}

\definecolor{color0}{rgb}{0.215686274509804,0.494117647058824,0.72156862745098}
\definecolor{color1}{rgb}{1,0.498039215686275,0}
\definecolor{color2}{rgb}{0.301960784313725,0.686274509803922,0.290196078431373}
\definecolor{color3}{rgb}{0.968627450980392,0.505882352941176,0.749019607843137}
\definecolor{color4}{rgb}{0.650980392156863,0.337254901960784,0.156862745098039}

\begin{axis}[
axis line style={white!80!black},
tick align=outside,
tick pos=both,
x grid style={white!80!black},
xmajorgrids,
xmin=-0.64, xmax=4.64,
xtick style={color=white!15!black},
xtick={0,1,2,3,4},
xticklabel style = {rotate=45.0,align=center},
xticklabels={Gatys \cite{gatys_texture_2015},{Gram +\\ MSInit},{Gram +\\ Spectrum +\\ MSInit},Snelgrove \cite{snelgrove_highresolution_2017},Deep Corr \cite{sendik_deep_2017}},
y grid style={white!80!black},
ylabel={Winning Prob},
ymajorgrids,
ymin=0, ymax=1,
ytick style={color=white!15!black},
ytick={0,0.2,0.4,0.6,0.8,1},
yticklabels={\(\displaystyle 0.0\),\(\displaystyle 0.2\),\(\displaystyle 0.4\),\(\displaystyle 0.6\),\(\displaystyle 0.8\),\(\displaystyle 1.0\)}
]
\draw[draw=white,fill=color0,opacity=0.5] (axis cs:-0.4,0) rectangle (axis cs:0.4,0.219335805332342);
\draw[draw=white,fill=color1,opacity=0.5] (axis cs:0.6,0) rectangle (axis cs:1.4,0.745010055740691);
\draw[draw=white,fill=color2,opacity=0.5] (axis cs:1.6,0) rectangle (axis cs:2.4,0.763720233022794);
\draw[draw=white,fill=color3,opacity=0.5] (axis cs:2.6,0) rectangle (axis cs:3.4,0.612053448517138);
\draw[draw=white,fill=color4,opacity=0.5] (axis cs:3.6,0) rectangle (axis cs:4.4,0.159880457387034);
\path [draw=black, semithick]
(axis cs:0,0.212306923204194)
--(axis cs:0,0.226364687460491);

\path [draw=black, semithick]
(axis cs:1,0.737401339491014)
--(axis cs:1,0.752618771990369);

\path [draw=black, semithick]
(axis cs:2,0.756240189568435)
--(axis cs:2,0.771200276477153);

\path [draw=black, semithick]
(axis cs:3,0.60417006576285)
--(axis cs:3,0.619936831271426);

\path [draw=black, semithick]
(axis cs:4,0.153282104240516)
--(axis cs:4,0.166478810533551);

\addplot [semithick, black, mark=-, mark size=10, mark options={solid}, only marks]
table {%
0 0.212306923204194
1 0.737401339491014
2 0.756240189568435
3 0.60417006576285
4 0.153282104240516
};
\addplot [semithick, black, mark=-, mark size=10, mark options={solid}, only marks]
table {%
0 0.226364687460491
1 0.752618771990369
2 0.771200276477153
3 0.619936831271426
4 0.166478810533551
};
\end{axis}

\end{tikzpicture}}
     \end{minipage}
      \begin{minipage}[b]{0.3\textwidth}
   \centering \small{Irregular images}  \\
   \resizebox{\columnwidth}{!}{
 \begin{tikzpicture}

\definecolor{color0}{rgb}{0.215686274509804,0.494117647058824,0.72156862745098}
\definecolor{color1}{rgb}{1,0.498039215686275,0}
\definecolor{color2}{rgb}{0.301960784313725,0.686274509803922,0.290196078431373}
\definecolor{color3}{rgb}{0.968627450980392,0.505882352941176,0.749019607843137}
\definecolor{color4}{rgb}{0.650980392156863,0.337254901960784,0.156862745098039}

\begin{axis}[
axis line style={white!80!black},
tick align=outside,
tick pos=both,
x grid style={white!80!black},
xmajorgrids,
xmin=-0.64, xmax=4.64,
xtick style={color=white!15!black},
xtick={0,1,2,3,4},
xticklabel style = {rotate=45.0,align=center},
xticklabels={Gatys \cite{gatys_texture_2015},{Gram +\\ MSInit},{Gram +\\ Spectrum +\\ MSInit},Snelgrove \cite{snelgrove_highresolution_2017},Deep Corr \cite{sendik_deep_2017}},
y grid style={white!80!black},
ylabel={Winning Prob},
ymajorgrids,
ymin=0, ymax=1,
ytick style={color=white!15!black},
ytick={0,0.2,0.4,0.6,0.8,1},
yticklabels={\(\displaystyle 0.0\),\(\displaystyle 0.2\),\(\displaystyle 0.4\),\(\displaystyle 0.6\),\(\displaystyle 0.8\),\(\displaystyle 1.0\)}
]
\draw[draw=white,fill=color0,opacity=0.5] (axis cs:-0.4,0) rectangle (axis cs:0.4,0.281097568181724);
\draw[draw=white,fill=color1,opacity=0.5] (axis cs:0.6,0) rectangle (axis cs:1.4,0.769561124502862);
\draw[draw=white,fill=color2,opacity=0.5] (axis cs:1.6,0) rectangle (axis cs:2.4,0.680236405479519);
\draw[draw=white,fill=color3,opacity=0.5] (axis cs:2.6,0) rectangle (axis cs:3.4,0.631393329572408);
\draw[draw=white,fill=color4,opacity=0.5] (axis cs:3.6,0) rectangle (axis cs:4.4,0.137711572263487);
\path [draw=black, semithick]
(axis cs:0,0.275938843663703)
--(axis cs:0,0.286256292699745);

\path [draw=black, semithick]
(axis cs:1,0.764286126737983)
--(axis cs:1,0.77483612226774);

\path [draw=black, semithick]
(axis cs:2,0.674644014241977)
--(axis cs:2,0.68582879671706);

\path [draw=black, semithick]
(axis cs:3,0.62570090145208)
--(axis cs:3,0.637085757692737);

\path [draw=black, semithick]
(axis cs:4,0.133374787577258)
--(axis cs:4,0.142048356949716);

\addplot [semithick, black, mark=-, mark size=10, mark options={solid}, only marks]
table {%
0 0.275938843663703
1 0.764286126737983
2 0.674644014241977
3 0.62570090145208
4 0.133374787577258
};
\addplot [semithick, black, mark=-, mark size=10, mark options={solid}, only marks]
table {%
0 0.286256292699745
1 0.77483612226774
2 0.68582879671706
3 0.637085757692737
4 0.142048356949716
};
\end{axis}

\end{tikzpicture}}
     \end{minipage}
\caption{Winning probabilities $W_i$ with standard error $\Sigma_i$ for the different methods for both global and local cases.}   
\label{fig:Wi_both}
\end{figure}
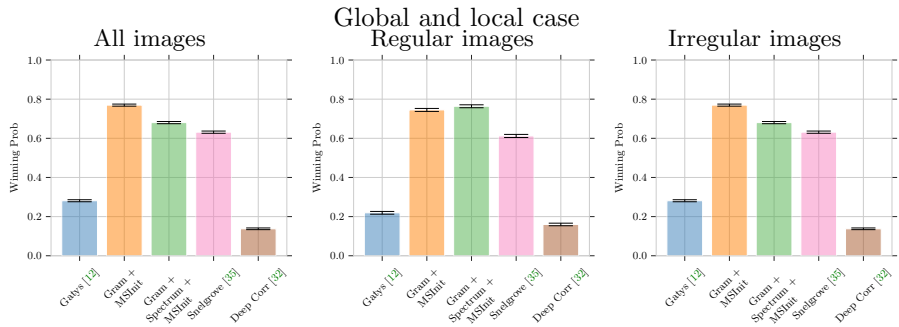

These winning probabilities are displayed on \Cref{fig:Wi_global,fig:Wi_local,fig:Wi_both} and confirm the duel results.   

 \begin{figure}[!tbp]
\centering
  Global case \\
   \begin{minipage}[b]{0.3\textwidth}
   \centering \small{ All images} \\
   \resizebox{\columnwidth}{!}{
 \begin{tikzpicture}

\definecolor{color0}{rgb}{0.215686274509804,0.494117647058824,0.72156862745098}
\definecolor{color1}{rgb}{1,0.498039215686275,0}
\definecolor{color2}{rgb}{0.301960784313725,0.686274509803922,0.290196078431373}
\definecolor{color3}{rgb}{0.968627450980392,0.505882352941176,0.749019607843137}
\definecolor{color4}{rgb}{0.650980392156863,0.337254901960784,0.156862745098039}

\begin{axis}[
axis line style={white!80!black},
tick align=outside,
tick pos=both,
x grid style={white!80!black},
xmajorgrids,
xmin=-0.64, xmax=4.64,
xtick style={color=white!15!black},
xtick={0,1,2,3,4},
xticklabel style = {rotate=45.0,align=center},
xticklabels={Gatys \cite{gatys_texture_2015},{Gram +\\ MSInit},{Gram +\\ Spectrum +\\ MSInit},Snelgrove \cite{snelgrove_highresolution_2017},Deep Corr \cite{sendik_deep_2017}},
y grid style={white!80!black},
ylabel={Winning Prob},
ymajorgrids,
ymin=0, ymax=1,
ytick style={color=white!15!black},
ytick={0,0.2,0.4,0.6,0.8,1},
yticklabels={\(\displaystyle 0.0\),\(\displaystyle 0.2\),\(\displaystyle 0.4\),\(\displaystyle 0.6\),\(\displaystyle 0.8\),\(\displaystyle 1.0\)}
]
\draw[draw=white,fill=color0,opacity=0.5] (axis cs:-0.4,0) rectangle (axis cs:0.4,0.183770275727956);
\draw[draw=white,fill=color1,opacity=0.5] (axis cs:0.6,0) rectangle (axis cs:1.4,0.765178861077769);
\draw[draw=white,fill=color2,opacity=0.5] (axis cs:1.6,0) rectangle (axis cs:2.4,0.742600395460819);
\draw[draw=white,fill=color3,opacity=0.5] (axis cs:2.6,0) rectangle (axis cs:3.4,0.631716970510367);
\draw[draw=white,fill=color4,opacity=0.5] (axis cs:3.6,0) rectangle (axis cs:4.4,0.176733497223088);
\path [draw=black, semithick]
(axis cs:0,0.17693304889855)
--(axis cs:0,0.190607502557363);

\path [draw=black, semithick]
(axis cs:1,0.757621393056576)
--(axis cs:1,0.772736329098962);

\path [draw=black, semithick]
(axis cs:2,0.734963402291551)
--(axis cs:2,0.750237388630088);

\path [draw=black, semithick]
(axis cs:3,0.623882466192626)
--(axis cs:3,0.639551474828108);

\path [draw=black, semithick]
(axis cs:4,0.169935955359951)
--(axis cs:4,0.183531039086224);

\addplot [semithick, black, mark=-, mark size=10, mark options={solid}, only marks]
table {%
0 0.17693304889855
1 0.757621393056576
2 0.734963402291551
3 0.623882466192626
4 0.169935955359951
};
\addplot [semithick, black, mark=-, mark size=10, mark options={solid}, only marks]
table {%
0 0.190607502557363
1 0.772736329098962
2 0.750237388630088
3 0.639551474828108
4 0.183531039086224
};
\end{axis}

\end{tikzpicture}}
     \end{minipage}
      \begin{minipage}[b]{0.3\textwidth}
   \centering \small{ Regular images } \\
   \resizebox{\columnwidth}{!}{
 \begin{tikzpicture}

\definecolor{color0}{rgb}{0.215686274509804,0.494117647058824,0.72156862745098}
\definecolor{color1}{rgb}{1,0.498039215686275,0}
\definecolor{color2}{rgb}{0.301960784313725,0.686274509803922,0.290196078431373}
\definecolor{color3}{rgb}{0.968627450980392,0.505882352941176,0.749019607843137}
\definecolor{color4}{rgb}{0.650980392156863,0.337254901960784,0.156862745098039}

\begin{axis}[
axis line style={white!80!black},
tick align=outside,
tick pos=both,
x grid style={white!80!black},
xmajorgrids,
xmin=-0.64, xmax=4.64,
xtick style={color=white!15!black},
xtick={0,1,2,3,4},
xticklabel style = {rotate=45.0,align=center},
xticklabels={Gatys \cite{gatys_texture_2015},{Gram +\\ MSInit},{Gram +\\ Spectrum +\\ MSInit},Snelgrove \cite{snelgrove_highresolution_2017},Deep Corr \cite{sendik_deep_2017}},
y grid style={white!80!black},
ylabel={Winning Prob},
ymajorgrids,
ymin=0, ymax=1,
ytick style={color=white!15!black},
ytick={0,0.2,0.4,0.6,0.8,1},
yticklabels={\(\displaystyle 0.0\),\(\displaystyle 0.2\),\(\displaystyle 0.4\),\(\displaystyle 0.6\),\(\displaystyle 0.8\),\(\displaystyle 1.0\)}
]
\draw[draw=white,fill=color0,opacity=0.5] (axis cs:-0.4,0) rectangle (axis cs:0.4,0.166285651519951);
\draw[draw=white,fill=color1,opacity=0.5] (axis cs:0.6,0) rectangle (axis cs:1.4,0.706603844671324);
\draw[draw=white,fill=color2,opacity=0.5] (axis cs:1.6,0) rectangle (axis cs:2.4,0.868951357114896);
\draw[draw=white,fill=color3,opacity=0.5] (axis cs:2.6,0) rectangle (axis cs:3.4,0.553112450864799);
\draw[draw=white,fill=color4,opacity=0.5] (axis cs:3.6,0) rectangle (axis cs:4.4,0.205046695829029);
\path [draw=black, semithick]
(axis cs:0,0.156650766888364)
--(axis cs:0,0.175920536151537);

\path [draw=black, semithick]
(axis cs:1,0.696078162829777)
--(axis cs:1,0.717129526512871);

\path [draw=black, semithick]
(axis cs:2,0.860215602157201)
--(axis cs:2,0.877687112072591);

\path [draw=black, semithick]
(axis cs:3,0.542502252778937)
--(axis cs:3,0.563722648950662);

\path [draw=black, semithick]
(axis cs:4,0.195030104637514)
--(axis cs:4,0.215063287020544);

\addplot [semithick, black, mark=-, mark size=10, mark options={solid}, only marks]
table {%
0 0.156650766888364
1 0.696078162829777
2 0.860215602157201
3 0.542502252778937
4 0.195030104637514
};
\addplot [semithick, black, mark=-, mark size=10, mark options={solid}, only marks]
table {%
0 0.175920536151537
1 0.717129526512871
2 0.877687112072591
3 0.563722648950662
4 0.215063287020544
};
\end{axis}

\end{tikzpicture}}
     \end{minipage}
      \begin{minipage}[b]{0.3\textwidth}
   \centering \small{ Irregular images } \\
   \resizebox{\columnwidth}{!}{
 \begin{tikzpicture}

\definecolor{color0}{rgb}{0.215686274509804,0.494117647058824,0.72156862745098}
\definecolor{color1}{rgb}{1,0.498039215686275,0}
\definecolor{color2}{rgb}{0.301960784313725,0.686274509803922,0.290196078431373}
\definecolor{color3}{rgb}{0.968627450980392,0.505882352941176,0.749019607843137}
\definecolor{color4}{rgb}{0.650980392156863,0.337254901960784,0.156862745098039}

\begin{axis}[
axis line style={white!80!black},
tick align=outside,
tick pos=both,
x grid style={white!80!black},
xmajorgrids,
xmin=-0.64, xmax=4.64,
xtick style={color=white!15!black},
xtick={0,1,2,3,4},
xticklabel style = {rotate=45.0,align=center},
xticklabels={Gatys \cite{gatys_texture_2015},{Gram +\\ MSInit},{Gram +\\ Spectrum +\\ MSInit},Snelgrove \cite{snelgrove_highresolution_2017},Deep Corr \cite{sendik_deep_2017}},
y grid style={white!80!black},
ylabel={Winning Prob},
ymajorgrids,
ymin=0, ymax=1,
ytick style={color=white!15!black},
ytick={0,0.2,0.4,0.6,0.8,1},
yticklabels={\(\displaystyle 0.0\),\(\displaystyle 0.2\),\(\displaystyle 0.4\),\(\displaystyle 0.6\),\(\displaystyle 0.8\),\(\displaystyle 1.0\)}
]
\draw[draw=white,fill=color0,opacity=0.5] (axis cs:-0.4,0) rectangle (axis cs:0.4,0.183770275727956);
\draw[draw=white,fill=color1,opacity=0.5] (axis cs:0.6,0) rectangle (axis cs:1.4,0.765178861077769);
\draw[draw=white,fill=color2,opacity=0.5] (axis cs:1.6,0) rectangle (axis cs:2.4,0.742600395460819);
\draw[draw=white,fill=color3,opacity=0.5] (axis cs:2.6,0) rectangle (axis cs:3.4,0.631716970510367);
\draw[draw=white,fill=color4,opacity=0.5] (axis cs:3.6,0) rectangle (axis cs:4.4,0.176733497223088);
\path [draw=black, semithick]
(axis cs:0,0.17693304889855)
--(axis cs:0,0.190607502557363);

\path [draw=black, semithick]
(axis cs:1,0.757621393056576)
--(axis cs:1,0.772736329098962);

\path [draw=black, semithick]
(axis cs:2,0.734963402291551)
--(axis cs:2,0.750237388630088);

\path [draw=black, semithick]
(axis cs:3,0.623882466192626)
--(axis cs:3,0.639551474828108);

\path [draw=black, semithick]
(axis cs:4,0.169935955359951)
--(axis cs:4,0.183531039086224);

\addplot [semithick, black, mark=-, mark size=10, mark options={solid}, only marks]
table {%
0 0.17693304889855
1 0.757621393056576
2 0.734963402291551
3 0.623882466192626
4 0.169935955359951
};
\addplot [semithick, black, mark=-, mark size=10, mark options={solid}, only marks]
table {%
0 0.190607502557363
1 0.772736329098962
2 0.750237388630088
3 0.639551474828108
4 0.183531039086224
};
\end{axis}

\end{tikzpicture}}
     \end{minipage}
 \caption{Winning probabilities $W_i$ with standard error $\Sigma_i$ for the different methods for the global case.}   
 \label{fig:Wi_global}
 \end{figure}
 
\begin{figure}[!tbp]
\centering
    Local case \\
\begin{minipage}[b]{0.3\textwidth}
   \centering  \small{All images} \\
   \resizebox{\columnwidth}{!}{
 \begin{tikzpicture}

\definecolor{color0}{rgb}{0.215686274509804,0.494117647058824,0.72156862745098}
\definecolor{color1}{rgb}{1,0.498039215686275,0}
\definecolor{color2}{rgb}{0.301960784313725,0.686274509803922,0.290196078431373}
\definecolor{color3}{rgb}{0.968627450980392,0.505882352941176,0.749019607843137}
\definecolor{color4}{rgb}{0.650980392156863,0.337254901960784,0.156862745098039}

\begin{axis}[
axis line style={white!80!black},
tick align=outside,
tick pos=both,
x grid style={white!80!black},
xmajorgrids,
xmin=-0.64, xmax=4.64,
xtick style={color=white!15!black},
xtick={0,1,2,3,4},
xticklabel style = {rotate=45.0,align=center},
xticklabels={Gatys \cite{gatys_texture_2015},{Gram +\\ MSInit},{Gram +\\ Spectrum +\\ MSInit},Snelgrove \cite{snelgrove_highresolution_2017},Deep Corr \cite{sendik_deep_2017}},
y grid style={white!80!black},
ylabel={Winning Prob},
ymajorgrids,
ymin=0, ymax=1,
ytick style={color=white!15!black},
ytick={0,0.2,0.4,0.6,0.8,1},
yticklabels={\(\displaystyle 0.0\),\(\displaystyle 0.2\),\(\displaystyle 0.4\),\(\displaystyle 0.6\),\(\displaystyle 0.8\),\(\displaystyle 1.0\)}
]
\draw[draw=white,fill=color0,opacity=0.5] (axis cs:-0.4,0) rectangle (axis cs:0.4,0.3787127374193);
\draw[draw=white,fill=color1,opacity=0.5] (axis cs:0.6,0) rectangle (axis cs:1.4,0.774216031987564);
\draw[draw=white,fill=color2,opacity=0.5] (axis cs:1.6,0) rectangle (axis cs:2.4,0.617248299916139);
\draw[draw=white,fill=color3,opacity=0.5] (axis cs:2.6,0) rectangle (axis cs:3.4,0.631192710393835);
\draw[draw=white,fill=color4,opacity=0.5] (axis cs:3.6,0) rectangle (axis cs:4.4,0.0986302202831629);
\path [draw=black, semithick]
(axis cs:0,0.370976856043788)
--(axis cs:0,0.386448618794811);

\path [draw=black, semithick]
(axis cs:1,0.766799303043799)
--(axis cs:1,0.781632760931328);

\path [draw=black, semithick]
(axis cs:2,0.609045363310448)
--(axis cs:2,0.62545123652183);

\path [draw=black, semithick]
(axis cs:3,0.622989728263861)
--(axis cs:3,0.639395692523809);

\path [draw=black, semithick]
(axis cs:4,0.0935524536099125)
--(axis cs:4,0.103707986956413);

\addplot [semithick, black, mark=-, mark size=10, mark options={solid}, only marks]
table {%
0 0.370976856043788
1 0.766799303043799
2 0.609045363310448
3 0.622989728263861
4 0.0935524536099125
};
\addplot [semithick, black, mark=-, mark size=10, mark options={solid}, only marks]
table {%
0 0.386448618794811
1 0.781632760931328
2 0.62545123652183
3 0.639395692523809
4 0.103707986956413
};
\end{axis}

\end{tikzpicture}}
     \end{minipage}
      \begin{minipage}[b]{0.3\textwidth}
   \centering \small{Regular images} \\
   \resizebox{\columnwidth}{!}{
 \begin{tikzpicture}

\definecolor{color0}{rgb}{0.215686274509804,0.494117647058824,0.72156862745098}
\definecolor{color1}{rgb}{1,0.498039215686275,0}
\definecolor{color2}{rgb}{0.301960784313725,0.686274509803922,0.290196078431373}
\definecolor{color3}{rgb}{0.968627450980392,0.505882352941176,0.749019607843137}
\definecolor{color4}{rgb}{0.650980392156863,0.337254901960784,0.156862745098039}

\begin{axis}[
axis line style={white!80!black},
tick align=outside,
tick pos=both,
x grid style={white!80!black},
xmajorgrids,
xmin=-0.64, xmax=4.64,
xtick style={color=white!15!black},
xtick={0,1,2,3,4},
xticklabel style = {rotate=45.0,align=center},
xticklabels={Gatys \cite{gatys_texture_2015},{Gram +\\ MSInit},{Gram +\\ Spectrum +\\ MSInit},Snelgrove \cite{snelgrove_highresolution_2017},Deep Corr \cite{sendik_deep_2017}},
y grid style={white!80!black},
ylabel={Winning Prob},
ymajorgrids,
ymin=0, ymax=1,
ytick style={color=white!15!black},
ytick={0,0.2,0.4,0.6,0.8,1},
yticklabels={\(\displaystyle 0.0\),\(\displaystyle 0.2\),\(\displaystyle 0.4\),\(\displaystyle 0.6\),\(\displaystyle 0.8\),\(\displaystyle 1.0\)}
]
\draw[draw=white,fill=color0,opacity=0.5] (axis cs:-0.4,0) rectangle (axis cs:0.4,0.27282917993662);
\draw[draw=white,fill=color1,opacity=0.5] (axis cs:0.6,0) rectangle (axis cs:1.4,0.784166190578217);
\draw[draw=white,fill=color2,opacity=0.5] (axis cs:1.6,0) rectangle (axis cs:2.4,0.657350402255334);
\draw[draw=white,fill=color3,opacity=0.5] (axis cs:2.6,0) rectangle (axis cs:3.4,0.670890308482021);
\draw[draw=white,fill=color4,opacity=0.5] (axis cs:3.6,0) rectangle (axis cs:4.4,0.114763918747808);
\path [draw=black, semithick]
(axis cs:0,0.262811445973641)
--(axis cs:0,0.282846913899599);

\path [draw=black, semithick]
(axis cs:1,0.773721722383465)
--(axis cs:1,0.794610658772969);

\path [draw=black, semithick]
(axis cs:2,0.646124104875714)
--(axis cs:2,0.668576699634954);

\path [draw=black, semithick]
(axis cs:3,0.659604974599173)
--(axis cs:3,0.682175642364868);

\path [draw=black, semithick]
(axis cs:4,0.106514172707956)
--(axis cs:4,0.12301366478766);

\addplot [semithick, black, mark=-, mark size=10, mark options={solid}, only marks]
table {%
0 0.262811445973641
1 0.773721722383465
2 0.646124104875714
3 0.659604974599173
4 0.106514172707956
};
\addplot [semithick, black, mark=-, mark size=10, mark options={solid}, only marks]
table {%
0 0.282846913899599
1 0.794610658772969
2 0.668576699634954
3 0.682175642364868
4 0.12301366478766
};
\end{axis}

\end{tikzpicture}}
     \end{minipage}
      \begin{minipage}[b]{0.3\textwidth}
   \centering \small{Irregular images}  \\
   \resizebox{\columnwidth}{!}{
 \begin{tikzpicture}

\definecolor{color0}{rgb}{0.215686274509804,0.494117647058824,0.72156862745098}
\definecolor{color1}{rgb}{1,0.498039215686275,0}
\definecolor{color2}{rgb}{0.301960784313725,0.686274509803922,0.290196078431373}
\definecolor{color3}{rgb}{0.968627450980392,0.505882352941176,0.749019607843137}
\definecolor{color4}{rgb}{0.650980392156863,0.337254901960784,0.156862745098039}

\begin{axis}[
axis line style={white!80!black},
tick align=outside,
tick pos=both,
x grid style={white!80!black},
xmajorgrids,
xmin=-0.64, xmax=4.64,
xtick style={color=white!15!black},
xtick={0,1,2,3,4},
xticklabel style = {rotate=45.0,align=center},
xticklabels={Gatys \cite{gatys_texture_2015},{Gram +\\ MSInit},{Gram +\\ Spectrum +\\ MSInit},Snelgrove \cite{snelgrove_highresolution_2017},Deep Corr \cite{sendik_deep_2017}},
y grid style={white!80!black},
ylabel={Winning Prob},
ymajorgrids,
ymin=0, ymax=1,
ytick style={color=white!15!black},
ytick={0,0.2,0.4,0.6,0.8,1},
yticklabels={\(\displaystyle 0.0\),\(\displaystyle 0.2\),\(\displaystyle 0.4\),\(\displaystyle 0.6\),\(\displaystyle 0.8\),\(\displaystyle 1.0\)}
]
\draw[draw=white,fill=color0,opacity=0.5] (axis cs:-0.4,0) rectangle (axis cs:0.4,0.3787127374193);
\draw[draw=white,fill=color1,opacity=0.5] (axis cs:0.6,0) rectangle (axis cs:1.4,0.774216031987564);
\draw[draw=white,fill=color2,opacity=0.5] (axis cs:1.6,0) rectangle (axis cs:2.4,0.617248299916139);
\draw[draw=white,fill=color3,opacity=0.5] (axis cs:2.6,0) rectangle (axis cs:3.4,0.631192710393835);
\draw[draw=white,fill=color4,opacity=0.5] (axis cs:3.6,0) rectangle (axis cs:4.4,0.0986302202831629);
\path [draw=black, semithick]
(axis cs:0,0.370976856043788)
--(axis cs:0,0.386448618794811);

\path [draw=black, semithick]
(axis cs:1,0.766799303043799)
--(axis cs:1,0.781632760931328);

\path [draw=black, semithick]
(axis cs:2,0.609045363310448)
--(axis cs:2,0.62545123652183);

\path [draw=black, semithick]
(axis cs:3,0.622989728263861)
--(axis cs:3,0.639395692523809);

\path [draw=black, semithick]
(axis cs:4,0.0935524536099125)
--(axis cs:4,0.103707986956413);

\addplot [semithick, black, mark=-, mark size=10, mark options={solid}, only marks]
table {%
0 0.370976856043788
1 0.766799303043799
2 0.609045363310448
3 0.622989728263861
4 0.0935524536099125
};
\addplot [semithick, black, mark=-, mark size=10, mark options={solid}, only marks]
table {%
0 0.386448618794811
1 0.781632760931328
2 0.62545123652183
3 0.639395692523809
4 0.103707986956413
};
\end{axis}

\end{tikzpicture}}
     \end{minipage}
\caption{Winning probabilities $W_i$ with standard error $\Sigma_i$ for the different methods for the local case.}   
\label{fig:Wi_local}
\end{figure}

\subsection{Influence of parameters}
\label{sec:EvalParam}

In this section, we display experiments illustrating the effects of two parameters of the proposed method : $K$, the number of considered scales, and $\beta$, the weighting of the spectrum term when using the method "Gram+Spectrum+MSInit".

\subsubsection{Multi-scale strategy}

\begin{figure}[!ht]
\centering
\begin{minipage}[b]{0.12\linewidth}
\centering Reference \\
\end{minipage}
\begin{minipage}[c]{0.21\linewidth}
\includegraphics[width=\linewidth]{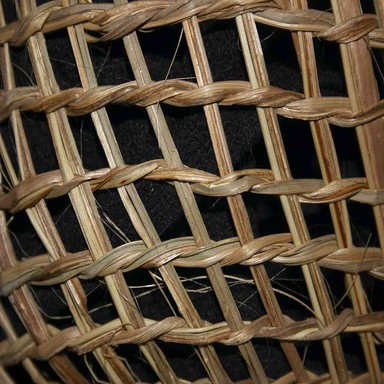}
\end{minipage}
\begin{minipage}[c]{0.21\linewidth}
\includegraphics[width=\linewidth]{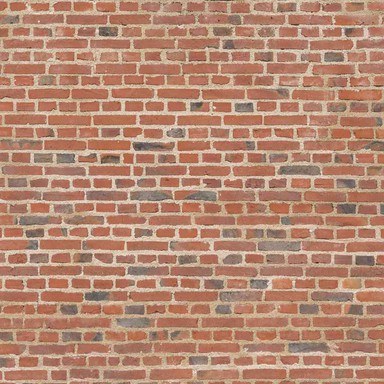}
\end{minipage}
\begin{minipage}[c]{0.21\linewidth}
\includegraphics[width=\linewidth]{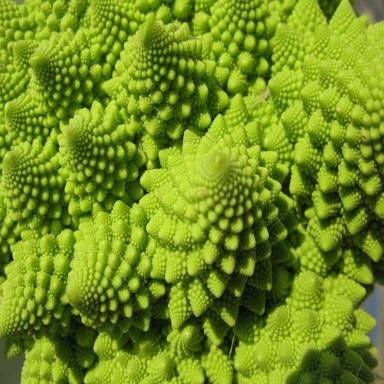}
\end{minipage}
\begin{minipage}[c]{0.21\linewidth}
\includegraphics[width=\linewidth]{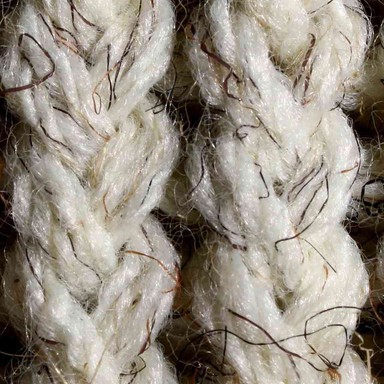}
\end{minipage}
\\
\vspace{1px}
\begin{minipage}[b]{0.12\linewidth}
\centering K = 0 \\
\end{minipage}
\begin{minipage}[c]{0.21\linewidth}
\includegraphics[width=\linewidth]{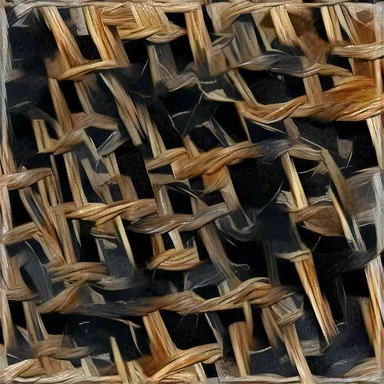}
\end{minipage}
\begin{minipage}[c]{0.21\linewidth}
\includegraphics[width=\linewidth]{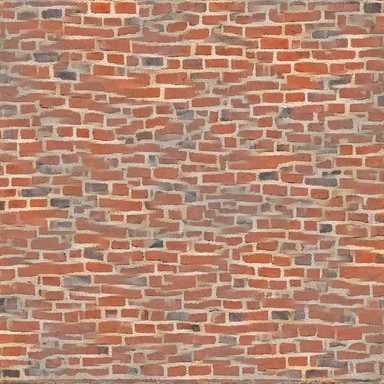}
\end{minipage}
\begin{minipage}[c]{0.21\linewidth}
\includegraphics[width=\linewidth]{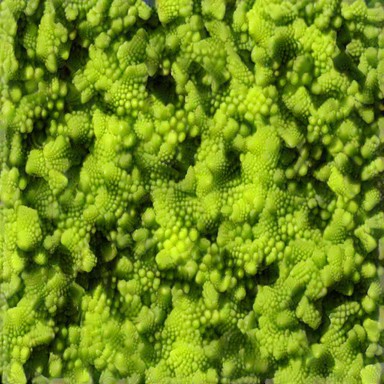}
\end{minipage}
\begin{minipage}[c]{0.21\linewidth}
\includegraphics[width=\linewidth]{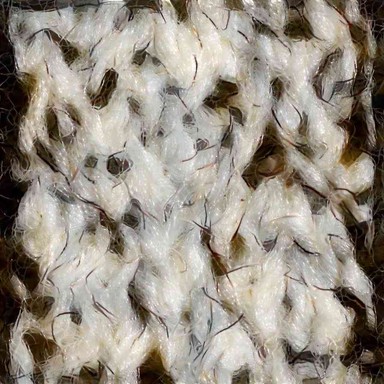}
\end{minipage}
\\
\vspace{1px}
\begin{minipage}[b]{0.12\linewidth}
\centering K = 1 \\
\end{minipage}
\begin{minipage}[c]{0.21\linewidth}
\includegraphics[width=\linewidth]{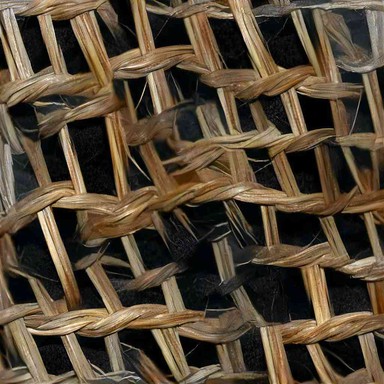}
\end{minipage}
\begin{minipage}[c]{0.21\linewidth}
\includegraphics[width=\linewidth]{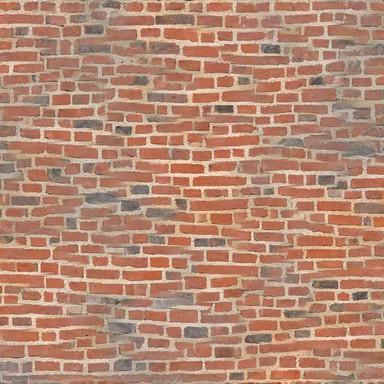}
\end{minipage}
\begin{minipage}[c]{0.21\linewidth}
\includegraphics[width=\linewidth]{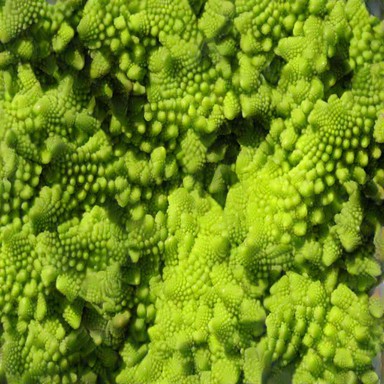}
\end{minipage}
\begin{minipage}[c]{0.21\linewidth}
\includegraphics[width=\linewidth]{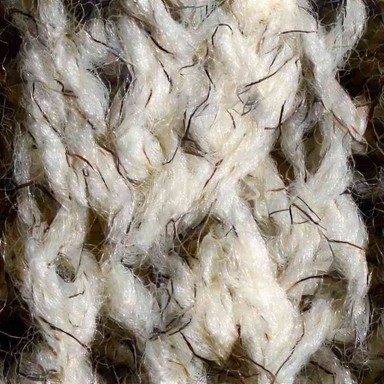}
\end{minipage}
\\
\vspace{1px}
\begin{minipage}[b]{0.12\linewidth}
\centering K = 2 \\
\end{minipage}
\begin{minipage}[c]{0.21\linewidth}
\includegraphics[width=\linewidth]{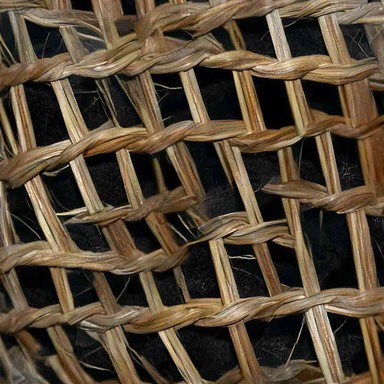}
\end{minipage}
\begin{minipage}[c]{0.21\linewidth}
\includegraphics[width=\linewidth]{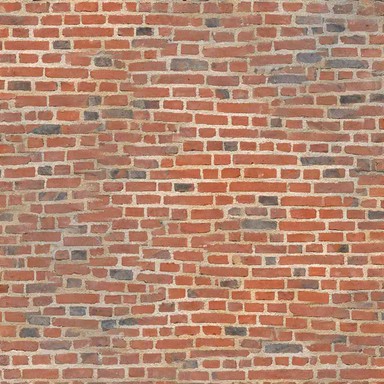}
\end{minipage}
\begin{minipage}[c]{0.21\linewidth}
\includegraphics[width=\linewidth]{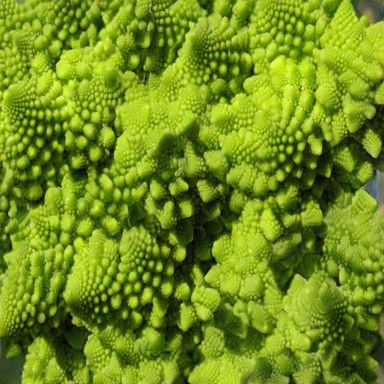}
\end{minipage}
\begin{minipage}[c]{0.21\linewidth}
\includegraphics[width=\linewidth]{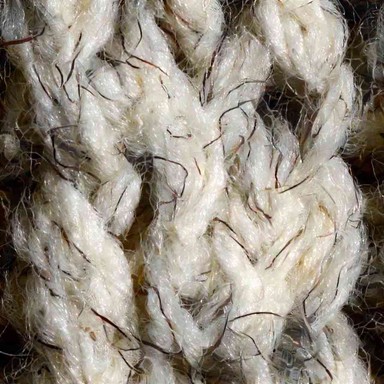}
\end{minipage}
\\
\vspace{1px}
\begin{minipage}[b]{0.12\linewidth}
\centering K = 3 \\
\end{minipage}
\begin{minipage}[c]{0.21\linewidth}
\includegraphics[width=\linewidth]{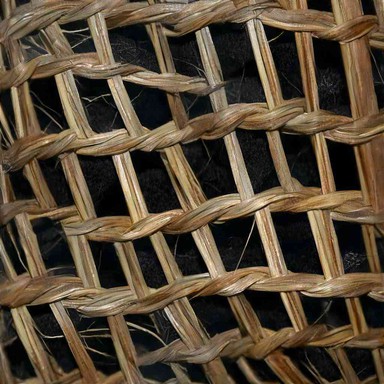}
\end{minipage}
\begin{minipage}[c]{0.21\linewidth}
\includegraphics[width=\linewidth]{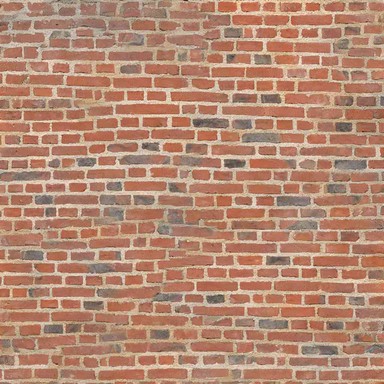}
\end{minipage}
\begin{minipage}[c]{0.21\linewidth}
\includegraphics[width=\linewidth]{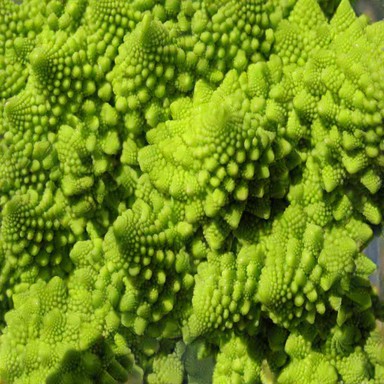}
\end{minipage}
\begin{minipage}[c]{0.21\linewidth}
\includegraphics[width=\linewidth]{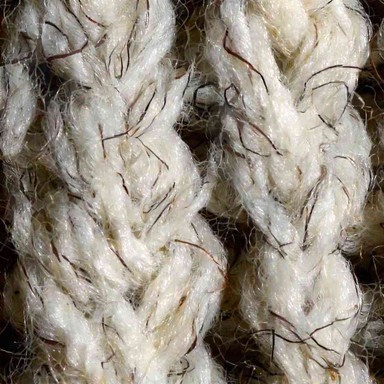}
\end{minipage}
\\
\vspace{1px}
\begin{minipage}[b]{0.12\linewidth}
\centering K = 4 \\
\end{minipage}
\begin{minipage}[c]{0.21\linewidth}
\includegraphics[width=\linewidth]{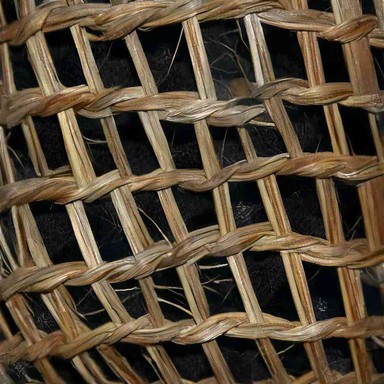}
\end{minipage}
\begin{minipage}[c]{0.21\linewidth}
\includegraphics[width=\linewidth]{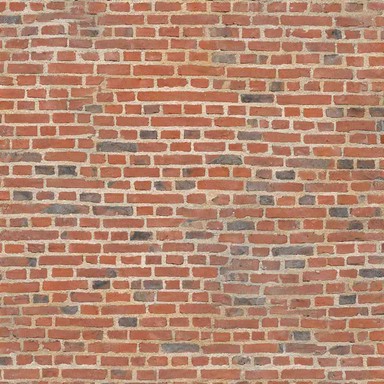}
\end{minipage}
\begin{minipage}[c]{0.21\linewidth}
\includegraphics[width=\linewidth]{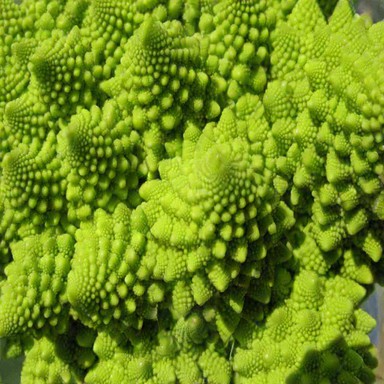}
\end{minipage}
\begin{minipage}[c]{0.21\linewidth}
\includegraphics[width=\linewidth]{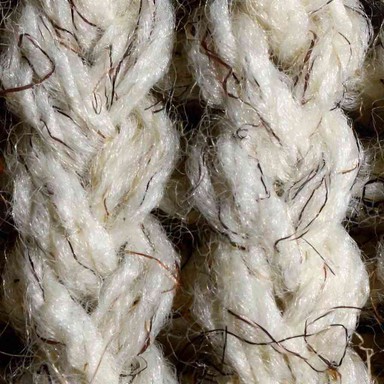}
\end{minipage}
\caption{Synthesis results using different numbers of scales $K$ in the multi-scale strategy. The case $K=0$ corresponds to the original method from~\cite{gatys_texture_2015}.}
\label{fig:multiscalesyn_dif_K}
\end{figure}

In \cref{fig:multiscalesyn_dif_K}, we display synthesis results with $K$ ranging from 0 (original method from\cite{gatys_texture_2015}) to 4. The quality of results increases up to $K=2$. This confirms the fact that the size of the filters in the VGG19 network is too small to describe large scales. Its also illustrates the fact than the VGG filters are versatile and provide good features at different scales, since the network has been trained on $224 \times 224$ input images. An interesting experiment in this respect would be to synthesize textures using the scale-invariant features from~\cite{vannoord_learning_2017}. 

From $K=3$, the method starts to produce results that are very similar to the reference, the case $K=4$ being almost a copy of the reference. This may be due to the fact that in these cases, the number of parameters of the synthesis model is up to two orders of magnitude larger than the number of pixels of the coarse image. In other words, the multi-scale strategy reduces too much the solution space for this optimization problem. In practice, $K=2$ appears a good choice for synthesizing $1024 \times 1024$ images.

\subsubsection{Weighting of the spectrum constraint}
\label{sec:spectrumConstraint}

\begin{figure}[!ht]
\centering
\begin{minipage}[c]{0.1\linewidth}
$\beta = 0$
\end{minipage}
\begin{minipage}[c]{0.21\linewidth}
\includegraphics[width=\linewidth]{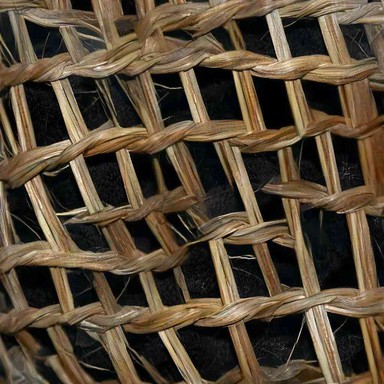}
\end{minipage}
\begin{minipage}[c]{0.21\linewidth}
\includegraphics[width=\linewidth]{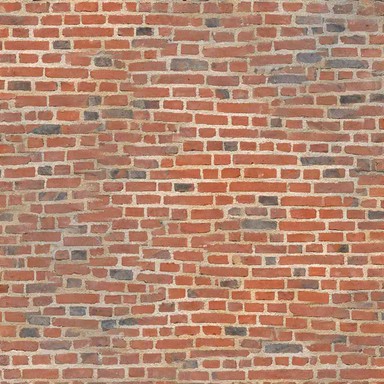}
\end{minipage}
\begin{minipage}[c]{0.21\linewidth}
\includegraphics[width=\linewidth]{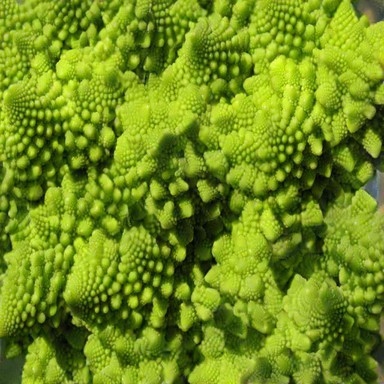}
\end{minipage}
\begin{minipage}[c]{0.21\linewidth}
\includegraphics[width=\linewidth]{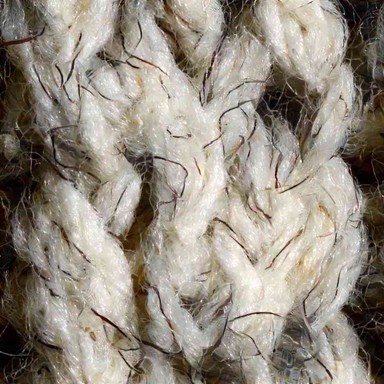}
\end{minipage}
\begin{minipage}[c]{0.1\linewidth}
$\beta = 0.1$
\end{minipage}
\begin{minipage}[c]{0.21\linewidth}
\includegraphics[width=\linewidth]{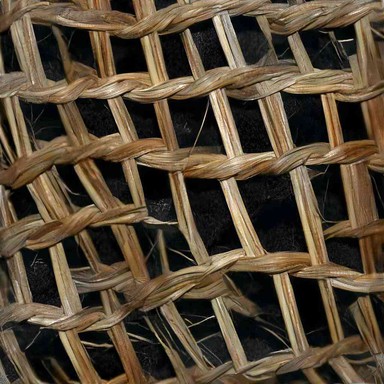}
\end{minipage}
\begin{minipage}[c]{0.21\linewidth}
\includegraphics[width=\linewidth]{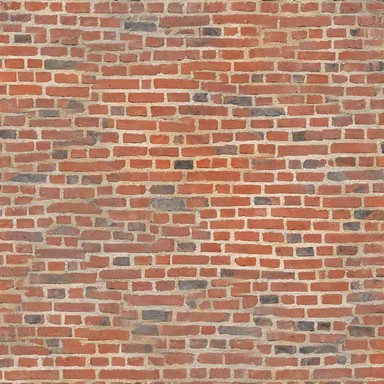}
\end{minipage}
\begin{minipage}[c]{0.21\linewidth}
\includegraphics[width=\linewidth]{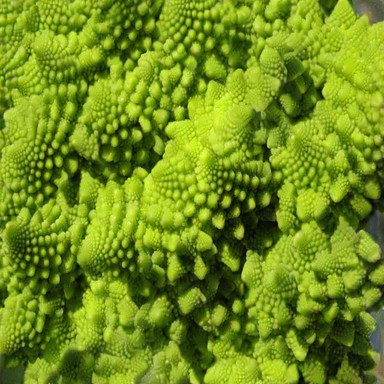}
\end{minipage}
\begin{minipage}[c]{0.21\linewidth}
\includegraphics[width=\linewidth]{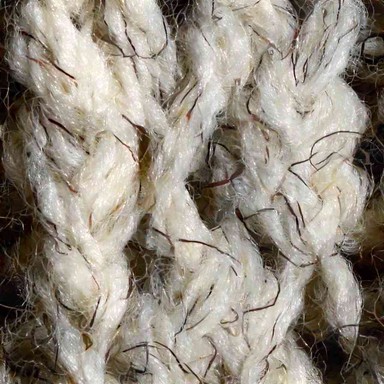}
\end{minipage}
\begin{minipage}[c]{0.1\linewidth}
$\beta = 10^2$
\end{minipage}
\begin{minipage}[c]{0.21\linewidth}
\includegraphics[width=\linewidth]{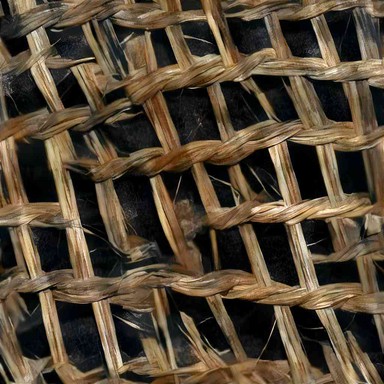}
\end{minipage}
\begin{minipage}[c]{0.21\linewidth}
\includegraphics[width=\linewidth]{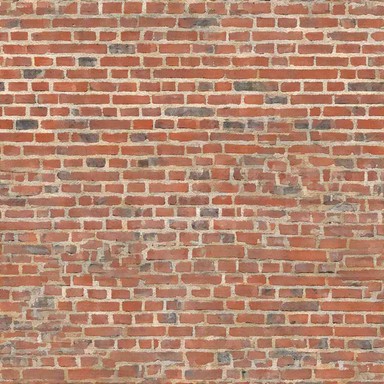}
\end{minipage}
\begin{minipage}[c]{0.21\linewidth}
\includegraphics[width=\linewidth]{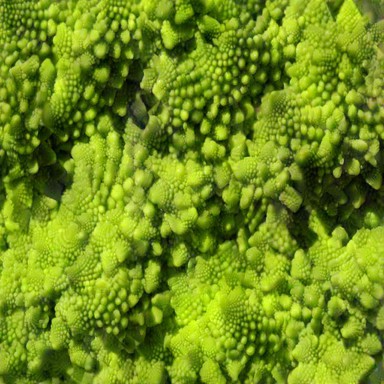}
\end{minipage}
\begin{minipage}[c]{0.21\linewidth}
\includegraphics[width=\linewidth]{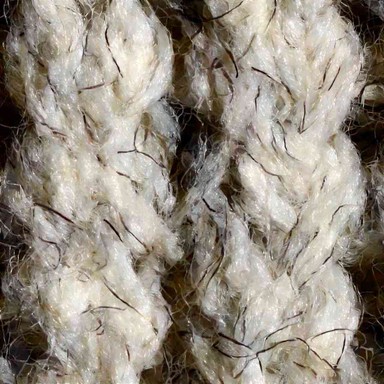}
\end{minipage}
\begin{minipage}[c]{0.1\linewidth}
$\beta = 10^5$
\end{minipage}
\begin{minipage}[c]{0.21\linewidth}
\includegraphics[width=\linewidth]{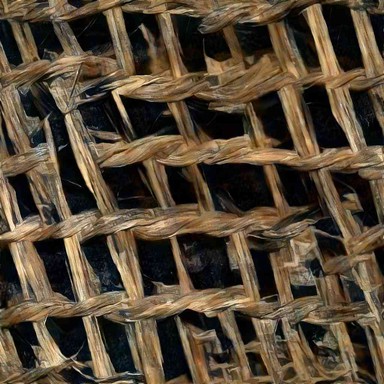}
\end{minipage}
\begin{minipage}[c]{0.21\linewidth}
\includegraphics[width=\linewidth]{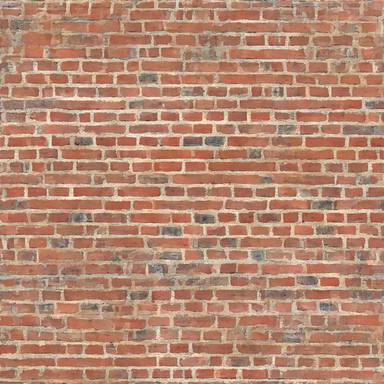}
\end{minipage}
\begin{minipage}[c]{0.21\linewidth}
\includegraphics[width=\linewidth]{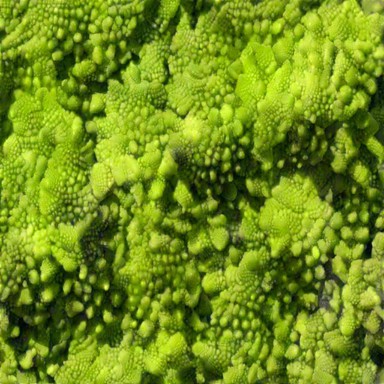}
\end{minipage}
\begin{minipage}[c]{0.21\linewidth}
\includegraphics[width=\linewidth]{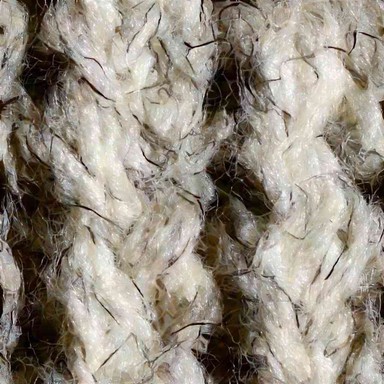}
\end{minipage}
\begin{minipage}[c]{0.1\linewidth}
$\beta = 10^8$
\end{minipage}
\begin{minipage}[c]{0.21\linewidth}
\includegraphics[width=\linewidth]{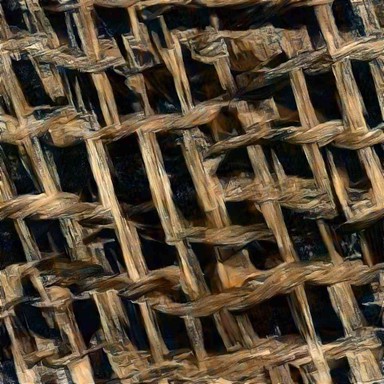}
\end{minipage}
\begin{minipage}[c]{0.21\linewidth}
\includegraphics[width=\linewidth]{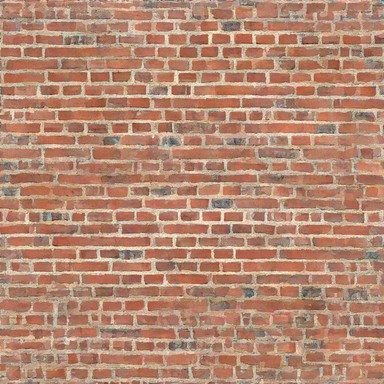}
\end{minipage}
\begin{minipage}[c]{0.21\linewidth}
\includegraphics[width=\linewidth]{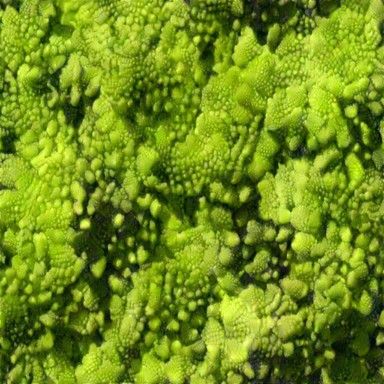}
\end{minipage}
\begin{minipage}[c]{0.21\linewidth}
\includegraphics[width=\linewidth]{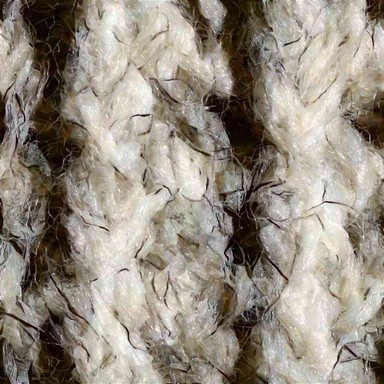}
\end{minipage}
\\
\caption{Synthesis results using different $\beta$ in Formula~\eqref{eq:lossall} (original can be seen in Fig.~\ref{fig:multiscalesyn_dif_K}), $\beta=0,10^{-1},10^2,10^{5},10^{8}$ with the multi-scale strategy and $K=2$.}
\label{fig:diff_beta_msinit}
\end{figure}

In \cref{fig:diff_beta_msinit}, we display the result of the synthesis for different values of $\beta$, the parameter weighting the Spectrum constraint, using the method "Gram+Spectrum+MSInit". For the structured textures for which the spectrum term is useful, the best results are obtained for a relatively large $\beta$, of the order of $10^5$ for the brick image (second column). For more irregular textures, such high values may deteriorate results. This is in agreement with the results from the previous evaluations, where a value $\beta=10^5$ was used. The problem of automatically setting this parameter is open.

\subsection{High resolution synthesis.}
\label{sec:HD_images}

We conclude this experimental section by showing synthesis results of higher resolution ($1024\times 1024$). We consider methods "Gatys", "Gram+MSInit", "Gram+Spectrum+MSInit" (both using $K=3$). %
 The results can be seen  \Cref{fig:pebbles_2048,fig:Tartan_2048}. More results can be seen in Supplementary Materials.%
 Unsurprisingly the interest of the multi-scale schemes is even stronger in this case and the mono-scale method fails. \Cref{fig:Tartan_2048} shows the ability of the spectrum constraint to enforce large scale regularity at this resolution.

\begin{figure}[!ht]
\centering
\begin{minipage}[b]{0.225\linewidth}
\centering Reference \\
\includegraphics[width=\linewidth]{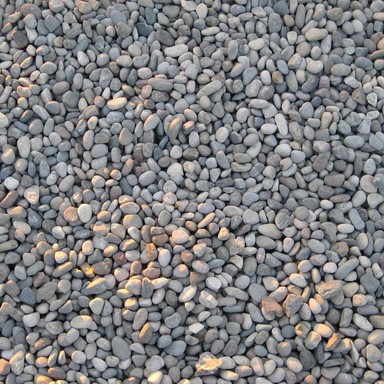}
\end{minipage}
\begin{minipage}[b]{0.225\linewidth}
\centering Gatys \cite{gatys_texture_2015} \\
\includegraphics[width=\linewidth]{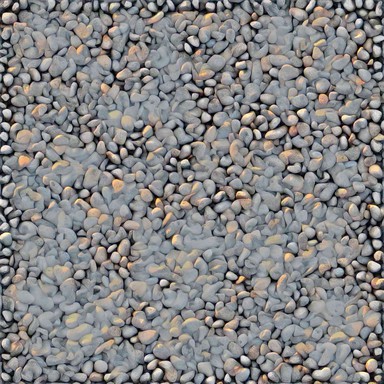}
\end{minipage}
\begin{minipage}[b]{0.225\linewidth}
\centering Gram + MSInit \\
\includegraphics[width=\linewidth]{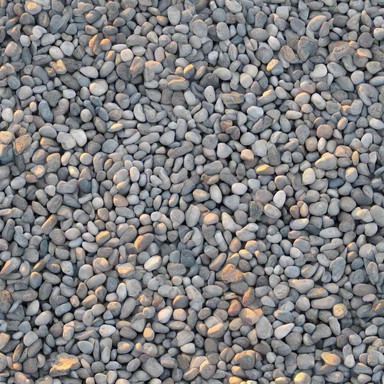}
\end{minipage}
\begin{minipage}[b]{0.225\linewidth}
\centering Gram + Spectrum + MSInit \\
\includegraphics[width=\linewidth]{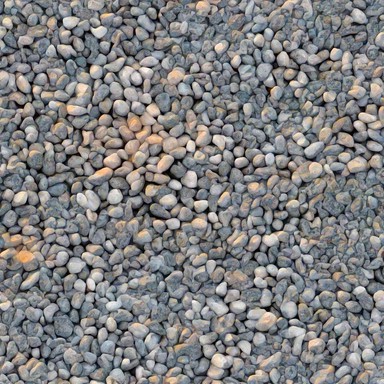}
\end{minipage}
\caption{Synthesis results using different methods for one given reference of size $2048 \times 2048$.}
\label{fig:pebbles_2048}
\end{figure}
\begin{figure}[!ht]
\centering
\begin{minipage}[b]{0.225\linewidth}
\centering Reference \\
\includegraphics[width=\linewidth]{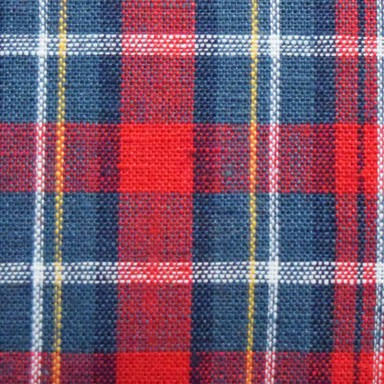}
\end{minipage}
\begin{minipage}[b]{0.225\linewidth}
\centering Gatys \cite{gatys_texture_2015} \\
\includegraphics[width=\linewidth]{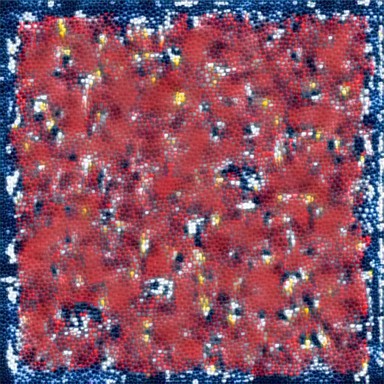}
\end{minipage}
\begin{minipage}[b]{0.225\linewidth}
\centering Gram + MSInit \\
\includegraphics[width=\linewidth]{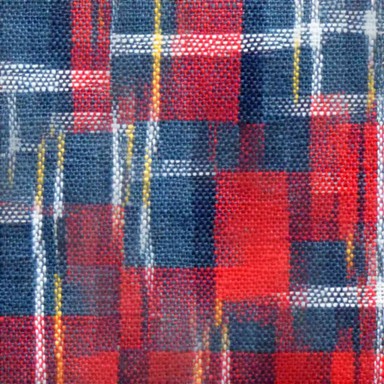}
\end{minipage}
\begin{minipage}[b]{0.225\linewidth}
\centering Gram + Spectrum + MSInit \\
\includegraphics[width=\linewidth]{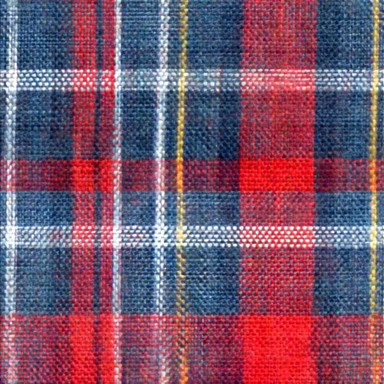}
\end{minipage}
\caption{Synthesis results using different methods for a given reference of size $2048 \times 2048$.}
\label{fig:Tartan_2048}
\end{figure}

\section{Conclusion}

In this paper, we have shown how a multi-resolution framework and additional statistical constraints related to long-range dependency enables one to significantly improve texture synthesis results in comparison to the seminal work~\cite{gatys_texture_2015}, especially for high resolution and possibly regular textures. 
A natural extension would be to investigate the use of such multi-resolution strategies for style transfer for high-resolution images, following \cite{gatys_neural_2015}. More generally, most generative methods dealing with high resolution images incorporate more or less explicitly some multi-resolution steps in their synthesis process. This is for instance the case for the very efficient StyleGan~\cite{karras2019style} approach to face synthesis. In this context, it is of great interest to investigate generic procedures to develop multi-resolution frameworks for such generative approaches.

A strong limitation of the neural methods investigated in this work is the unreasonably large number of parameters of the models. In this respect, the next question is not "what set of statistical constraint is sufficient", but "what is the minimal set of statistical constraints"  needed to produce realistic synthesis. Some works~\cite{debortoli_macrocanonical_2019} have shown that second order statistics between features are not necessary to get satisfying results. This, combined with the highly redundant nature of networks such as VGG19, trained for recognition, suggests that much room is available to reduce the number of parameters in these models.

\section*{Acknowledgments}
This work is supported by the "IDI 2017" project funded by the IDEX Paris-Saclay, ANR-11-IDEX-0003-02 and by the ANR project MISTIC, 19-CE40-0005-01.
We would like to thank Gang Liu, Kiwon Um and Gui-Song Xia for fruitful discussions. 
We would like to acknowledge the participation of volunteers in answering the perceptual evaluation survey.

\bibliographystyle{siamplain}
\bibliography{Texture_Paper}

\end{document}